\documentclass[letterpaper]{article} 
\usepackage{aaai24}  
\usepackage{times}  
\usepackage{helvet}  
\usepackage{courier}  
\usepackage[hyphens]{url}  
\usepackage{graphicx} 
\usepackage{natbib}  
\usepackage{caption} 
\usepackage{algorithm}
\usepackage{algorithmic}

\usepackage{newfloat}
\usepackage{listings}
\DeclareCaptionStyle{ruled}{labelfont=normalfont,labelsep=colon,strut=off} 
\floatstyle{ruled}
\newfloat{listing}{tb}{lst}{}
\floatname{listing}{Listing}
\usepackage{braket}
\usepackage{amsfonts}
\usepackage{amsmath}
\usepackage{physics}
\usepackage{subcaption}
\usepackage{multirow}

\newtheorem{theorem}{Theorem}

\newtheorem{assumption}{Assumption}
\newtheorem{lemma}{Lemma}

\newcommand{\thickhline}{%
    \noalign {\ifnum 0=`}\fi \hrule height 1pt
    \futurelet \reserved@a \@xhline
}

\title{Generating Universal Adversarial Perturbations for Quantum Classifiers}
\author {
    Gautham Anil\textsuperscript{\rm 1}\equalcontrib,
    Vishnu Vinod\textsuperscript{\rm 1}\equalcontrib,
    Apurva Narayan\textsuperscript{\rm 2,3,4}
}
\affiliations {
    \textsuperscript{\rm 1}Indian Institute of Technology Madras\\
    \textsuperscript{\rm 2}University of British Columbia\\
    \textsuperscript{\rm 3}University of Western Ontario\\
    \textsuperscript{\rm 4}University of Waterloo\\
    \{gauthamga.gga,vishnuvinod2001\}@gmail.com, apurva.narayan@uwo.ca
}

\begin{document}

\maketitle

\begin{abstract}
Quantum Machine Learning (QML) has emerged as a promising field of research, aiming to leverage the capabilities of quantum computing to enhance existing machine learning methodologies. Recent studies have revealed that, like their classical counterparts, QML models based on Parametrized Quantum Circuits (PQCs) are also vulnerable to adversarial attacks. Moreover, the existence of Universal Adversarial Perturbations (UAPs) in the quantum domain has been demonstrated theoretically in the context of quantum classifiers. In this work, we introduce QuGAP: a novel framework for generating UAPs for quantum classifiers. We conceptualize the notion of additive UAPs for PQC-based classifiers and theoretically demonstrate their existence. We then utilize generative models (QuGAP-A) to craft additive UAPs and experimentally show that quantum classifiers are susceptible to such attacks. Moreover, we formulate a new method for generating unitary UAPs (QuGAP-U) using quantum generative models and a novel loss function based on fidelity constraints. We evaluate the performance of the proposed framework and show that our method achieves state-of-the-art misclassification rates, while maintaining high fidelity between legitimate and adversarial samples.
\end{abstract}

\section{Introduction} 
Rapid advancements in the field of machine learning have led to development of solutions to problems which were previously intractable \cite{jordan2015machine, lecun2015deep, he2015delving, zhao2023survey, silver2016mastering}. Concurrently, rapid advances are being achieved in the domain of quantum computing. While development of fault-tolerant large-scale quantum computers holds the potential to unravel solutions to classically hard problems \cite{shor1994algorithms, cao2019quantum}, quantum computing has applications even in the current NISQ (Noisy Intermediate Scale Quantum) era \cite{preskill2018quantum, lau2022nisq, bharti2022noisy}.

A promising direction of research, termed quantum machine learning (QML) attempts to bridge the domains of machine learning and quantum computing in an attempt to leverage NISQ devices to enhance machine learning methodologies \cite{biamonte2017quantum, lloyd2014quantum, rebentrost2014quantum, dallaire2018quantum}. Within QML, parametrized quantum circuits (PQCs), recognized as a quantum analogue of classical neural networks, have garnered significant attention in recent times \cite{cerezo2021variational, benedetti2019parameterized}. While the application domains where QML can provide an advantage are being explored, recent advances suggest that classification is a promising candidate. \cite{Huang2021}. The rise in prominence of PQC-based classifiers stems from their capacity to achieve performance comparable to classical neural networks in multiple use cases while utilizing far fewer parameters \cite{abbas2021power, schuld2020circuit}.

Classical machine learning classifiers are well-known to be susceptible to adversarial attacks that severely degrade performance \cite{xu2020adversarial, akhtar2018threat, zhang2020adversarial}. Recent studies reveal that PQC-based classifiers too, are susceptible to adversarial attacks, mirroring the vulnerabilities observed in classical settings \cite{liu2020vulnerability, lu2020quantum}. Akin to the classical ML
, the existence of universal attacks has also been demonstrated in the realm of quantum classifiers \cite{gong2022universal}. In this work, we propose QuGAP (Quantum Generative Adversarial Perturbation), a framework to generate UAPs for quantum classifiers. Our main contributions are:
\begin{itemize}
    \item We theoretically demonstrate the existence of \textit{additive} UAPs for quantum classifiers used in the classification of amplitude-encoded classical data. 
    \item We propose a strategy for generating \textit{additive} UAPs using classical generative models and conduct experiments to validate the viability of the proposed approach.
    \item We propose a novel strategy for generating \textit{unitary} UAPs using explicit fidelity constraints. We empirically evaluate the performance of the proposed framework and achieve state-of-the-art results.
\end{itemize}

\section{Related Work}
\subsection{Adversarial Attack on Neural Networks}
The idea that carefully constructed adversarial samples, which differ only slightly from the true samples, could fool classical deep neural networks was first discussed in \cite{szegedy2014intriguing}. After the conception of this idea, a number of attacks were proposed in the white-box setting, where the adversary has full access to the target classifier. The proposed attacks adopt different strategies, including gradient-based methods \cite{goodfellow2015explaining, madry2019towards}, optimization-based methods \cite{carlini2017towards, liu2017delving} and generative model based methods \cite{xiao2019generating, bai2021ai}. However, these attacks are all input specific attacks as the perturbation applied on each input is different.

The notion of a ``universal'' adversarial perturbation (UAP) was initially introduced in  \cite{dezfooli2017}. The idea was to generate input-agnostic perturbations which can fool the target model in a classification problem across all classes. The idea of generating UAPs using generative networks was proposed in \cite{poursaeed2018}. This method of generating UAPs showed a marked improvement over the iterative method proposed in \cite{dezfooli2017}. Our work takes inspiration from the method proposed in \cite{poursaeed2018} to generate additive UAPs for quantum classifiers.

\subsection{Adversarial Attack on PQC-Based Classifiers}
The vulnerability of quantum classifiers to adversarial attacks has been studied in \cite{liu2020vulnerability, lu2020quantum}. Quantum adversarial attacks and adversarial learning were experimentally demonstrated in \cite{ren2022experimental}. The effect of noise in making quantum classifiers robust to adversarial attacks was studied in \cite{du2021quantum, huang2023certified} whereas the effect of encoding schemes in protecting quantum classifiers was studied in \cite{gong2022enhancing}. The transferability of attacks across classical and quantum classifiers was studied in \cite{west2023benchmarking}. The existence of unitary UAPs for quantum classifiers has recently been theoretically demonstrated in \cite{gong2022universal}; they also propose an iterative scheme for generating UAPs. In this work, we propose a more effective framework for generating unitary UAPs for quantum classifiers.

\section{Background}
We use ``ket'' notation, like $\ket{\psi}$ for instance, to denote a complex column vector. Unless specified otherwise, $\ket{\psi} \in \mathbb{C}^d$ where $\mathbb{C}^d$ is the complex $d-$space and $d$ is a positive integer. In general, a quantum state is represented by its density matrix $\sigma$. A pure quantum state is characterized by a density matrix of rank 1, and therefore can be equivalently represented by a complex column vector $\ket{\psi}$. For the rest of the paper, ``quantum state'' is used to refer to a pure quantum state represented by a column vector $\ket{\psi}$ unless specified otherwise. For a more detailed introduction to quantum states, density matrices and other concepts in quantum computing, we refer the readers to \cite{nielsen2010quantum}.

 \subsubsection{Quantum Classifiers}
 We denote a PQC-based quantum classifier by $\mathcal{Q}$. For a review of such classifiers, we refer the readers to \cite{cerezo2021variational}. The classifier $\mathcal{Q}$ takes in a quantum state $\ket{\psi}$ as input and outputs a label $\mathcal{Q}(\ket{\psi}) \in \{0, 1, ... k-1\}$ for a $k$-class classification problem. Note that the state $\ket{\psi}$ could be from a quantum dataset or could be from an encoded classical dataset. Analogous to classical classifiers, given a dataset $\mathcal{H}$ of samples $\left\{(\ket{\psi_i}, c_{\psi_i})\right\}_{i=1}^N$ where $c_{\psi_i}$ denotes the assigned label of $\ket{\psi_i}$, we loosely say a quantum classifier $\mathcal{Q}$ is trained if it achieves $\mathcal{Q}(\ket{\psi_i}) = c_{\psi_i}$ for ``most'' of the samples in $\mathcal{H}$. The exact accuracy that can be achieved depends on the classifier as well as the dataset. 

\subsubsection{Classical UAPs}
Consider a classical dataset $\mathcal{D}$ of samples $\left\{({x_i}, c_{x_i})\right\}_{i=1}^N$ where $x_i \in \mathbb{R}^d$ and $c_{x_i} $ denotes the assigned label of ${x_i}$. Let $\mathcal{D'} \subset \mathcal{D}$ be the subset of all samples such that $\mathcal{M}(x_j) =  c_{x_j}$ for every $x_j$ in $\mathcal{D'}$ and a trained classical classifier $\mathcal{M}$. In this context, a UAP for $\mathcal{M}$ is an \textit{additive perturbation} $\delta \in \mathbb{R}^d$ such that:
\begin{enumerate}
    \item $\mathcal{M}(x_j + \delta) \neq  c_{x_j}$ 
    \item $|| \delta ||_p \leq \epsilon $
\end{enumerate}
for ``most'' of $x_j$ in $\mathcal{D'}$. Effectiveness of $\delta$ is measured by the misclassification rate, which is the fraction of such $x_j$ in $\mathcal{D'}$. Note that $||.||_p$ denotes $L_p$ norm and $\epsilon$ is a user-defined threshold controlling the strength of the perturbation.  

\subsubsection{Quantum UAPs}
As in the classical case, let $\mathcal{H'} \subset \mathcal{H}$ be the subset of the dataset such that for every $\ket{\psi_j}$ in $\mathcal{H'}$, $\mathcal{Q}(\ket{\psi_{j}}) =  c_{\psi_j}$ for a trained quantum classifier $\textit{Q}$. A quantum UAP for $\textit{Q}$ is a \textit{unitary transformation} $U \in \mathbb{C}^{d \times d}$ such that:
\begin{enumerate}
    \item $\mathcal{Q}(U\ket{\psi_{j}}) \neq  c_{\psi_j}$ 
    \item $\ket{\psi_{j}}$ and $U\ket{\psi_{j}}$ are ``close''
\end{enumerate}
for a significant fraction of $\psi_{j}$. Again, the effectiveness of $U$ is measured by the misclassification rate. In the existing literature, closeness of $\ket{\psi_{j}}$ and $U\ket{\psi_{j}}$ is ensured by constraining $U$ to be close to the identity matrix \cite{gong2022universal}; in this work, we explore the effect of using a fidelity-based loss function instead.

\subsubsection{Encoding Schemes}
To use quantum classifiers for classifying classical data, it is necessary to encode the data into quantum states. There are a number of such proposed schemes for encoding classical data \cite{larose2020robust}. Amplitude encoding is one of the most commonly used data encoding schemes, due to the fact that it offers an exponential advantage in number of qubits required to encode classical data (compared to other quantum encoding schemes); to encode $d$-dimensional classical data we require only $\lceil\log_2 d\rceil$ qubits. The amplitude encoded state $\ket{\psi_{x}}$ for a classical sample $x$ can be computed as
$\ket{\psi_{x}} = \sum_{k = 0}^{d-1} x^{(k)} \ket{k} $
where $x^{(k)}$ is the $k^{\mathrm{th}}$ element of the \textit{normalized} $d-$dimensional vector $x$ and $\left\{ \ket{k} \right\}_{k = 0}^{d-1}$ is the set of computational basis states. While in general the amplitudes corresponding to the computational basis states of a quantum state may be complex, in practice, for classical data, the amplitudes are constrained to be real values.

\subsubsection{Adversarial Loss} 
In order to train generative models for adversarial sample generation, we define an adversarial or fooling loss inspired by the formulations in \cite{xiao2019generating, poursaeed2018}.\\ 
For targeted attacks, the fooling loss is defined as:
$$\mathcal{L}_{fool, targeted} = \sum_{x\in\mathcal{D}}\mathcal{L}_{CE}(\hat{y}_x, t)$$
where $\mathcal{L}_{CE}$ is the cross-entropy loss, $\hat{y}_x$ gives the prediction probabilities for input $x$ and $t$ denotes the target class. For untargeted attacks, two formulations exist: either the target class can be set as the least-likely class and targeted attack can be used or a separate fooling loss can be defined:
$$\mathcal{L}_{fool, untargeted} = - \sum_{x\in\mathcal{D}}\mathcal{L}_{CE}(\hat{y}_x, c_x)$$
where $c_x$ is the label of input $x$. Both formulations were observed to have competitive performance in \cite{poursaeed2018}. In our experiments, we use the latter formulation for untargeted attacks. We use $\mathcal{L}_{fool}$ to denote fooling loss in general. We also reiterate that the effectiveness of an attack on a dataset is measured by its misclassification rate: percentage of correctly predicted samples which are misclassified after the attack. This metric will be used throughout the paper to measure the effectiveness of adversarial attacks.

\section{Additive UAPs}   

\subsubsection{Motivation}
A proposed application of quantum classifiers is to carry out classification tasks on classical data encoded into quantum states \cite{benedetti2019parameterized}. In such cases, we propose to generate additive UAPs for classical data motivated by the following reasons:
\begin{itemize}
    \item Unlike a quantum dataset where the quantum states may be directly accessible, an adversary may not have access to the quantum states after encoding classical data as the encoded data is directly fed to the quantum classifier. 
    \item The classical notion of additive UAPs does not directly carry over to the quantum domain because quantum states, unlike classical data, can only be perturbed using unitary transformations.
\end{itemize}
It is thus valuable to study the effect of classical additive attacks on quantum classifiers. For the rest of the discussion in this section, we assume amplitude encoding of classical data. A justification for this choice may be found in the previous section.

\subsection{Existence of Additive UAPs}
The key idea behind using additive perturbations is to take advantage of the normalization step in the amplitude encoding scheme. By applying a large enough perturbation to a sample and re-normalizing the resulting vector, it might be possible to project the sample to a different decision region of the quantum classifier. We provide an intuitive demonstration of this idea in the supplementary material.

To develop a theoretical framework for formally proving the existence of additive UAPs, we make the following assumptions:

\begin{assumption}
The dataset under consideration is a classical dataset $\left\{({x_i}, c_{x_i})\right\}_{i=1}^N$, where $x_i \in \mathbb{R}^d$ are $d-$dimensional data samples, $c_{x_i} \in \{0, 1, ..., k-1\}$ are the labels for the $k-$class classification problem and $N$ is the total number of labelled data samples.
\end{assumption}

\begin{assumption}
Encoding of classical data into quantum states is carried out using the amplitude encoding scheme. We abuse notation and use $\ket{x}$ to denote the amplitude encoded quantum state corresponding to the data sample $x$. We also assume $x$ is normalized since this is necessary for amplitude encoding.
\end{assumption}

\begin{assumption}
The trained quantum classifier, $\mathcal{Q}$, is a noiseless PQC-based classifier with a total of $D + K$ qubits. $D = \lceil \log_2 d \rceil$ data qubits are used for encoding classical data and $K = \lceil \log_2 k \rceil$ ancillary qubits are used for measuring output probabilities. Inputs are padded with zeros to ensure $d = 2^D$.
\end{assumption}

\begin{assumption}
 $\mathcal{Q}$ assigns a prediction $\hat{c}_x$ to an input sample $x$ as follows: first, a global unitary transformation $U \in \mathbb{C}^{n \times n}$, where n = $2^{D+K}$, is applied to $\ket{x} \otimes \ket{0}^{\otimes K}$ to obtain a state $\ket{y}$. The prediction is then computed as $\hat{c}_x = \arg\max \{ |\braket{\mathbf{1}_{2^D} \otimes {i}} {{y}}|^2 \}_{i = 0}^{k-1} $, where $\mathbf{1}_{2^D}$ is a vector in $\mathbb{C}^{2^D}$ with all its elements as 1 and $\{\ket{i}\}_{i=0}^{k-1}$ are the first $k$ standard basis states of space $\mathbb{R}^{2^K}$ (j$^{\textrm{th}}$ standard basis state is a vector with its $j^{\textrm{th}}$ element equal to 1 and all other elements equal to 0).
 
\end{assumption}
A note regarding notation: all vectors and matrices are zero-indexed. For instance, the first entry of vector $x$ will be denoted as $x_0$ and the top-left entry of matrix $U$ will be denoted as $U_{0,0}$. Also, $\norm{.}$ denotes $L_2$ norm unless specified otherwise. With these assumptions and notations in place, we now state a few important lemmas and the main theorems. All detailed proofs are provided in the supplementary. First, we prove the following lemma:

\begin{lemma}
The probability of an input $x$ being classified as class $c$ is given by $P(\hat{c}_x = c) = x^\dagger M^c x$ where, $x^\dagger$ denotes the conjugate transpose of $x$ and $M^c\in\mathbb{C}^{d\times d}$ is a positive semi-definite matrix given by: $M^{c}_{ij} = \sum_{t=0}^{d-1} \;U_{k't+c,k'i}^* \; U_{k't+c,k'j}$ with $^*$ denoting the complex conjugate and $k' = 2^K$.
\end{lemma}
\textit{Proof sketch:} Straightforward to prove by expanding out and then rewriting the expression in Assumption 4.\\
Now, our objective is to examine whether a perturbation, when added to input samples, can cause the classifier to misclassify most of the input samples. Therefore, it is necessary to examine the effect of such a perturbation on output prediction probabilities. Towards this, we prove the following lemma:

\begin{lemma}
    Let $x,y\in \mathbb{R}^{d}$ such that $\norm{x} = 1$ and $\norm{y} = 1$. If  $\norm{x-y}\leq \epsilon$, where $\epsilon\in\mathbb{R}$, the probability of the point $y$ being classified as $c$ is bounded as
$\abs{P(\hat{c}_{y} = c) - P(\hat{c}_{x} = c)} \leq d\cdot\big(\epsilon^2 + 2\epsilon\big)$.
\end{lemma}
\textit{Proof sketch:} The key idea is to write $y$ as $x + \delta$ for some $\delta$ and use Lemma 1 to express $P(\hat{c}_{y} = c)$ in terms of $P(\hat{c}_{x} = c)$ and some additional terms involving $M^c$. The expression can then be simplified using the triangle inequality and the fact that $M^c$ is PSD. The result then follows from the fact that $\text{Tr}(M^c) \leq d$, where Tr$()$ denotes the trace operator.

Next, we explore how introducing a perturbation impacts the amplitude encoding of data samples. The following lemma is established for large perturbations:

\begin{lemma}
    For any $x\in \mathbb{R}^{d}$ with $\norm{x} = 1$ on which a perturbation $p$ is applied, the resultant vector is given by $p + x $. If we normalize this vector to obtain $y = \frac{p + x}{\norm{p + x}}$, then we have $\norm{\frac{p}{\norm{p}} - y} \;\leq\; \sqrt{2 - 2\sqrt{1 - \frac{1}{\norm{p}^2}}}$ whenever $\norm{p} > 1$.
\end{lemma}
\textit{Proof sketch:} The expression of norm is first expanded out in terms of $p$ and $x$. The bound can then be computed by maximizing the expression and imposing the condition that  $\norm{p} > 1$.

With Lemmas 2 and 3 in place, we are ready to state the following theorem for perturbations with norm greater than 1:

\begin{theorem}
For an additive universal adversarial perturbation $p$ applied on inputs of classifier $\mathcal{Q}$, a strength of perturbation $\norm{p}\in\mathbb{R}$ will cause $\mathcal{Q}$ to classify all inputs as $c$ (class to which $p/\norm{p}$ belongs) if:$$\norm{p} \geq \frac{2}{\epsilon_c\sqrt{4 - \epsilon_c^2}}$$ where $\epsilon_c$ is given by: $$\epsilon_c = \sqrt{1 + \frac{1}{2d}\cdot \Big( \hat{p}^TM^c\hat{p}  - \hat{p}^TM^{c'}\hat{p}}\Big) - 1$$ where $\hat{p} = p/\norm{p}$ and $c'$ is the class with highest output probability for $\hat{p}$ after $c$. 
\end{theorem}
\textit{Proof sketch:} A lower bound is first placed on the probability of classifying the input sample after perturbation as class $c$. Lemma 2 is then invoked to rewrite the bound in terms of $\epsilon_c$. The bound is then related to $\norm{p}$ by using Lemma 3.

By using Theorem 1, it can be concluded that a sufficiently large perturbation can cause the classifier $\mathcal{Q}$ to classify all input samples as belonging to the class of the perturbation. This result is then directly applicable to targeted UAP generation; a sample belonging to the target class can be chosen as the perturbation. This perturbation, when appropriately scaled, can cause $\mathcal{Q}$ to misclassify all samples as belonging to the target class. For untargeted attacks, the target class can be chosen as the least probable class in the dataset.

Next, we focus on the case where the norm of the perturbation is constrained. In such cases, it may not be possible to guarantee a result as strong as Theorem 1. However, it is still possible to characterize the set of inputs for which we can ensure a required classification. To establish such a characterisation, we state a more general version of Lemma 3:

\begin{lemma}
    For any $x\in\mathbb{R}^{d}$ with $\norm{x} = 1$ on which a perturbation $p$, such that $\norm{p} \geq \delta$, is applied, the resultant vector is given by $p + x $. If we normalize this vector to obtain $y = \frac{p + x}{\norm{p + x}}$, then we have $\norm{\hat{p} - y} \;\leq\; \sqrt{2 - 2\Big( \frac{\delta + \hat{p}^Tx}{\sqrt{\delta^2 + 2\delta \hat{p}^Tx + 1}}\Big)}$ where ${\hat{p}} = \frac{p}{\norm{p}}$.
\end{lemma}
\textit{Proof sketch:} Similar to the proof of Lemma 3.

With Lemma 4 in place, we are ready to state the following theorem:

\begin{theorem}
For an additive universal adversarial perturbation $p$ applied on inputs of classifier $\mathcal{Q}$, with the constraint $\norm{p} \leq \delta$, an input $x$ is predicted as belonging to class $c$ (class to which $p/\norm{p}$ belongs) if any one of the following conditions hold true:
\begin{enumerate}
    \item If $\delta \geq \frac{2}{\epsilon_c\sqrt{4 - \epsilon_c^2}}$ 
    \item If $1 \leq \delta < \frac{2}{\epsilon_c\sqrt{4 - \epsilon_c^2}}$ and ($\hat{p}^Tx \leq t_1$ or $\hat{p}^Tx \geq t_2$)
    \item If $\delta < 1 $ and $\hat{p}^Tx \geq t_2$
\end{enumerate}
where $\hat{p} = p/\norm{p}$, thresholds $t_1, t_2$  are in $[-1, 1]$, $t_1 \leq t_2$ and $t_1, t_2$ are the solutions of the quadratic equation:
$$ t^2 + 2\delta \epsilon't + (\epsilon' (\delta^2+1)-1) = 0 $$
where $\epsilon' = \epsilon_c^2 - \frac{\epsilon_c^4}{4}$
($\epsilon_c$ defined in Theorem 1).
\end{theorem}
\textit{Proof sketch:} Similar to Theorem 1, except that now we place a bound on $\norm{p}$. The bound is then used to compute allowed values of $\hat{p}^Tx$ using Lemma 4. 

With Theorem 1, we have established the existence of additive UAPs which can cause a quantum classifier to classify all samples as belonging to a target class. Further, Theorem 2 establishes a connection between the perturbation strength and the subset of input space which gets classified as belonging to a target class. Theorem 2 implies that for any given $\delta$, there is a subset of vectors in $\mathbb{R}^d$ which get classified as belonging to class $c$.

Before moving forward, we would like to emphasize that the statements of both Theorem 1 and Theorem 2 involve sufficient conditions and not necessary conditions; it is possible that other perturbations exist which are ``better'' than the ones which satisfy conditions in Theorem 1 or 2. The objective of the developed theoretical framework is to prove the existence of additive UAPs for quantum classifiers at various perturbation strengths; no claim is made with respect to the optimality of the perturbations. A strategy for generating effective additive UAPs is discussed in the next section.

 \subsection{Generative Framework}
As described above, it is not straightforward to determine the perturbation $p$, which causes the highest misclassification under a given norm constraint. The generation of such UAPs is further complicated by the fact that the effectiveness of a UAP also depends on how the samples are distributed and therefore on the dataset under consideration. Therefore, in this section, we introduce an effective framework for obtaining additive UAPs in a constrained-norm setting. 

\begin{figure*}[!htbp]
    \centering
    \includegraphics[width = 0.9\textwidth]{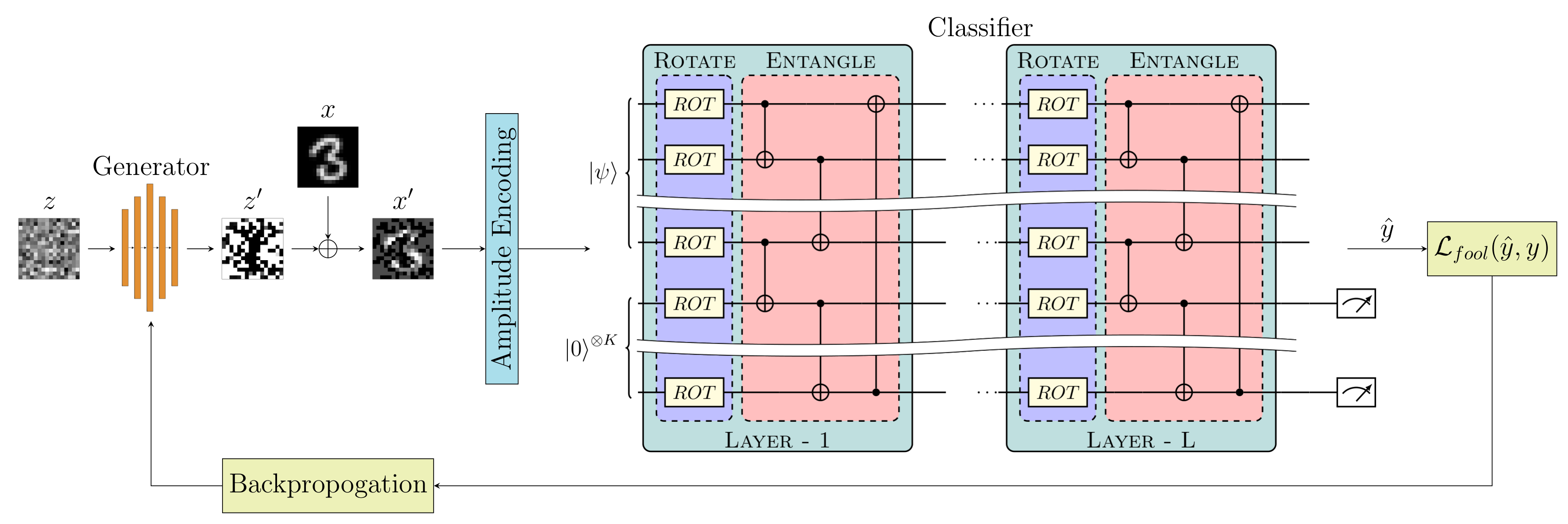}
    \caption{QuGAP-A: A framework for generating additive UAPs for quantum classifiers. A random vector $z$ sampled from $\mathbb{R}^m$ is passed through a classical generative network. The generated perturbation $z'$ is then scaled to impose the norm constraint and then added to an input sample $x$. The perturbed input sample $x'$ is then amplitude-encoded and passed through the trained quantum classifier $\mathcal{Q}$. Output predictions from $\mathcal{Q}$ are then used to compute the fooling loss $\mathcal{L}_{fool}$. Gradients computed are backpropagated to update the generator parameters. The process is repeated for all input samples over multiple epochs. }
    \label{fig:additive}
\end{figure*}

The proposed strategy for generating additive UAPs, denoted henceforth as QuGAP-A, for an amplitude-encoded classical dataset is illustrated in Figure \ref{fig:additive}. A brief description of the training procedure is given in the caption.

The key idea is to train the classical generator $\mathcal{G}$ to convert a given random vector $z$ to an additive UAP $z'$ for the dataset under consideration. Note that the generated UAP $z'$ is scaled to ensure that the $L_p$ norm is below a fixed threshold. The training is done by performing backpropagation and updating the parameters of $\mathcal{G}$ using the fooling loss $\mathcal{L}_{fool}$.  Typically, $p = 2$ or $p = \infty$ is used. The gradient computation can either be done completely classically by simulating $\mathcal{Q}$ (straightforward once the parameters of $\mathcal{Q}$ are known), or it can be done in a hybrid fashion where gradients from $\mathcal{Q}$ can be directly computed using the parameter-shift method. Additional details such as the structure of $\mathcal{G}$, hyperparameters for training and software packages used as well as experiments for targeted attacks are detailed in the supplementary.

Once the generator $\mathcal{G}$ training is complete, we no longer require access to the quantum classifier to generate adversarial samples. This makes our attack a {semi-whitebox} attack \cite{xiao2019generating}. The noise $z$ used during training is passed to the trained generator to generate the attack. The generated perturbation is added to a clean input to generate an adversarial sample.

\begin{figure}[h!tbp]
\centering
\begin{subfigure}{0.45\columnwidth}
\centering
\includegraphics[width=0.95\textwidth]{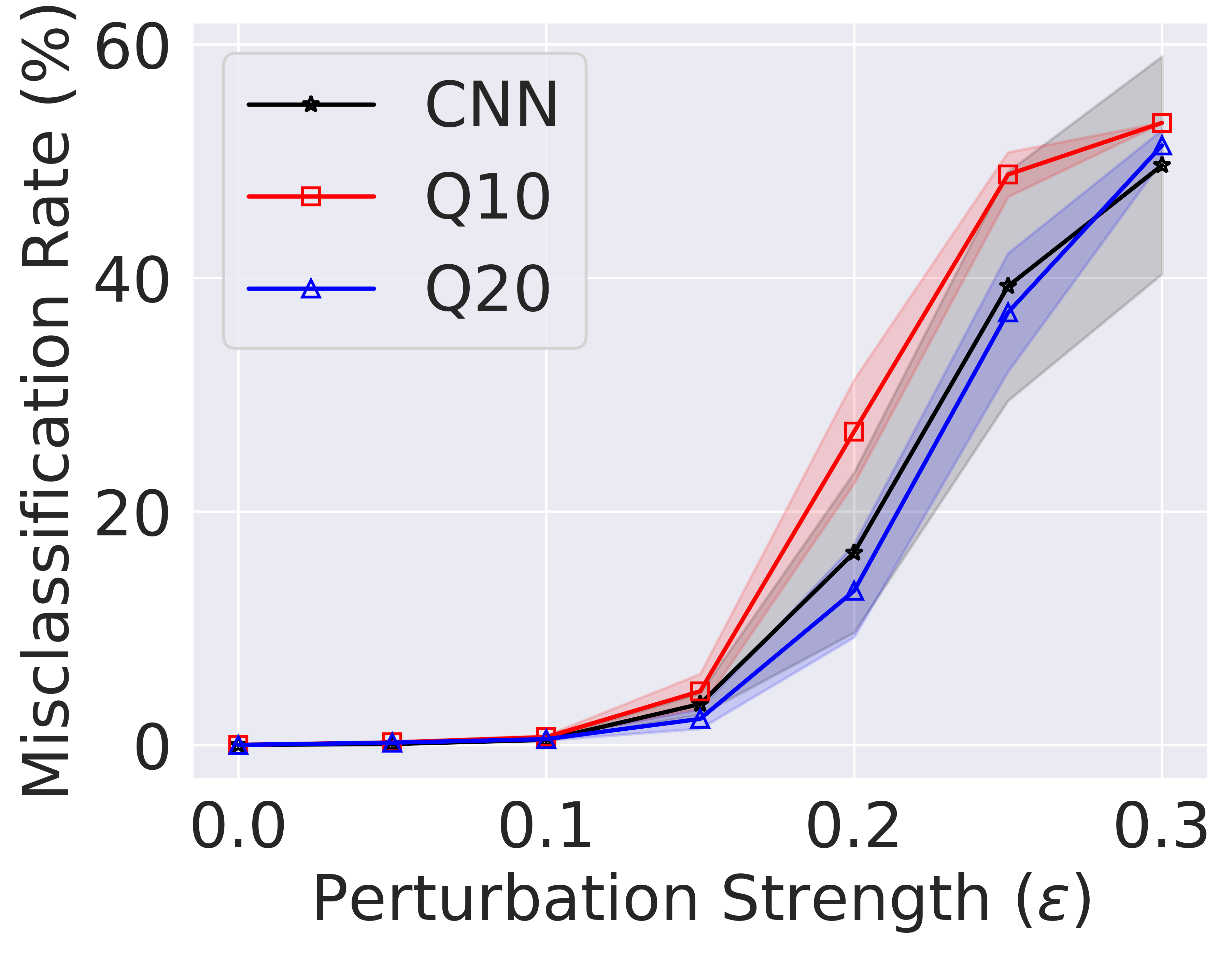}
\caption{MNIST: 2 Class}
\end{subfigure}
\begin{subfigure}{0.45\columnwidth}
\centering
\includegraphics[width=0.95\textwidth]{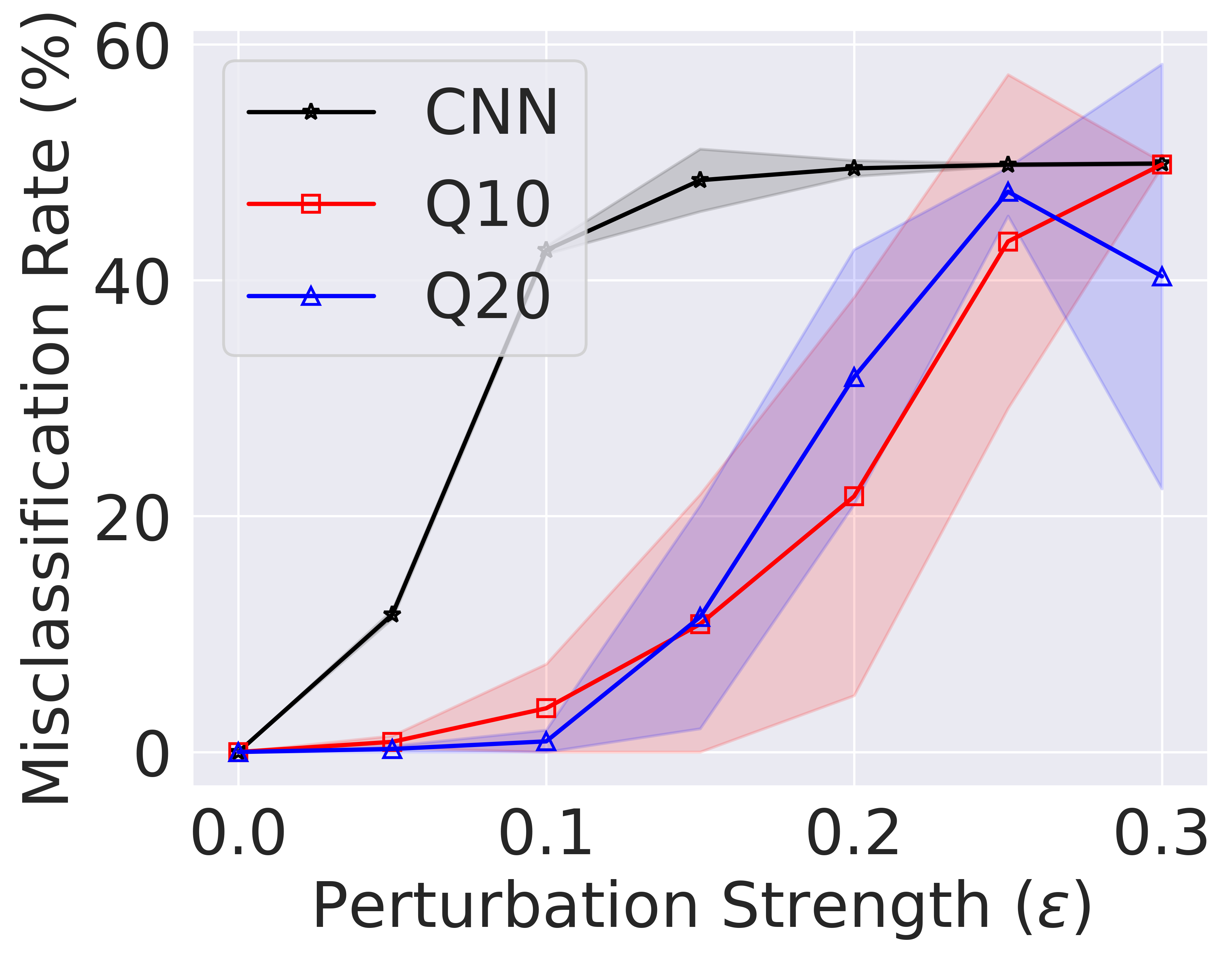}
\caption{FMNIST: 2 Class}
\end{subfigure}
\begin{subfigure}{0.45\columnwidth}
\centering
\includegraphics[width=0.95\textwidth]{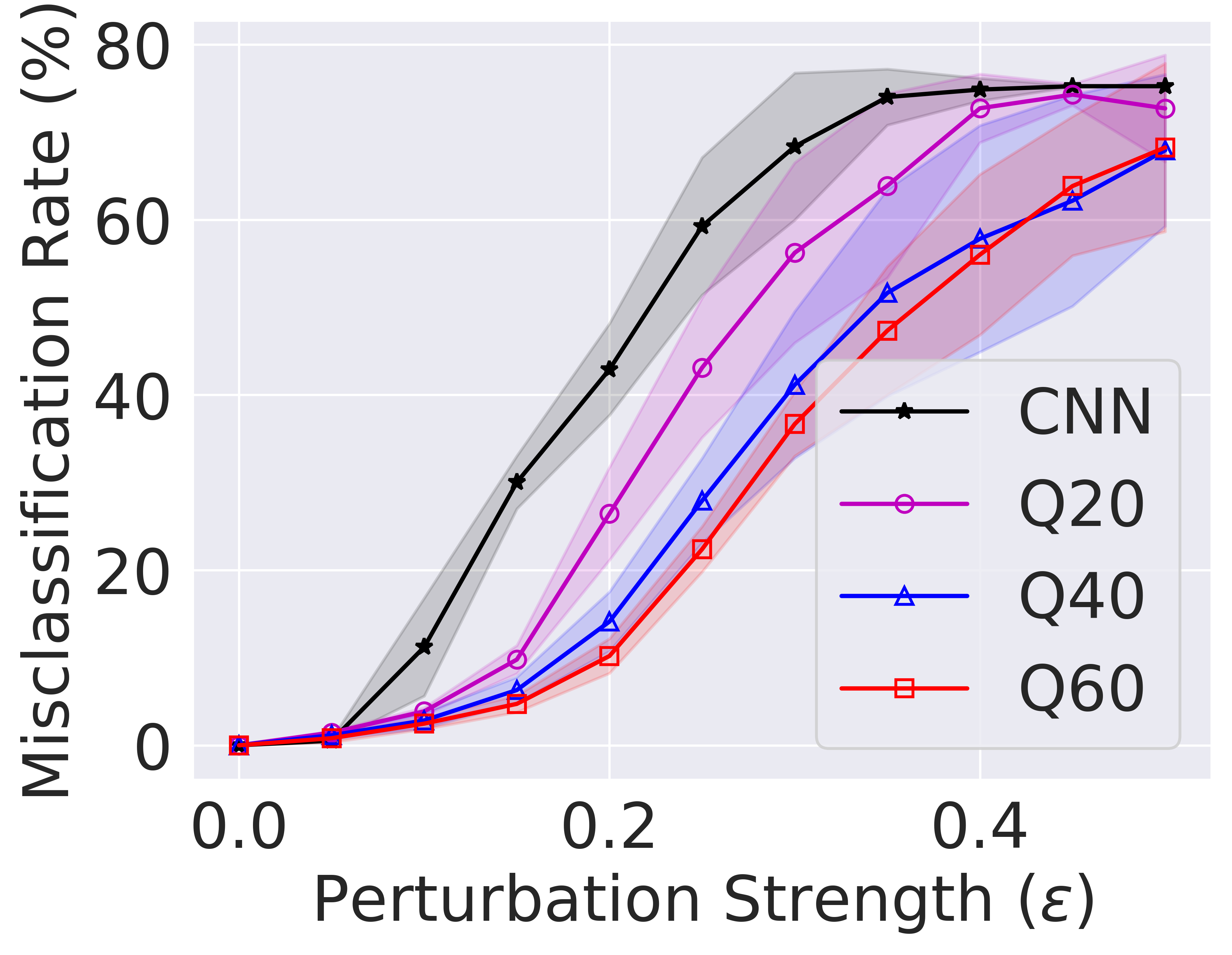}
\caption{MNIST: 4 Class}
\end{subfigure}
\begin{subfigure}{0.45\columnwidth}
\centering
\includegraphics[width=0.95\textwidth]{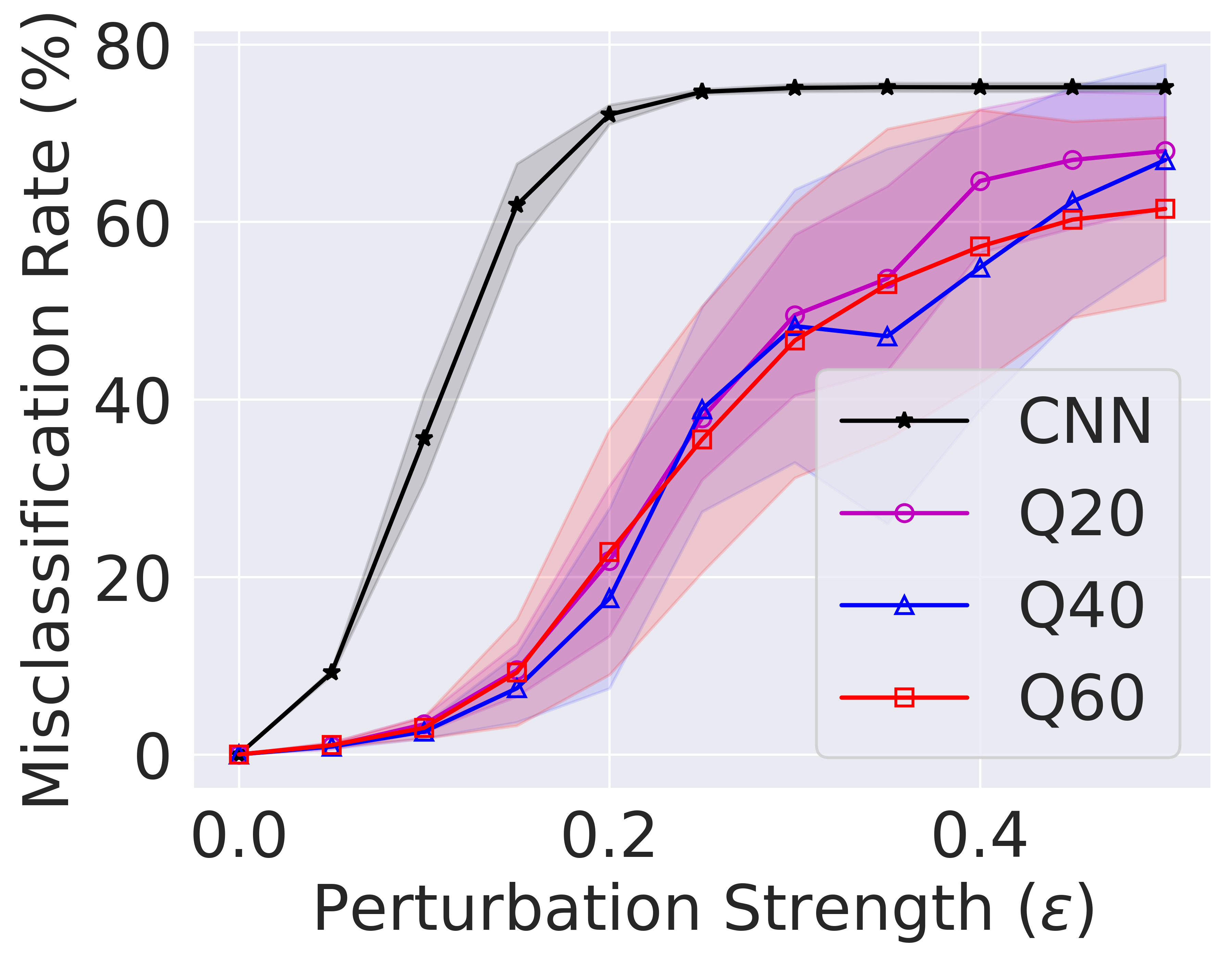}
\caption{FMNIST: 4 Class}
\end{subfigure}
\caption{The misclassification rates for $16\times16$ MNIST and FMNIST using additive untargeted UAPs. We report results for binary classification between classes $0$ and $1$ and 4-class classification between classes $0,1,2$ and $3$.}
\label{fig:expadditive}
\end{figure}
 
\subsection{Experimental Results}
 We test the generative framework by attacking quantum classifiers of different depths trained on two tasks: binary classification and four-class classification, and two datasets: MNIST \cite{lecun2010mnist} and FMNIST \cite{xiao2017fashion}. We downsample both datasets to $16\times16$ pixels due to computational limitations. We constrain the $L_{\infty}$ norm of the attack to be less than $\epsilon$. UAPs are generated with different values of the bound $\epsilon$ for the classifier. The UAP generation is stochastic in nature as we sample a random vector $z$ initially. To enable a fair evaluation, we test each setting $10$ times to ensure reasonably small standard deviations. While dealing with images, an additional step of clipping the data after perturbation is required to ensure that the pixels are in the range $[0,1]$. The implications of this additional step on the UAP generation as well as more specifics regarding the experiments, such as the training procedure and hyperparameters used, are detailed in the supplementary.

 To illustrate the effectiveness of our framework, we plot the  misclassification rates (along with standard deviation) for the experiments in Figure \ref{fig:expadditive}. CNN represents a classical convolutional neural network, whereas Q10, Q20, Q40 and Q60 represent PQCs with depths 10, 20, 40 and 60, respectively. As expected, the misclassification rates increase with an increase in $\epsilon$. Note that the misclassification rate plateaus around $50\%$ for binary classification and around $75\%$ for 4-class classification. This is in line with the developed theory; for a $k-$class classification problem, assuming even distribution of samples across all classes and an ideal classifier, the misclassification rate when all samples are classified as one particular class will be $\frac{k-1}{k}$. A naive approach to generate additive UAPs for quantum classifiers would be to generate perturbations for a classical classifier and transfer them to quantum classifiers. However, empirically such attacks do not transfer well. Transferability studies are presented in the supplementary material.
 
\begin{figure*}[!htbp]
    \centering
    \includegraphics[width=0.9\textwidth]{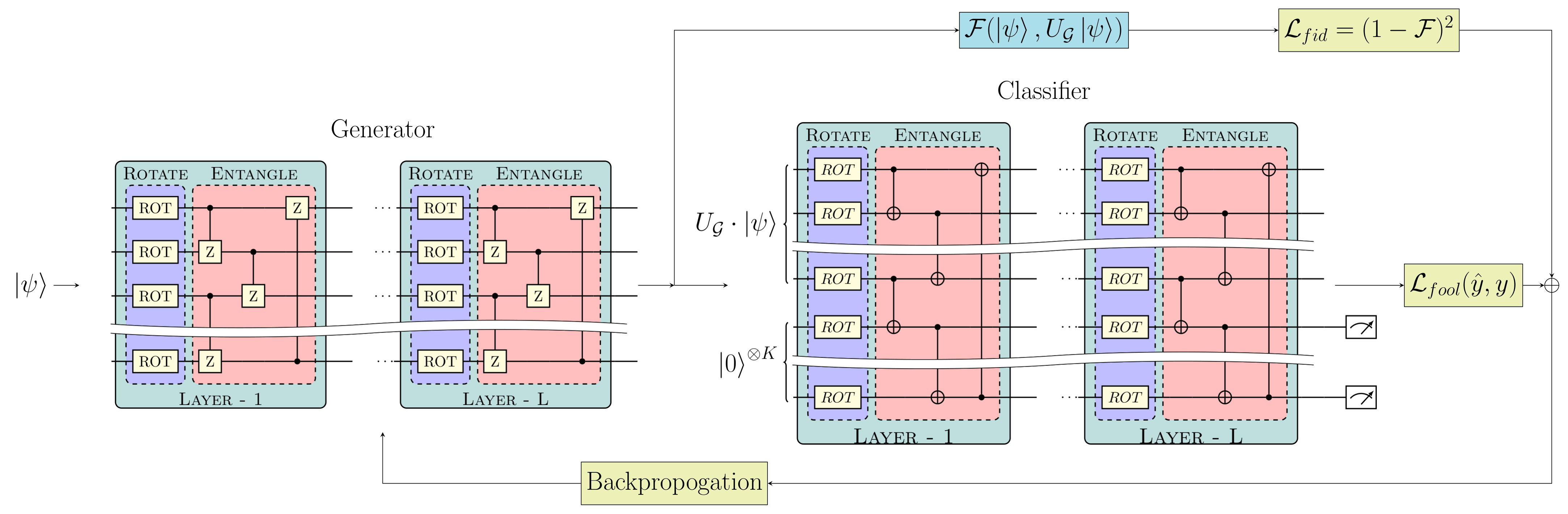}
    \caption{QuGAP-U: A framework for generating unitary UAPs for quantum classifiers. The quantum generator $\mathcal{G_Q}$ takes in an input state $\ket{\psi_{i}}$ and transforms it into a perturbed state $\ket{\phi_{i}} = U_\mathcal{G}\ket{\psi_{i}}$. The fidelity between $\ket{\psi_{i}}$  and $\ket{\phi_{i}}$ is computed from which $\mathcal{L}_{fid}$ is calculated. $\ket{\phi_{i}}$ is also passed through a trained quantum classifier $\mathcal{Q}$ to compute $\mathcal{L}_{fool}$. Gradients are computed using the total loss $\mathcal{L}_{fool} + \alpha \mathcal{L}_{fid}$ and used to update the parameters of $\mathcal{G_Q}$ over all training samples for multiple epochs}
\label{fig:unitary}
\end{figure*}

\section{Unitary UAPs}

In this section, we propose a strategy for generating unitary transformations in the quantum domain which can act as UAPs. In contrast to the previous section where additive perturbations are added to classical data before encoding, in this section we focus on the case where adversarial manipulations directly transform the quantum state. This enables us to attack encoded classical data as well as quantum data. 
\subsection{Motivation}

The existence of quantum UAPs has been theoretically established in \cite{gong2022universal}; furthermore they employ the iterative qBIM \cite{lu2020quantum} algorithm to generate quantum UAPs. However, a drawback of the algorithm is that the search space for the global unitary ${U}$ is limited as $U$ is constrained to be a product of local unitaries near the identity matrix. This limitation is further worsened by the fact that only a single variational layer is used to generate the UAP.

We propose a novel strategy to overcome these limitations; instead  of constraining $U$ to be a product of local unitaries near identity, we implement a fidelity-based loss function to control the perturbation strength. Further, a PQC-based generative network is used instead of a single variational layer to search over a larger space of unitaries.

\subsection{Proposed Framework}
The proposed framework for generating unitary UAPs, denoted as QuGAP-U, is illustrated in Figure \ref{fig:unitary}. A brief description of the training procedure is given in the caption. 

We introduce a novel loss function $\mathcal{L}_U$ of the form:
$$\mathcal{L}_U = \mathcal{L}_{fool} + \alpha \mathcal{L}_{fid}$$
where $\mathcal{L}_{fool}$ is the fooling loss and $\mathcal{L}_{fid}$ is a fidelity-based loss of the form: 
$$\mathcal{L}_{fid} = (1-\mathcal{F}(\ket{\psi}, \ket{\phi}))^2$$
$\mathcal{F}(\ket{\psi}, \ket{\phi}) = \abs{\braket{\psi}{\phi}}^2$ is the fidelity between the input state and the perturbed state. $\alpha$ is a hyper-parameter controlling the trade-off between misclassification and fidelity; a higher value of $\alpha$ generates UAPs which ensure a higher fidelity of perturbed states but may have lower misclassification rates. Some details regarding implementation on quantum hardware: since $\ket{\psi_{i}}$ and $\ket{\phi_{i}}$ are pure states, the fidelity can be computed exactly using the SWAP test \cite{stein2021qugan}. The gradients can also be computed using parameter-shift rule \cite{schuld2019evaluating, mitarai2018quantum}. 

\subsection{Classical Simulation}

To empirically verify the viability of the proposed framework, we simulate it classically by optimizing for a unitary $U_\mathcal{G_Q}$ which acts as a proxy for the quantum generator $\mathcal{G_Q}$. The objective then is to learn $U_\mathcal{G_Q}$ which minimizes the loss $\mathcal{L}_\mathcal{Q}$. To perform the optimization, we take inspiration from \cite{kiani2022projunn} and project the matrix learned after each step of gradient descent into the space of unitary matrices. The simulation is done on two datasets: MNIST and Transverse-field Ising Model (TIM) Dataset. The synthetic TIM dataset maps the states of the transverse-field Ising model described in \cite{PFEUTY197079} to the phase of the system (ferromagnetic or paramagnetic). We model this physical system as a binary classification task on pure quantum data. More details regarding the TIM dataset as well as the complete pseudocode for classical optimization are given in the supplementary. 

\begin{figure}[!htbp]
 \centering
 \begin{subfigure}{0.45\columnwidth}
\centering
\includegraphics[width=0.95\textwidth]{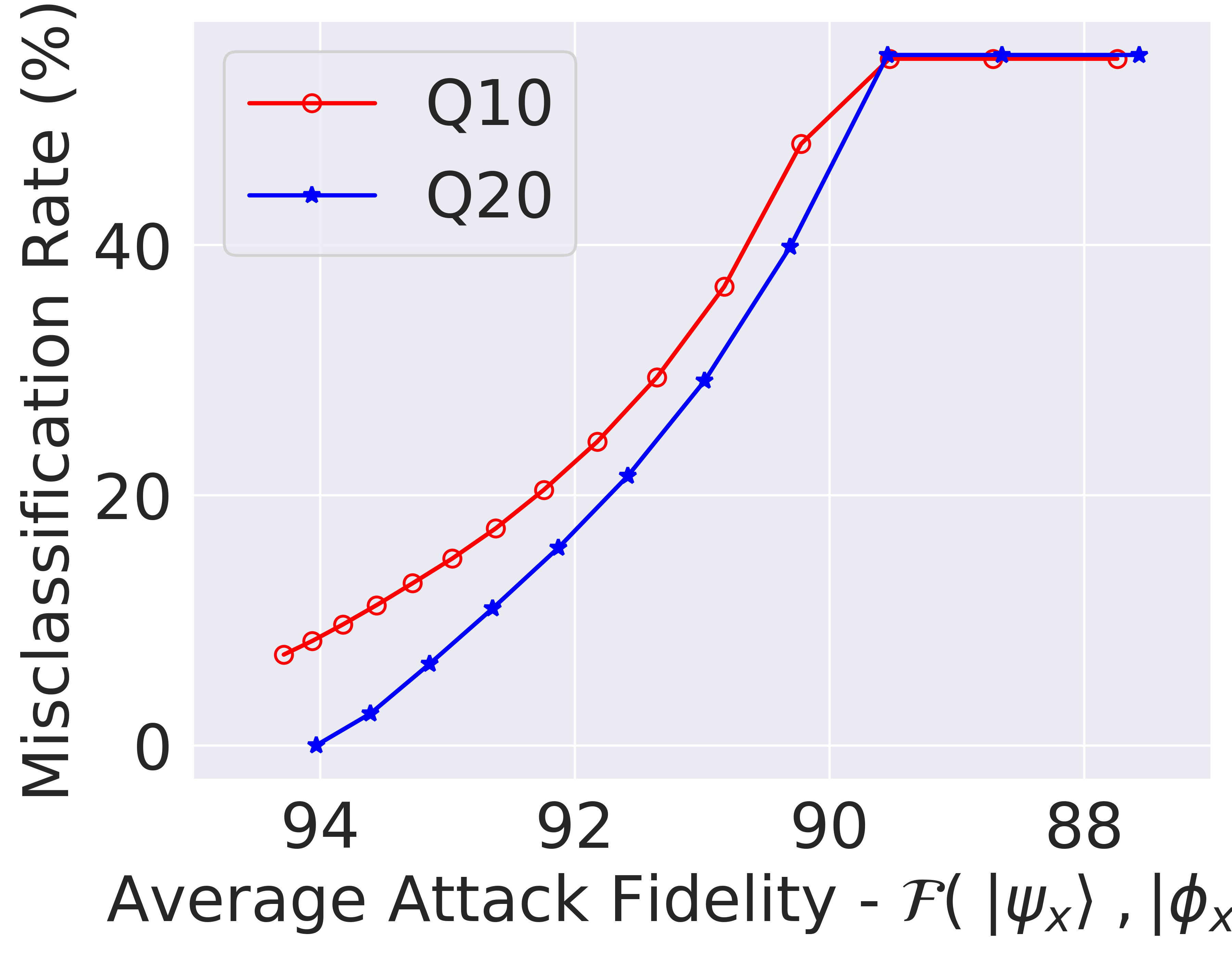}
\caption{TIM: 2 Class}
\end{subfigure}
\begin{subfigure}{0.45\columnwidth}
\centering
\includegraphics[width=0.95\textwidth]{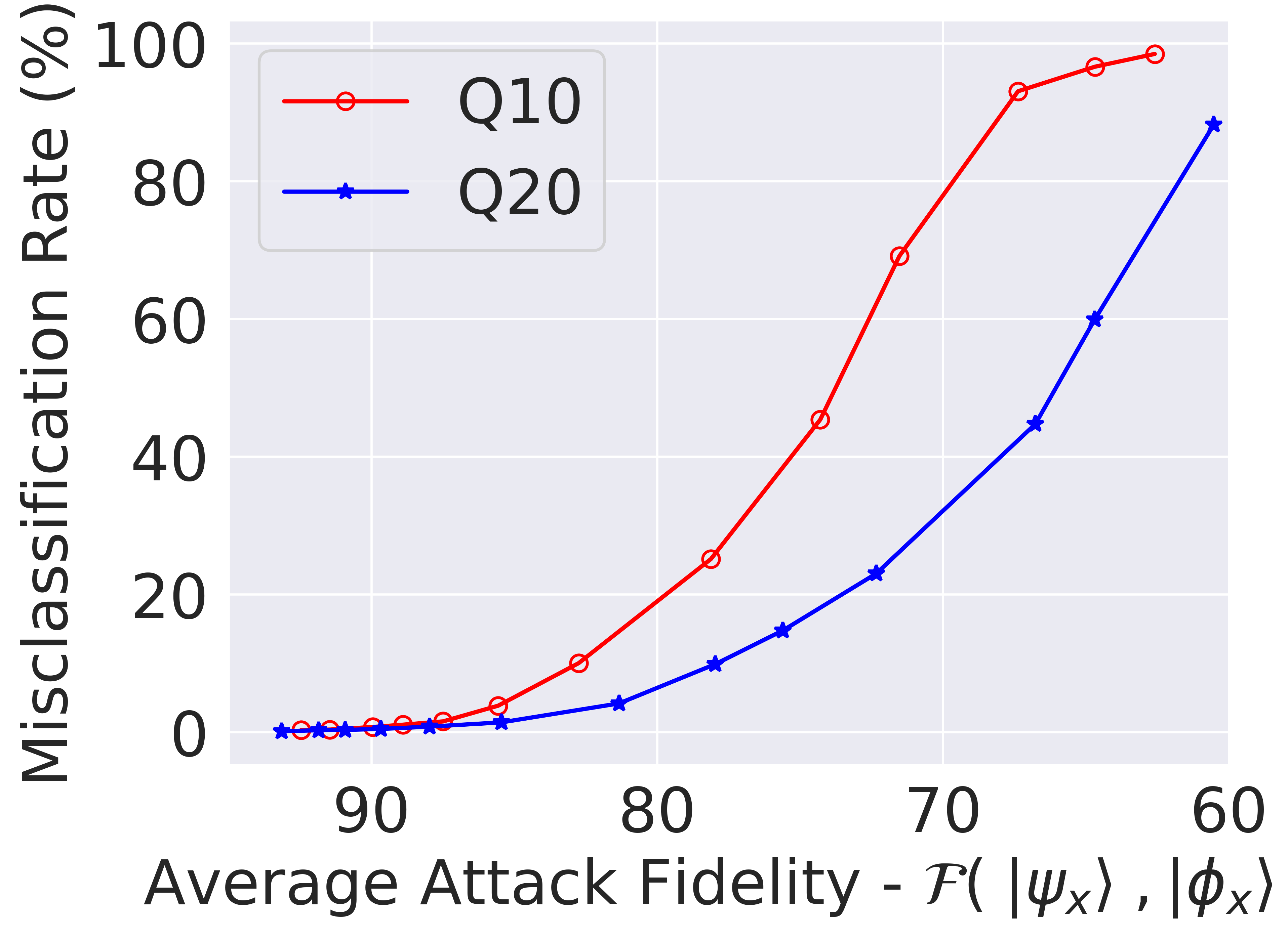}
\caption{MNIST: 2 Class}
\end{subfigure}

\caption{Classical simulation of a framework for generating unitary UAPs. Misclassification rates for TIM binary classification are given in (a), and for 8x8 downsampled MNIST binary classification are given in (b). Achieving competitive misclassification rates for MNIST results in much lower attack fidelities.}
\label{fig:class_sim} 
 
 \end{figure}
The results are illustrated in Figure \ref{fig:class_sim}. The plots validate the hypothesis that by varying the value of $\alpha$, we can achieve fine-grained control over the fidelity of the perturbed states produced by the learned UAP. We also note that while the misclassification rate approaches $100\%$ for MNIST  classification as you decrease $\alpha$, misclassification rate for TIM classification stays at around $54\%$. An explanation for this disparity is given in \cite{gong2022universal}; the maximum allowed misclassification rate has an upper bound which depends on the dataset distribution.

\subsection{PQC Simulation}
In practice, quantum UAPs are implemented using PQCs. While we have demonstrated the effectiveness of the proposed approach through classical simulation in the previous section, it must be noted that PQCs, by construction, have access to only local unitaries. In such a scenario, it has been shown that for constructing an arbitrary unitary in a Hilbert space of dimension $d$, one would require $\mathcal{O}(d^2)$ gates \cite{shende2005synthesis}. Therefore, we expect the depth of the PQC to have a significant impact in determining the quality of the generated UAP. 

\begin{table}[!htbp]
    \centering
    \begin{tabular}{ c | c | c | c| c }
    \hline \hline
     Fidelity  &  \multicolumn{2}{c|}{TIM} &  \multicolumn{2}{c}{MNIST}\\
     \cline{2-5} 
     constraint & qBIM & QuGAP-U & qBIM & QuGAP-U \\
     \hline 
     \hline
      0.95 &  3.18 & $\mathbf{10.54}$ & 0.00& {0.00} \\
      \hline
      0.90 &6.37 & $\mathbf{54.88}$  & 0.00& {0.00}\\
      \hline
      0.85 & 9.44& $\mathbf{54.88}$  & 0.00& \textbf{12.56}\\
      \hline
      0.80 &13.39 & $\mathbf{54.88}$  & 0.00& \textbf{36.65}\\
      \hline
      0.75 & 16.68& $\mathbf{54.88}$ & 0.05& \textbf{50.78}\\
      \hline
      0.70 &23.27 &$\mathbf{54.88}$  & 0.15&\textbf{54.97}\\
      \hline \hline
    \end{tabular}
    \caption{Comparison of misclassification rates for a PQC classifier of depth $10$ trained on TIM and MNIST datasets after qBIM and QuGAP attacks. Fidelity constraint is the minimum average fidelity to be maintained while attacking. }
    \label{tab:unitary_uap}
\end{table}

We repeat the classical simulation experiments with a PQC-based quantum generative model, implemented using the Pennylane library \cite{bergholm2022pennylane}. For benchmarking performance, we compare the performance of our method with the qBIM-based approach proposed in \cite{gong2022universal}. Tests are run for binary classification on the TIM dataset and an $8\times8$ downsampled version of MNIST. QuGAP-U uses a quantum generator with $30$ layers for the TIM dataset and $200$ layers for the MNIST dataset. The number of parameters in the generative models is chosen to be around $d^2$ in each case. Here $d$ denotes the Hilbert space dimension or equivalently the number of input features. The effect of varying generator depth on the performance of QuGAP-U is detailed in the supplementary. Note that qBIM can utilize only a single variational layer; deeper circuits will significantly degrade the fidelity of the generated samples as the method does not explicitly rely on fidelity constraints. The benchmarking results are presented in Table \ref{tab:unitary_uap}. We observe that the performance of qBIM is much poorer than QuGAP-U, possibly because QuGAP-U utilizes deeper quantum generators and hence has a significantly larger search space. A more detailed analysis, as well as further discussion, can be found in the supplementary.
\begin{figure}[!tbp]
 \centering
    \begin{subfigure}{0.45\columnwidth}
    \centering
     \includegraphics[width=0.95\textwidth]{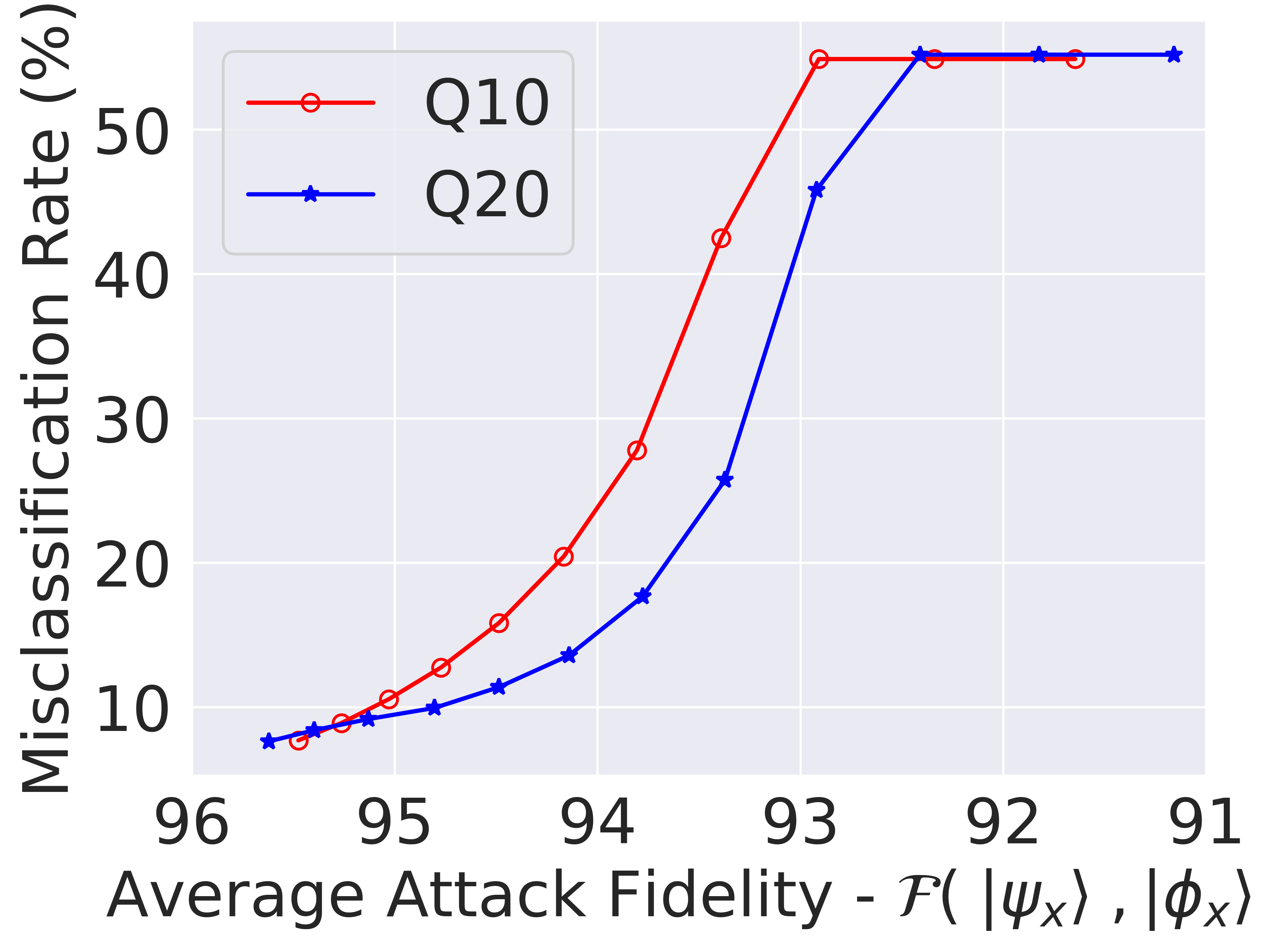}
     \caption{TIM: 2 class}
    \end{subfigure}
    \begin{subfigure}{0.45\columnwidth}
    \centering
     \includegraphics[width=0.95\textwidth]{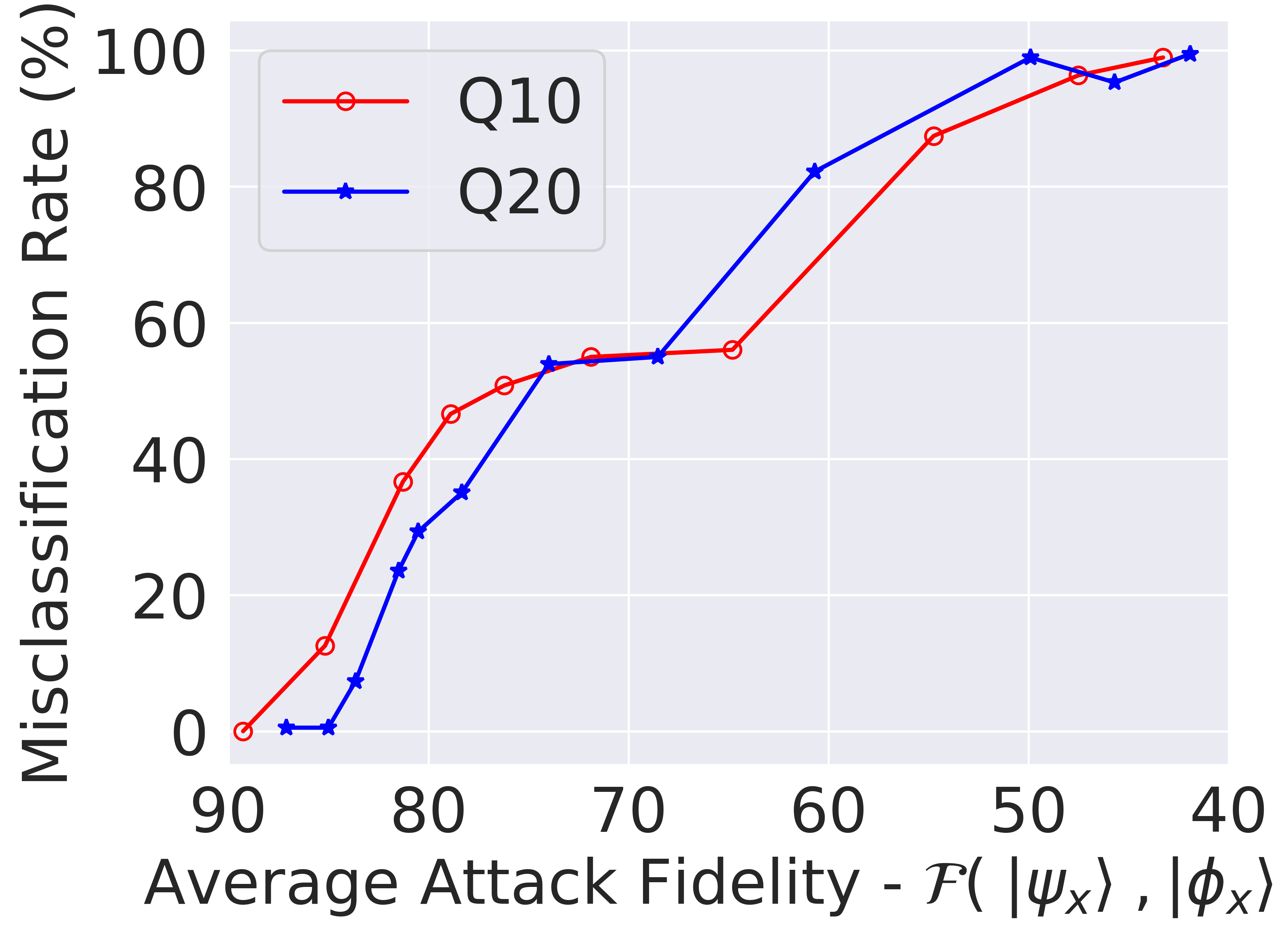}
     \caption{MNIST: 2 class}
    \end{subfigure}
     \caption{Misclassification rate evolution of QuGAP-U on classifiers with depths 10(Q10) and 20(Q20). (a) TIM dataset; (b) $8\times 8$ downsampled MNIST dataset}
     \label{uap_unit}
 \end{figure}

QuGAP-U clearly outperforms qBIM, the only other existing method for generating unitary UAPs, on both tasks, thereby achieving state-of-the-art performance in unitary UAP generation. We also observe that for the TIM dataset, misclassification saturates at 90\% fidelity. This is because the misclassification rate rises drastically in the $90\%$ - $95\%$ fidelity range, as evidenced in Figure \ref{uap_unit} (a). We also note that the performance closely matches the classical simulation (Figure \ref{fig:class_sim} (a)) for the TIM dataset. We plot the performance of QuGAP-U on MNIST in Figure \ref{uap_unit} (b). While TIM can be attacked with a quantum generator depth of just 30, a generator depth of around 200 is required to achieve good results for MNIST. Even then, the attacks are weaker than the classical simulation (Figure \ref{fig:class_sim} (b)). This observation further supports our hypothesis that higher dimensional datasets require deeper PQCs for generating effective unitary attacks.

\section{Discussion and Conclusion}
With the rising prominence of quantum classifiers, it is necessary to study and mitigate the effect of adversarial attacks on such classifiers. In this work, we have analyzed universal adversarial perturbations in the context of quantum classifiers. We theoretically proved existence of additive UAPs and proposed a framework for UAP generation. We then established a novel framework for generating unitary UAPs and empirically demonstrated its advantages over existing methods. 

A natural extension to our work would be to examine the effect of additive perturbations on encoding schemes other than amplitude encoding. Moreover, while we have designed the framework with considerations for practical implementations, our experiments are limited to simulations. Implementation of the proposed schemes on actual quantum computers may also be a worthwhile avenue for future research. Finally, it might be interesting to analyze the impact of quantum noise on the performance of the generated attacks.

\section*{Ethics Statement}
In this work, we introduce algorithms to generate Universal Adversarial Perturbations (UAPs) for quantum classifiers which act on amplitude-encoded classical, and quantum datasets. Using the methods we propose, adversaries may tailor adversarial attacks which show reasonable efficiency even on real-world noisy quantum classifiers. However, due to the limited presence of PQC-based quantum classifiers in real-world applications at the time of publication of this work, we believe that our work does not pose any immediate security threats. On the other hand, the theoretical and empirical demonstrations of the efficacy of such attacks in our research, pre-emptively highlights the need to develop effective defense strategies capable of mitigating such attacks. We also concretely conceptualize the two types of UAPs for quantum classifiers: Additive UAPs (generated using QuGAP-A) and Unitary UAPs (generated using QuGAP-U); allowing future work in developing defense strategies to be broadly centered around these areas. To the best of our knowledge our work does not raise any ethical concerns, other than those addressed above.

\section*{Acknowledgements}
The research presented in this work was done partially while the authors were at the Intelligent Data Science Lab, University of British Columbia. We thank the anonymous reviewers for their feedback. We also thank the Digital Research Alliance of Canada for access to computational resources. This work was supported by the MITACS Globalink Research Internship award 2022 and by the NSERC Discovery Grant No. RGPIN-2019-05163. All source code used for this research may be found at: https://github.com/Idsl-group/QuGAP along with links to the supplementary material.

\section*{Technical Appendix}
\appendix
\section{Theoretical proofs}

In this section, we provide complete proofs for all lemmas and theorems stated in the main paper. We reiterate that all vectors and matrices are zero-indexed. Furthermore, a vector is represented by a column matrix.

\subsection{Lemma 1}

\subsubsection{Statement:} {The probability of an input $x$ being classified as class $c$ is given by $P(\hat{c}_x = c) = x^\dagger M^c x$ where, $x^\dagger$ denotes the conjugate transpose of $x$ and $M^c\in\mathbb{C}^{d\times d}$ is a positive semi-definite matrix given by: $M^{c}_{ij} = \sum_{t=0}^{d-1} \;U_{k't+c,k'i}^* \; U_{k't+c,k'j}$ with $^*$ denoting the complex conjugate and $k' = 2^K$.}
\subsubsection{Proof:} Recall that according to Assumption 4:
\begin{equation*}
    P(\hat{c}_x = c) = |\braket{\mathbf{1}_{2^D} \otimes {c}} {{y}}|^2
\end{equation*}
where $\ket{c}$ is the $c^{\textrm{th}}$ standard basis state in $\mathbb{R}^{2^K}$. Let $P = P(\hat{c}_x = c)$. Expanding out the inner product, we have:
\begin{equation}
    \label{eq:prob}
    P = \sum_{t = 0}^{2^{D}-1}|y_{2^K \cdot t + c}|^2
\end{equation}
where $y_j$ denotes the $j^\textrm{th}$ element of $\ket{y}$. Note that:
\begin{equation*}
    \ket{y} = U({\ket{x} \otimes \ket{0}^{\otimes K}})
\end{equation*}
which implies:
\begin{equation}
\label{eq:yj}
    y_j = \sum_{s = 0}^{2^{D}-1}U_{j, 2^K \cdot s}x_{s}
\end{equation}
where $U_{a, b}$ denotes the element at position $(a, b)$ of $U$ and $x_a$ denotes the $a^{\textrm{th}}$ element of $x$. Substituting \eqref{eq:yj} in \eqref{eq:prob} yields:

\begin{align}
    P &= \sum_{t = 0}^{2^{D}-1}\;\abs{ \sum_{s = 0}^{2^{D}-1}(U_{2^Kt+c, 2^K  s })x_{s} }^2 \nonumber \\
    &= \sum_{t = 0}^{2^{D}-1}\;\Big ( \sum_{s = 0}^{2^{D}-1}U^*_{2^Kt+c, 2^K s}x^*_{s} \Big)\Big ( \sum_{q = 0}^{2^{D}-1}U_{2^Kt+c, 2^K \cdot q }x_{q} \Big) \nonumber \\
    &= \sum_{s = 0}^{2^{D}-1} \sum_{q = 0}^{2^{D}-1} x^*_{s} \Big ( \sum_{t = 0}^{2^{D}-1} U^*_{2^Kt+c, 2^K \cdot s} U_{2^Kt+c, 2^K \cdot q }  \Big )  x_{q} 
    \label{eq:big_p}
\end{align}
We define:
\begin{equation}
    M_{s, q}^c = \sum_{t = 0}^{2^{D}-1} U^*_{2^Kt+c, k's} U_{2^Kt+c, k'q }
    \label{eq:mc}
\end{equation}
where $k' = 2^K$. Then we can rewrite \eqref{eq:big_p} as:
\begin{equation}
    P = \sum_{s = 0}^{2^{D}-1} \sum_{q = 0}^{2^{D}-1} x^*_{s} M_{s, q} x_{q}
    \label{eq:pf}
\end{equation}
Rewriting \eqref{eq:pf} in terms of matrices, we have:
\begin{equation*}
    P = x^{\dagger}M^cx
 \end{equation*}
where $M_c$ is defined by \eqref{eq:mc}. Note that since $P\geq0$ (P is a probability), $x^{\dagger}M^cx \geq 0$ for all $x$ and hence $M^c$ is a positive semi-definite matrix.

\subsection{Lemma 2}

\subsubsection{Statement:} Let $x,y\in \mathbb{R}^{d}$ such that $\norm{x} = 1$ and $\norm{y} = 1$. If  $\norm{x-y}\leq \epsilon$, where $\epsilon\in\mathbb{R}$, the probability of the point $y$ being classified as $c$ is bounded as
$\abs{P(\hat{c}_{y} = c) - P(\hat{c}_{x} = c)} \leq d\cdot\big(\epsilon^2 + 2\epsilon\big)$.

\subsubsection{Proof:} Let us write:
\begin{equation*}
    y = x + \delta 
\end{equation*}
for some $\delta \in \mathbb{R}^d$. Then, from Lemma 1 we have:
\begin{align*}
    P(\hat{c}_{y} = c) &= y^{\dagger}M^cy \\
    &= (x + \delta)^{\dagger}M^c(x + \delta) \\
    &= x^\dagger M^cx + x^\dagger M^c\delta + \delta^\dagger M^cx + \delta^\dagger M^c\delta
\end{align*}
Noting that $ x^\dagger M^cx  = P(\hat{c}_{x} = c)$, we have:
\begin{equation*}
    P(\hat{c}_{y} = c) - P(\hat{c}_{x} = c) = x^\dagger M^c\delta + \delta^\dagger M^cx + \delta^\dagger M^c\delta
\end{equation*}
By using triangle inequality, we have:
\begin{equation}
    \label{eq:pyx}
    \abs{P(\hat{c}_{y} = c) - P(\hat{c}_{x} = c)} \leq \abs{x^\dagger M^c\delta} + \abs{\delta^\dagger M^cx} + \abs{\delta^\dagger M^c\delta}
\end{equation}
Recognizing the three terms on the RHS as inner products, we apply Cauchy-Schwartz inequality to obtain:
\begin{align}
    \label{eq:xd}\abs{x^\dagger M^c\delta} &\leq \norm{x}\norm{M^c\delta} \\
    \label{eq:dx}\abs{\delta^\dagger M^cx} &\leq \norm{\delta}\norm{M^cx} \\
    \label{eq:dd}\abs{\delta^\dagger M^c\delta} &\leq \norm{\delta}\norm{M^c\delta} 
\end{align}
Substituting \eqref{eq:xd}, \eqref{eq:dx}, \eqref{eq:dd} back in \eqref{eq:pyx}, we have:
\begin{align}
    \abs{P(\hat{c}_{y} = c) - P(\hat{c}_{x} = c)} &\leq \norm{x}\norm{M^c\delta}\\ &+ \norm{\delta}\norm{M^cx}
    + \norm{\delta}\norm{M^c\delta} \label{eq:sar}
\end{align}
Note that $M^c$ is a PSD matrix and therefore has a spectral decomposition of the form:
\begin{equation*}
    M^c = Q \Lambda Q^{\dagger}  
\end{equation*}
where $Q \in \mathbb{C}^{d\times d}$ is a unitary matrix and $\Lambda \in \mathbb{C}^{d\times d}$ is a diagonal matrix with $\Lambda_{i, i} = \lambda_{i}$ where $\lambda_{i} \in \mathbb{R}$ is the $i^{\textrm{th}}$ eigenvalue of $M^c$. Note that $\lambda_{i}$ is guaranteed to be real since $M^c$ is PSD and therefore Hermitian. Furthermore, since $Q$ is unitary, the columns of $Q$ form an orthonormal basis in $\mathbb{C}^{d}$. Therefore, any vector $z \in \mathbb{C}^{d}$ can be written as:
\begin{equation*}
    z = Qr
\end{equation*}
for some $r \in \mathbb{C}^{d}$. Therefore we can write:
\begin{align*}
    x &= Qr^{(1)} \\
    \delta &= Qr^{(2)}
\end{align*}
for some $r^{(1)}, r^{(2)} \in \mathbb{C}^{d}$. Using the spectral decomposition of $M^c$, we can write:
\begin{align}
    \norm{M^cx} &= \norm{ Q \Lambda Q^{\dagger}x} \nonumber\\
    &= \norm{ Q \Lambda Q^{\dagger} Q r^{(1)}} \nonumber \\
    &= \norm{\Lambda r^{(1)}} \label{eq:mclambda}
\end{align}
using the fact that $Q$ is unitary. Similarly, we have:
\begin{equation*}
    \norm{M^c \delta} = \norm{\Lambda r^{(2)}}
\end{equation*}
Using the fact that $\Lambda$ is diagonal, \eqref{eq:mclambda} can be rewritten as:
\begin{equation}
    \norm{M^cx} = \sqrt{\sum_{i = 0}^{d-1}\abs{\lambda_ir^{(1)}_i}^2}\label{eq:mcsqrt}
\end{equation}
Let $\lambda_m$ denote the largest eigenvalue of $M^c$. Then we have:
\begin{equation}
    \sum_{i = 0}^{d-1}\abs{\lambda_ir^{(1)}_i}^2 \leq \lambda_m^2\sum_{i = 0}^{d-1}\abs{r^{(1)}_i}^2 \label{eq:lambsum}
\end{equation}
Using \eqref{eq:lambsum}, \eqref{eq:mcsqrt} can be written as:
\begin{equation*}
    \norm{M^cx} \leq  \sqrt{\lambda_m^2\sum_{i = 0}^{d-1}\abs{r^{(1)}_i}^2}
\end{equation*}
Which can be simplified as:
\begin{equation}
    \norm{M^cx} \leq  \lambda_m\norm{r^{(1)}} \label{eq:pap}
\end{equation}
Note that since $x = Qr^{(1)}$ and $Q$ is unitary, $\norm{x} = \norm{r^{(1)}}$. Therefore, \eqref{eq:pap} can be rewritten as:
\begin{equation}
    \norm{M^cx} \leq  \lambda_m\norm{x} \label{eq:cle}
\end{equation}
Similarly,
\begin{equation}
    \norm{M^c \delta} \leq  \lambda_m\norm{\delta} \label{eq:abu}
\end{equation}
Substituting \eqref{eq:cle} and \eqref{eq:abu} in \eqref{eq:sar} we have:
\begin{align*}
    \abs{P(\hat{c}_{y} = c) - P(\hat{c}_{x} = c)} &\leq \norm{x} \lambda_m\norm{\delta}\\ &+ \norm{\delta}\lambda_m\norm{x}
    + \norm{\delta} \lambda_m\norm{\delta} 
\end{align*}
Now, since inputs are normalized we have $\norm{x} = 1$. Furthermore, it is given that $\norm{\delta} = \norm{y - x} \leq \epsilon$. Using these facts, we have:
\begin{align}
    \abs{P(\hat{c}_{y} = c) - P(\hat{c}_{x} = c)} &\leq  \lambda_m(\epsilon^2+2\epsilon) \label{eq:lameps}
\end{align} 
Now, since the sum of the eigenvalues of a matrix must equal its trace, we have:
\begin{equation}
    \sum_{i = 0}^{d-1}\lambda_i = \textrm{Tr}(M^c) \label{eq:lameps2}
\end{equation}
Since $M^c$ is PSD, all eigenvalues are non-negative. Therefore:
\begin{equation}
    \lambda_m \leq \sum_{i = 0}^{d-1}\lambda_i \label{eq:lameps3}
\end{equation}
From the definition of trace operator, we have:
\begin{align}
    \textrm{Tr}(M^c) &= \sum_{i = 1}^{d}M^c_{i, i} \nonumber\\
    &= \sum_{i = 0}^{d-1}   \sum_{t=0}^{d-1} \;U_{2t+c,k'i}^* \; U_{2t+c,k'i} \nonumber\\
    &= \sum_{i = 0}^{d-1}   \sum_{t=0}^{d-1} \abs{U_{2t+c,k'i}}^2 \nonumber\\
    &\leq \sum_{i = 0}^{d-1}   \sum_{t=0}^{2^{D+K}-1} \abs{U_{t,k'i}}^2 \label{eq:kili}
\end{align}
Note that since $U$ is a unitary matrix and therefore $\sum_{t=0}^{2^{D+K}-1} \abs{U_{t,k'i}}^2 = 1$ since columns are orthonormal. Therefore, we can rewrite \eqref{eq:kili} as:
\begin{equation}
    \textrm{Tr}(M^c) \leq d \label{eq:dtrace}
\end{equation}
Finally, using \eqref{eq:lameps}, \eqref{eq:lameps2}, \eqref{eq:lameps3} and \eqref{eq:dtrace} we obtain the desired result:
\begin{equation*}
    \abs{P(\hat{c}_{y} = c) - P(\hat{c}_{x} = c)} \leq  d (\epsilon^2+2\epsilon)
\end{equation*}

\subsection{Lemma 3}

\subsubsection{Statement:} For any $x\in \mathbb{R}^{d}$ with $\norm{x} = 1$ on which a perturbation $p$ is applied, the resultant vector is given by $p + x $. If we normalize this vector to obtain $y = \frac{p + x}{\norm{p + x}}$, then we have $\norm{\frac{p}{\norm{p}} - y} \;\leq\; \sqrt{2 - 2\sqrt{1 - \frac{1}{\norm{p}^2}}}$ whenever $\norm{p} > 1$.

\subsubsection{Proof: } Let $\hat{p} = \frac{p}{\norm{p}}$. Expanding out the expression for norm:
\begin{align*}
\norm{\hat{p} - y} &= \sqrt{(\hat{p}-y)^{\dagger}(\hat{p}-y)}\\
&=\sqrt{\Big( \hat{p} - \frac{ p + x}{\norm{ p + x}}\Big)^{\dagger}\cdot \Big( \hat{p} - \frac{ p + x}{\norm{p + x}}\Big)}\\
       &= \sqrt{2 - \Big( \frac{(p+x)^{\dagger}\hat{p} + \hat{p}^{\dagger}(p+x)}{\norm{ p + x}}\Big)}
\end{align*}
where we have used the fact that $\norm{\hat{p}} = 1$. Since $p$ and $x$ are real, we have:
\begin{equation*}
    (p+x)^{\dagger}\hat{p} = \hat{p}^{\dagger}(p+x) = \norm{p} + \hat{p}^Tx
\end{equation*}
Therefore, we have:
\begin{equation*}
    \norm{\hat{p} - y} = \sqrt{2 - 2\Big( \frac{\norm{p} + \hat{p}^Tx}{\norm{ p + x}}\Big)}
\end{equation*}
Expanding out the term $\norm{ p + x}$, we have:
\begin{equation}
    \norm{\hat{p} - y} = \sqrt{2 - 2\Big( \frac{\norm{p} + \hat{p}^Tx}{\sqrt{\norm{p}^2 + 2\norm{p}\hat{p}^Tx + 1}}\Big)} \label{eq:reclemm3}
\end{equation}
using the fact that $\norm{x} = 1$ and that $x$ and $p$ are real. Let $z = \hat{p}^Tx$. Clearly, $-1 \leq z \leq 1$ since $\norm{\hat{p}} = \norm{x} = 1$. We write:
\begin{equation}
    \norm{\hat{p} - y}  = \sqrt{2 - 2f(z)} \label{eq:pysqrt}
\end{equation}
where
\begin{equation*}
    f(z) = \frac{\norm{p} + z}{\sqrt{\norm{p}^2 + 2\norm{p}z + 1}}
\end{equation*}
Computing the first derivative function, we have:
\begin{align*}
    f'(z) &=  \frac{{\norm{p}^2 + 2\norm{p}z + 1} - (\norm{p}+z){\norm{p}}}{({\norm{p}^2 + 2\norm{p}z + 1})^{\frac{3}{2}}}  \\
    &= \frac{{\norm{p}z + 1}}{({\norm{p}^2 + 2\norm{p}z + 1})^{\frac{3}{2}}} 
\end{align*}
Now,
\begin{equation*}
   {\norm{p}^2 + 2\norm{p}z + 1} = {(\norm{p} + z)^2 + (1 -z)^2}
\end{equation*}
Note that since $-1 \leq z \leq 1$, this expression is strictly positive provided $\norm{p} > 1$. Hence, the derivative function $f'(z)$ is well defined provided $\norm{p} > 1$. Therefore, for $\norm{p} > 1$, we can compute the stationary points by setting $f'(z)$ to zero to obtain:
\begin{align*}
    z_s &= -\frac{1}{\norm{p}}
\end{align*}
where $z_s$ denotes the stationary point. Note that $f'(z)$ is negative for $z < z_s$ and positive for $z > z_s$, implying $z_s$ is the global minimum for $f(z)$. Therefore, the minimum value of $f(z)$, $f_m$ is given by:
\begin{align*}
    f_m &= f(z_s) \\
        &= \sqrt{1 - \frac{1}{\norm{p}^2}}
\end{align*}
From \eqref{eq:pysqrt} it is clear that  $\norm{\hat{p} - y}$ is maximum when $f(z)$ is minimum. Therefore, we have:
\begin{align*}
    \norm{\hat{p} - y} &\leq \sqrt{2 - 2f_m} \\
    &\leq \sqrt{2 - 2\sqrt{1 - \frac{1}{\norm{p}^2}}}
\end{align*}
provided $\norm{p} > 1$, which is the required result.

\subsection{Theorem 1}

\subsubsection{Statement:}
For an additive universal adversarial perturbation $p$ applied on inputs of classifier $\mathcal{Q}$, a strength of perturbation $\norm{p}\in\mathbb{R}$ will cause $\mathcal{Q}$ to classify all inputs as $c$ (class to which $p/\norm{p}$ belongs) if:$$\norm{p} \geq \frac{2}{\epsilon_c\sqrt{4 - \epsilon_c^2}}$$ where $\epsilon_c$ is given by: $$\epsilon_c = \sqrt{1 + \frac{1}{2d}\cdot \Big( \hat{p}^TM^c\hat{p}  - \hat{p}^TM^{c'}\hat{p}}\Big) - 1$$ where $\hat{p} = p/\norm{p}$ and $c'$ is the class with highest output probability for $\hat{p}$ after $c$. 

\subsubsection{Proof:}
Let $y$ denote the vector obtained when perturbation $p$ is applied to an input $x$ after normalization. Clearly,
\begin{equation*}
    y = \frac{p + x}{\norm{p + x}}
\end{equation*}
Using Lemma 2, we can write:
\begin{equation}
    \abs{P(\hat{c}_{y} = c) - P(\hat{c}_{\hat{p}} = c)} \leq d\big(\epsilon^2 + 2\epsilon\big) \label{eq:pcypcp}
\end{equation}
where $\epsilon = \norm{\hat{p} - y}$. Now, for all inputs to be classified as $c$, we must have:
\begin{equation}
    P(\hat{c}_{y} = c)\geq \max_{\Tilde{c}\neq c} (P(\hat{c}_{y} = \Tilde{c})) \label{eq:pcneqc}
\end{equation}
Using Lemma 2 for $\Tilde{c}$, we have:
\begin{equation*}
    \abs{P(\hat{c}_{y} = \Tilde{c}) - P(\hat{c}_{\hat{p}} = \Tilde{c})} \leq d\big(\epsilon^2 + 2\epsilon\big)
\end{equation*}
We get an upper bound of the form:
\begin{equation}
    {P(\hat{c}_{y} = \Tilde{c})  \leq P(\hat{c}_{\hat{p}} = \Tilde{c})} + d\big(\epsilon^2 + 2\epsilon\big) \label{eq:pcyup}
\end{equation}
We also obtain a lower bound on $P(\hat{c}_{y} = c)$ from \eqref{eq:pcypcp}(64):
\begin{equation}
    {P(\hat{c}_{y} = {c})  \geq P(\hat{c}_{\hat{p}} = {c})} - d\big(\epsilon^2 + 2\epsilon\big) \label{eq:pcylow}
\end{equation}
For \eqref{eq:pcneqc} to hold true, it is sufficient to have:
\begin{equation*}
    \min (P(\hat{c}_{y} = {c})) \geq \max_{\Tilde{c}\neq c} ({P(\hat{c}_{y} = \Tilde{c}}) 
\end{equation*}
Using bounds \eqref{eq:pcyup} and \eqref{eq:pcylow}, we rewrite this as:
\begin{align*}
    P(\hat{c}_{\hat{p}} = {c}) - d\big(\epsilon^2 + 2\epsilon\big) \geq \max_{\Tilde{c} \neq c} &[ P(\hat{c}_{\hat{p}} = \Tilde{c}) \\
    & + d\big(\epsilon^2 + 2\epsilon\big)] 
\end{align*}
Since $d\big(\epsilon^2 + 2\epsilon\big)$ is independent of $\Tilde{c}$, we rewrite this as:
\begin{align}
    P(\hat{c}_{\hat{p}} = {c}) - d\big(\epsilon^2 + 2\epsilon\big) \geq \max_{\Tilde{c} \neq c} &[ P(\hat{c}_{\hat{p}} = \Tilde{c})] \nonumber\\
    & + d\big(\epsilon^2 + 2\epsilon\big) \label{eq:depst}
\end{align}
Let $c'$ be the class with second highest probability for $\hat{p}$. Then clearly,
\begin{equation}
    \max_{\Tilde{c} \neq c} [ P(\hat{c}_{\hat{p}} = \Tilde{c})] =  P(\hat{c}_{\hat{p}} = {c'}) \label{eq:maxpc}
\end{equation}
Substituting \eqref{eq:maxpc} in \eqref{eq:depst} and rearranging, we have:
\begin{equation*}
    (2d)\epsilon^2 + (4d)\epsilon - (P(\hat{c}_{\hat{p}} = {c}) - P(\hat{c}_{\hat{p}} = {c'})) \leq 0
\end{equation*}
Let $\alpha = P(\hat{c}_{\hat{p}} = {c}) - P(\hat{c}_{\hat{p}} = {c'})$. Since this is a quadratic equation, this inequality holds true if and only if:
\begin{equation*}
    -1 - \sqrt{1+\frac{\alpha}{2d}} \leq \epsilon \leq -1 + \sqrt{1+\frac{\alpha}{2d}}
\end{equation*}
Note that by definition $\epsilon \geq 0$. Therefore, we have:
\begin{equation*}
    \epsilon \leq -1 + \sqrt{1+\frac{\alpha}{2d}} \label{eq:epsminus}
\end{equation*}
But from Lemma 1, we have:
\begin{equation*}
    \alpha =  \hat{p}^\dagger M^c \hat{p} - \hat{p}^\dagger M^{c'} \hat{p}
\end{equation*}
We can then rewrite \eqref{eq:epsminus} as:
\begin{equation}
    \norm{\hat{p}-y} \leq -1 + \sqrt{1+\frac{(\hat{p}^\dagger M^c \hat{p} - \hat{p}^\dagger M^{c'} \hat{p})}{2d}} \label{eq:allyc}
\end{equation}
However, using Lemma 3, we can bound the norm of this displacement as:
\begin{equation}
    \norm{\hat{p} - y} \;\leq\; \sqrt{2 - 2\sqrt{1 - \frac{1}{\norm{p}^2}}} \label{eq:pyupper}
\end{equation}
For all $y$ to be classified as $c$, \eqref{eq:allyc} must be satisfied. This would imply that the maximum value of $\norm{\hat{p} - y}$ should satisfy \eqref{eq:allyc}. A sufficient condition for this to happen, provided $\norm{p} > 1$, is:
\begin{equation}
    \sqrt{2 - 2\sqrt{1 - \frac{1}{\norm{p}^2}}} \leq -1 + \sqrt{1+\frac{(\hat{p}^\dagger M^c \hat{p} - \hat{p}^\dagger M^{c'} \hat{p})}{2d}} \label{eq:aana}
\end{equation}
since \eqref{eq:pyupper} provides an upper bound on $\norm{\hat{p} - y}$. Let us define:
\begin{equation*}
    \epsilon_c = -1 + \sqrt{1+\frac{(\hat{p}^\dagger M^c \hat{p} - \hat{p}^\dagger M^{c'} \hat{p})}{2d}}
\end{equation*}
\eqref{eq:aana} then can be rewritten as:
\begin{equation*}
    \sqrt{2 - 2\sqrt{1 - \frac{1}{\norm{p}^2}}} \leq \epsilon_c
\end{equation*}
Since LHS and RHS are non-negative, the inequality holds true after squaring both sides:
\begin{equation*}
    {2 - 2\sqrt{1 - \frac{1}{\norm{p}^2}}} \leq \epsilon_c^2
\end{equation*}
Rearranging, we obtain:
\begin{equation}
    {\sqrt{1 - \frac{1}{\norm{p}^2}}} \geq 1 - \frac{\epsilon_c^2}{2} \label{eq:sqrt1p}
\end{equation}
To square both sides, we require RHS to be non-negative since LHS is non-negative. To check if this is the case, we note that:
\begin{equation*}
    \epsilon_c \leq -1 + \sqrt{1+\frac{1}{2d}}
\end{equation*}
since both $\hat{p}^\dagger M^c \hat{p}$ and $\hat{p}^\dagger M^{c'} \hat{p}$ are bounded between $0$ and $1$. Since $d$ represents the dimensionality of input vectors, $d \geq 1$. Therefore, we have:
\begin{align}
    \epsilon_c &\leq -1 + \sqrt{\frac{3}{2}} \label{eq:ecroot3}
\end{align}
And therefore:
\begin{equation*}
    1 - \frac{\epsilon_c^2}{2} > 0
\end{equation*}
Now that we have established that RHS is also non-negative, we can rewrite \eqref{eq:sqrt1p} as:
\begin{equation*}
    {1 - \frac{1}{\norm{p}^2}} \geq \Big(1 - \frac{\epsilon_c^2}{2} \Big)^2
\end{equation*}
which we simplify to obtain:
\begin{equation*}
    \frac{1}{\norm{p}^2} \leq {\epsilon_c^2} - \frac{\epsilon_c^4}{4}
\end{equation*}
Since both LHS and RHS are non-negative, this inequality can be rearranged to obtain:
\begin{equation*}
    \norm{p} \geq \frac{2}{\epsilon_c \sqrt{4 -\epsilon_c^2 }}
\end{equation*}
which is the required result. 

\subsection{Lemma 4}

\subsubsection{Statement: }
For any $x\in\mathbb{R}^{d}$ with $\norm{x} = 1$ on which a perturbation $p$, such that $\norm{p} \geq \delta$, is applied, the resultant vector is given by $p + x $. If we normalize this vector to obtain $y = \frac{p + x}{\norm{p + x}}$, then we have $\norm{\hat{p} - y} \;\leq\; \sqrt{2 - 2\Big( \frac{\delta + \hat{p}^Tx}{\sqrt{\delta^2 + 2\delta \hat{p}^Tx + 1}}\Big)}$ where ${\hat{p}} = \frac{p}{\norm{p}}$.

\subsubsection{Proof: }
The key idea behind the lemma is to explore the effect of setting perturbation norm to be greater than $\delta$. We aim to set an upper limit on the distance between original and reprojected vectors after such a perturbation is applied.

Recall \eqref{eq:reclemm3} from the proof of Lemma 3:

\begin{equation*}
    \norm{\hat{p} - y} = \sqrt{2 - 2\Big( \frac{\norm{p} + \hat{p}^Tx}{\sqrt{\norm{p}^2 + 2\norm{p}\hat{p}^Tx + 1}}\Big)}
\end{equation*}
Let $z = \norm{p}$. Consider $f(z)$ defined as:
\begin{equation*}
    f(z) = \frac{z + \hat{p}^Tx}{\sqrt{z^2 + 2z\hat{p}^Tx + 1}}
\end{equation*}
Computing the first derivative, we have:
\begin{equation*}
    f'(z) = \frac{1 - (\hat{p}^Tx)^2}{({z^2 + 2z\hat{p}^Tx + 1})^{\frac{3}{2}}}
\end{equation*}
Since $(\hat{p}^Tx)^2 \leq 1$, $f'(z) \geq 0$ always. Therefore $f(z)$ is increasing with $z$. For $\norm{\hat{p} - y}$ to be maximum, $f(z)$ must be minimum since:
\begin{equation*}
    \norm{\hat{p} - y} = \sqrt{2 - 2f(z)}
\end{equation*}
Since $z \geq \delta$, minimum of $f(z)$ occurs at $z = \delta$. Therefore, $\norm{\hat{p} - y}$ is maximum at  $z = \delta$. Then we have:
\begin{equation*}
    \norm{\hat{p} - y} \;\leq\; \sqrt{2 - 2\Big( \frac{\delta + \hat{p}^Tx}{\sqrt{\delta^2 + 2\delta \hat{p}^Tx + 1}}\Big)}
\end{equation*}
which is the required result.

\subsection{Theorem 2}

\subsubsection{Statement:}

For an additive universal adversarial perturbation $p$ applied on inputs of classifier $\mathcal{Q}$, with the constraint $\norm{p} \leq \delta$, an input $x$ is predicted as belonging to class $c$ (class to which $p/\norm{p}$ belongs) if any one of the following conditions hold true:
\begin{enumerate}
    \item If $\delta \geq \frac{2}{\epsilon_c\sqrt{4 - \epsilon_c^2}}$ 
    \item If $1 \leq \delta < \frac{2}{\epsilon_c\sqrt{4 - \epsilon_c^2}}$ and ($\hat{p}^Tx \leq t_1$ or $\hat{p}^Tx \geq t_2$)
    \item If $\delta < 1 $ and $\hat{p}^Tx \geq t_2$
\end{enumerate}
where $\hat{p} = p/\norm{p}$, thresholds $t_1, t_2$  are in $[-1, 1]$, $t_1 \leq t_2$ and $t_1, t_2$ are the solutions of the quadratic equation:
$$ t^2 + 2\delta \epsilon't + (\epsilon' (\delta^2+1)-1) = 0 $$
where $\epsilon' = \epsilon_c^2 - \frac{\epsilon_c^4}{4}$
($\epsilon_c$ defined in Theorem 1).

\subsubsection{Proof:} From Theorem 1, we know that:
\begin{equation}
    \norm{\hat{p}-y} \leq \epsilon_c \label{eq:th2:py}
\end{equation}
where $y = \frac{p + x}{\norm{p + x}}$ for $x$ to be classified as class $c$ after applying perturbation $p$. Now, $\norm{p}$ is constrained to be less than $\delta$. Recall from the proof of Lemma 4 that as $\norm{p}$ increases, $\norm{\hat{p}-y}$ decreases. To achieve tightest possible bound within the constraint, we choose $\norm{p} = \delta$. Therefore, to achieve \eqref{eq:th2:py}, it is sufficient to have:
\begin{equation*}
    \sqrt{2 - 2\Big( \frac{\delta + \hat{p}^Tx}{\sqrt{\delta^2 + 2\delta \hat{p}^Tx + 1}}\Big)} \leq \epsilon_c
\end{equation*}
where we again use Lemma 4. Let us define $z = \hat{p}^Tx$. Then we can rewrite this as:
\begin{equation}
    \sqrt{2 - 2f(z)} \leq \epsilon_c \label{eq:th2:fz}
\end{equation}
 where:
 \begin{equation}
    f(z) = \frac{\delta + z}{\sqrt{\delta^2 + 2\delta z + 1}} \label{eq:thm2:fz}
\end{equation}
Computing the derivative:
\begin{equation*}
    f'(z) = \frac{{\delta z + 1}}{({\delta ^2 + 2\delta z + 1})^{\frac{3}{2}}} 
\end{equation*}
We consider two cases: $\delta \geq 1$ and $\delta < 1$.

\subsubsection{Case 1: $\delta \geq 1$}

In this case, $f'(z) = 0$ when $z = z_s = -1/\delta$. Note that $z$ is constrained to be between $-1$ and $1$. Clearly, $f'(z) < 0$ for $z < z_s$ and $f'(z) > 0$ for $z > z_s$. Therefore, $f(z)$ is minimum at $z = z_s$. Now, let $f(z_s)$ be such that:
\begin{equation}
    \sqrt{2 - 2f(z_s)} \leq \epsilon_c \label{eq:cs1:fzs}
\end{equation}
Note that from the proof of Theorem 1, it can be seen that this is equivalent to:
\begin{equation*}
    \delta \geq \frac{2}{\epsilon_c \sqrt{4 -\epsilon_c^2 }}
\end{equation*}
Since $f(z_s)$ is the minimum of $f(z)$, \eqref{eq:cs1:fzs} implies \eqref{eq:th2:fz}.This proves condition 1 in the statement of Theorem 2. 

Let $g(z) = \sqrt{2 - 2f(z)}$. Now, if $1 \leq \delta < \frac{2}{\epsilon_c \sqrt{4 -\epsilon_c^2 }}$, $z_s$ must be such that:
\begin{equation}
    g(z_s) > \epsilon_c \label{eq:cs1:gz}
\end{equation}
Now, consider the value of $f(z)$ at $z = -1$:
\begin{align*}
    f(-1) &= \frac{\delta - 1}{\sqrt{\delta^2 + 2\delta(-1) + 1}} \\
    &= \frac{\delta - 1}{\sqrt{(\delta - 1)^2}} \\
    &= \frac{\delta - 1}{\abs{(\delta - 1)}} = 1
\end{align*}
since $\delta > 1$. Similarly, $f(1) = 1$. Therefore, we have
\begin{equation}
    g(-1) = g(1) = 0 \label{eq:cs1:g1}
\end{equation}
 Also note that since $f'(z) < 0$ for $0 \leq z < z_s$, $f(z)$ is strictly decreasing and therefore $g(z)$ is strictly increasing. Therefore, using Intermediate Value Theorem (IVT), there exists a unique $z_1 \in [-1, z_s)$ such that  $g(z_1) = \epsilon_c$ since we have:
 \begin{equation*}
     g(-1) \leq \epsilon_c < g(z_s)
 \end{equation*}
  from \eqref{eq:cs1:gz} and \eqref{eq:cs1:g1}. Similarly, there exists a unique $z_2 \in (z_s, 1]$ such that  $g(z_2) = \epsilon_c$. Note that we require $z$ such that $g(z) <= \epsilon_c$. For this to happen:
  \begin{equation*}
      -1 \leq z \leq z_1 \textrm{ or } z_2 \leq z \leq 1
  \end{equation*}
To compute $z_1$ and $z_2$, we solve:
\begin{equation*}
    g(z) = \epsilon_c
\end{equation*}
which can be rewritten to yield:
\begin{equation*}
    (2 - 2f(z)) - \epsilon_c^2 = 0
\end{equation*}
Simplifying the expression using \eqref{eq:thm2:fz}, we have:
\begin{equation}
    z^2 + 2 \delta \epsilon' z + (\epsilon'(\delta^2+1)  -1) = 0 \label{eq:cs1:quad}
\end{equation}
where $\epsilon' = \epsilon_c^2 - \epsilon_c^4/4$.  $z1$ and $z2$ are the solutions to this quadratic equation. Note that $z_1 < z_2$. Note that we have already proved the existence of $z_1$ and $z_2$ and hence valid solutions are guaranteed, provided $1 \leq \delta < \frac{2}{\epsilon_c \sqrt{4 -\epsilon_c^2 }}$. This proves condition 2 of Theorem 2.

\subsubsection{Case 2: $\delta < 1$} In this case, $f'(z) > 0$ for all $z$ in $[-1, 1]$ since $\delta z + 1$ is always positive. Therefore, $f(z)$ is strictly increasing and $g(z)$ is strictly decreasing. Note that now we have:
\begin{equation*}
    f(-1) = \frac{\delta - 1}{\abs{(\delta - 1)}} = -1
\end{equation*}
since $\delta < 1$. Therefore, $g(-1) = 4$. Similar computations show that $f(1) = 1$ and therefore $g(1) = 0$. In other words, as $z$ increases from $-1$ to $1$, $g(z)$ decreases monotonously from $4$ to $0$. From \eqref{eq:ecroot3} we know that $\epsilon_c \in [0, -1+\sqrt{3/2}]$. Therefore, again using Intermediate Value Theorem, there exists a unique $z_3 \in [-1, 1]$ such that $g(z_3) = \epsilon_c$. Again, to compute $z_3$, we solve \eqref{eq:cs1:quad}. To find out which solution of \eqref{eq:cs1:quad} is the required solution, we first note that if  $g(z_3) = \epsilon_c$:
\begin{equation*}
    \sqrt{2 - 2f(z_3)} < \sqrt{2}
\end{equation*}
because of \eqref{eq:ecroot3}. This implies that $f(z_3) > 0$ and therefore $z_3 > -\delta$. Now, consider the smaller solution of \eqref{eq:cs1:quad}:
\begin{equation*}
    z' = -\delta \epsilon' - \sqrt{(\delta \epsilon'-1)(\epsilon'-1)}
\end{equation*}
Clearly,
\begin{equation}
    z' \leq -\delta \epsilon' \label{eq:cs2:zdelta}
\end{equation}
Again, using \eqref{eq:ecroot3}, we have:
\begin{equation*}
    0 \leq \epsilon' < 1
\end{equation*}
Using this, we can write:
\begin{equation*}
    {\delta \epsilon'} < {\delta}
\end{equation*}
And therefore
\begin{equation}
    -{\delta \epsilon'} > -{\delta} \label{eq:cs2:zless}
\end{equation}
Using \eqref{eq:cs2:zdelta} and \eqref{eq:cs2:zless}, we have:
\begin{equation*}
    z' < -\delta
\end{equation*}
But we know $z_3 > -\delta$. Therefore $z_3 \neq z'$, i.e., $z_3$ cannot be the smaller solution of \eqref{eq:cs1:quad}. Therefore, $z_3$ must be the larger solution of \eqref{eq:cs1:quad}. And since $g(z)$ is monotonously decreasing, we have:
\begin{equation*}
    g(z) \leq \epsilon_c
\end{equation*}
for all $z\geq z_3 $, when $\delta < 1$. This proves condition 3 of Theorem 2, thereby completing the proof for Theorem 2.

\section{Existence of Additive UAPs}

We develop an intuition for additive UAPs by considering the specific example of single qubit binary classification. Consider a quantum classifier $\mathcal{Q}$ with one input qubit, in state $\ket{\psi}$, and one ancillary qubit. Without loss of generality we assume the ancillary qubit to be in state $\ket{0}$ so that the system is in state $\ket{\psi} \otimes \ket{0}$. Note that we restrict $\ket{\psi}$ to be of the form $\ket{\psi} =\begin{bmatrix} x & y \end{bmatrix}^T$ where $x, y \in \mathbb{R}$. This would mean that the input states can be represented as points on a unit circle centered at the origin. In practice, this is the case when classical data is embedded using amplitude encoding. 

\subsection{Dataset Details: XOR}
For illustration purposes, we generate a simple XOR dataset in $\mathbb{R}^2$. Samples $\{\begin{bmatrix}x_i & y_i\end{bmatrix}\}_{i = 1}^{N}$ are uniformly sampled from the unit circle $x^2 + y^2 = 1$, and for a given sample $\begin{bmatrix}x_i\ & y_i\end{bmatrix}^T$ in the dataset, the label is calculated as $c_i = s({x_i})\oplus  s({y_i})$ where $\oplus$ denotes the XOR operation and the function $s()$ is defined as:
\begin{equation}
    s(x)= 
\begin{cases}
    1 ,& \text{if } x\geq 0\\
    0,              & \text{otherwise}
\end{cases}
\end{equation}
We generate a total of $N = 1000$ data samples, out of which $850$ samples are used for training the classifier $\mathcal{Q}$. The remaining $150$ samples are used for evaluating the performance of the trained classifier. Since the dataset is rather simple, the classifier achives $\approx 100\%$ accuracy on both train and test sets.

\begin{figure*}
\centering
    \begin{subfigure}{0.23\textwidth}
    \centering
    \includegraphics[width=0.8\textwidth]{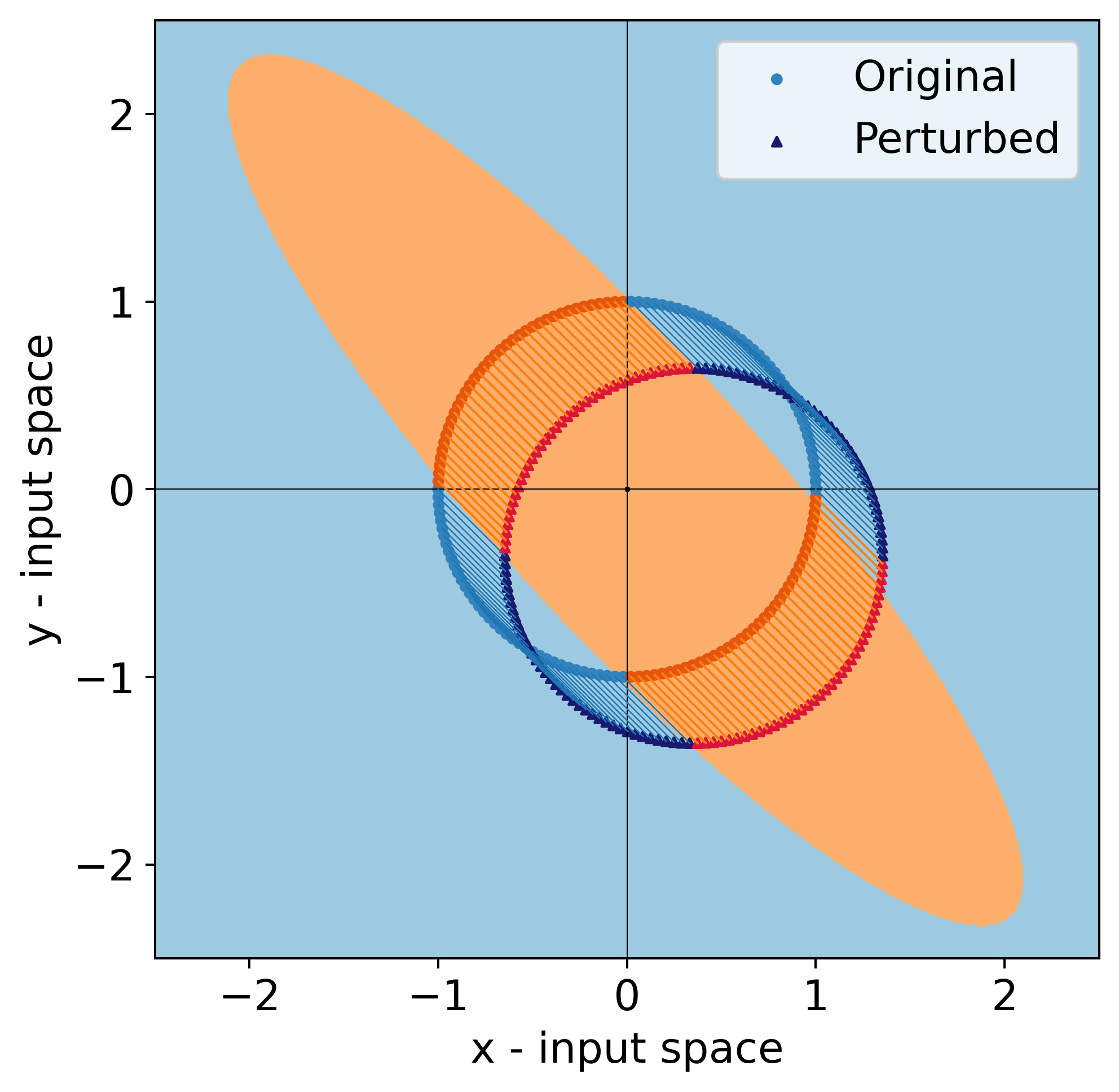}
    \caption{$\delta=0.5$}
    \end{subfigure}
    \begin{subfigure}{0.23\textwidth}
    \centering
    \includegraphics[width=0.8\textwidth]{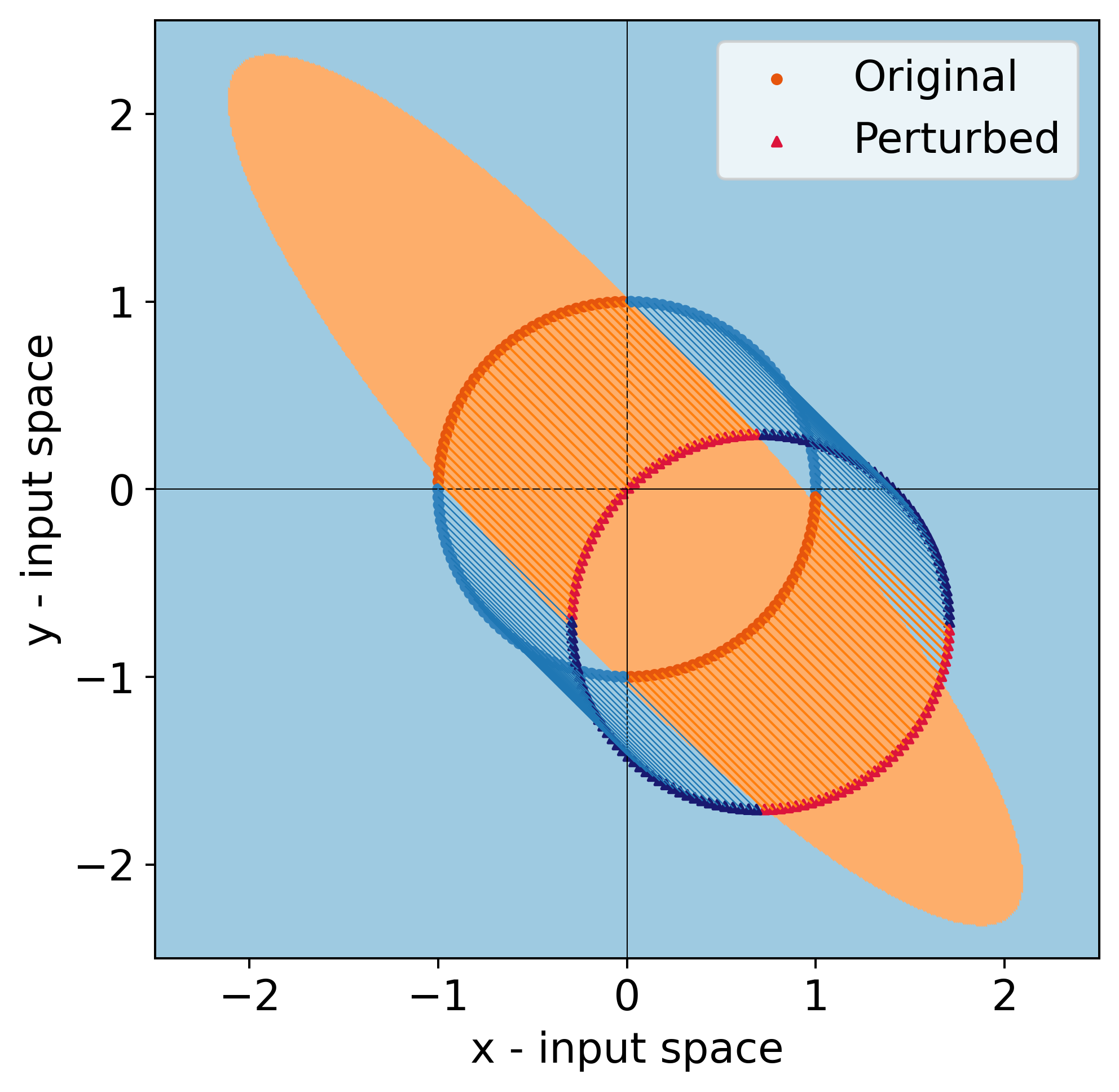}
    \caption{$\delta=1$}
    \end{subfigure}
    \begin{subfigure}{0.23\textwidth}
    \centering
    \includegraphics[width=0.8\textwidth]{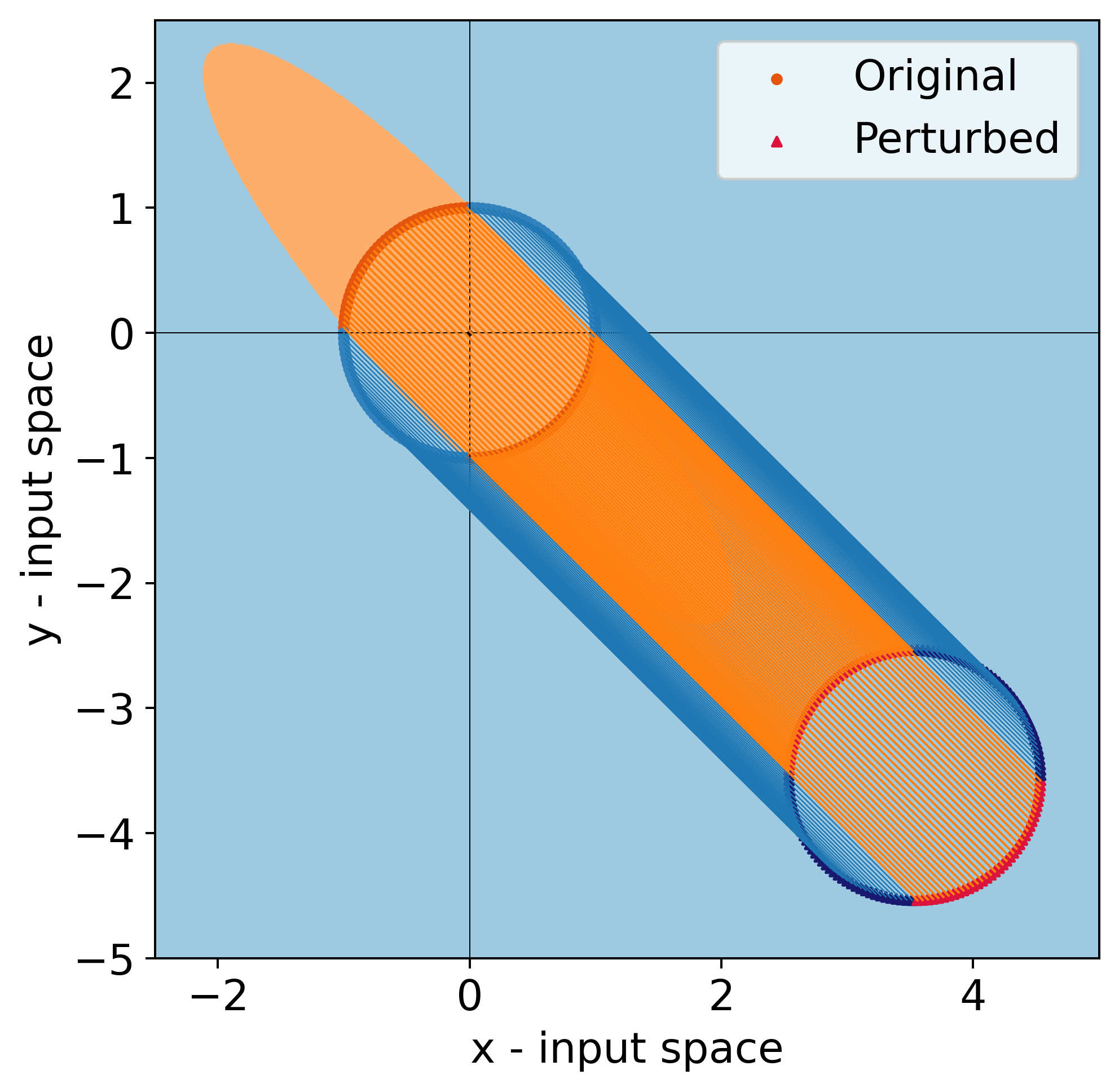}
    \caption{$\delta=5$}
    \end{subfigure}
    \begin{subfigure}{0.23\textwidth}
    \centering
    \includegraphics[width=0.8\textwidth]{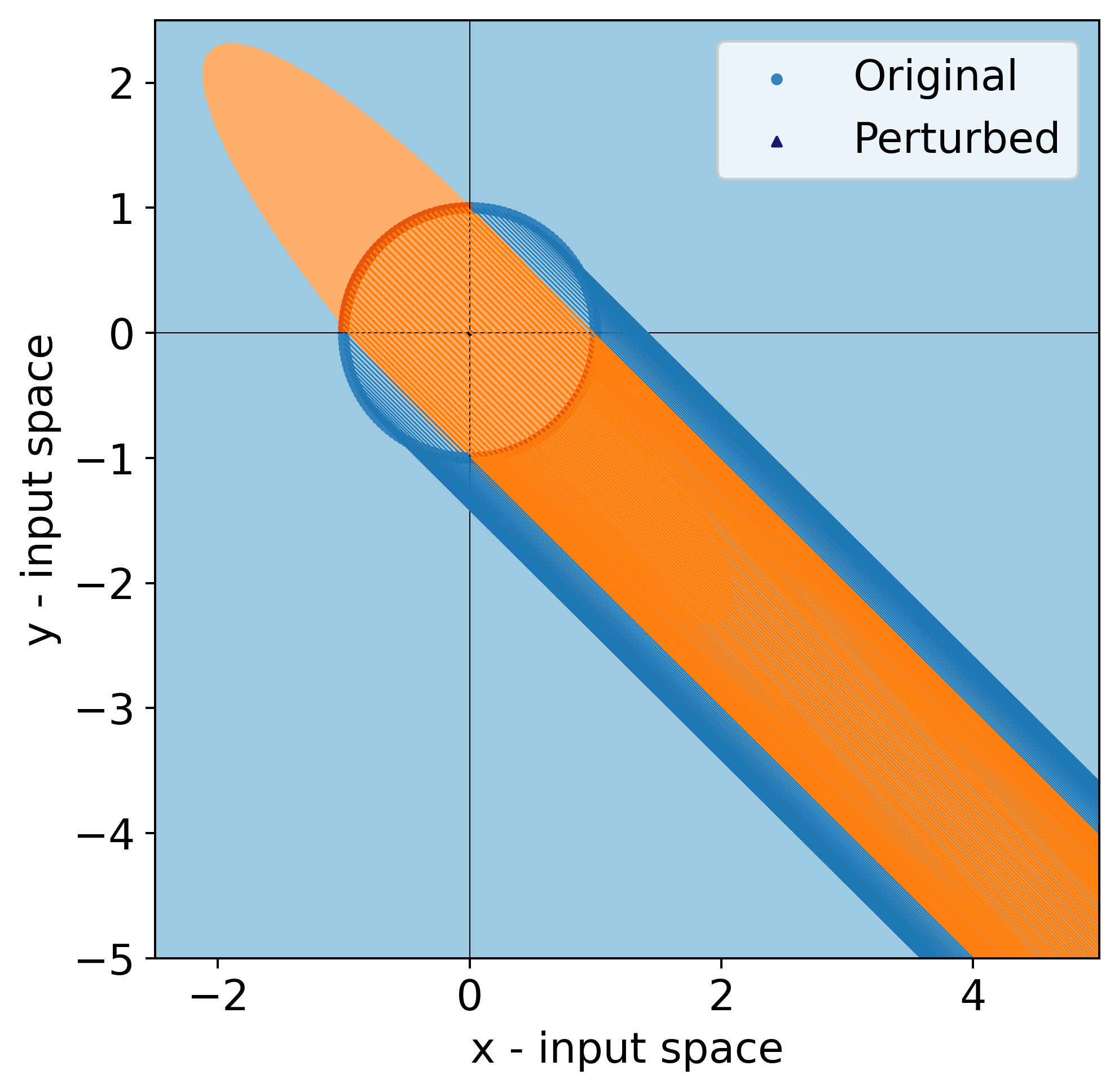}
    \caption{$\delta=50$}
    \end{subfigure}
    \begin{subfigure}{0.23\textwidth}
    \centering
    \includegraphics[width=0.8\textwidth]{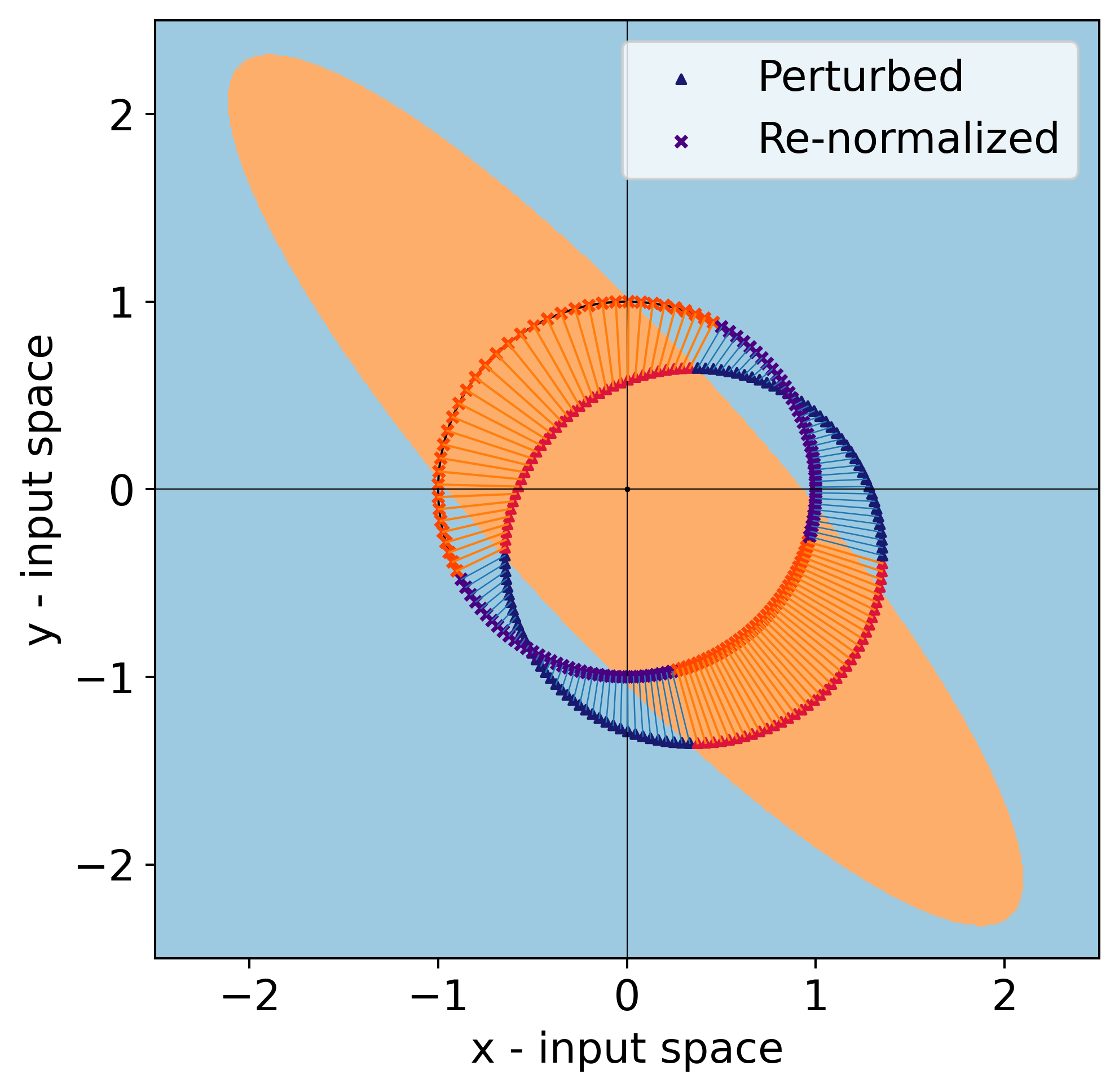}
    \caption{$\delta=0.5$}
    \end{subfigure}
    \begin{subfigure}{0.23\textwidth}
    \centering
    \includegraphics[width=0.8\textwidth]{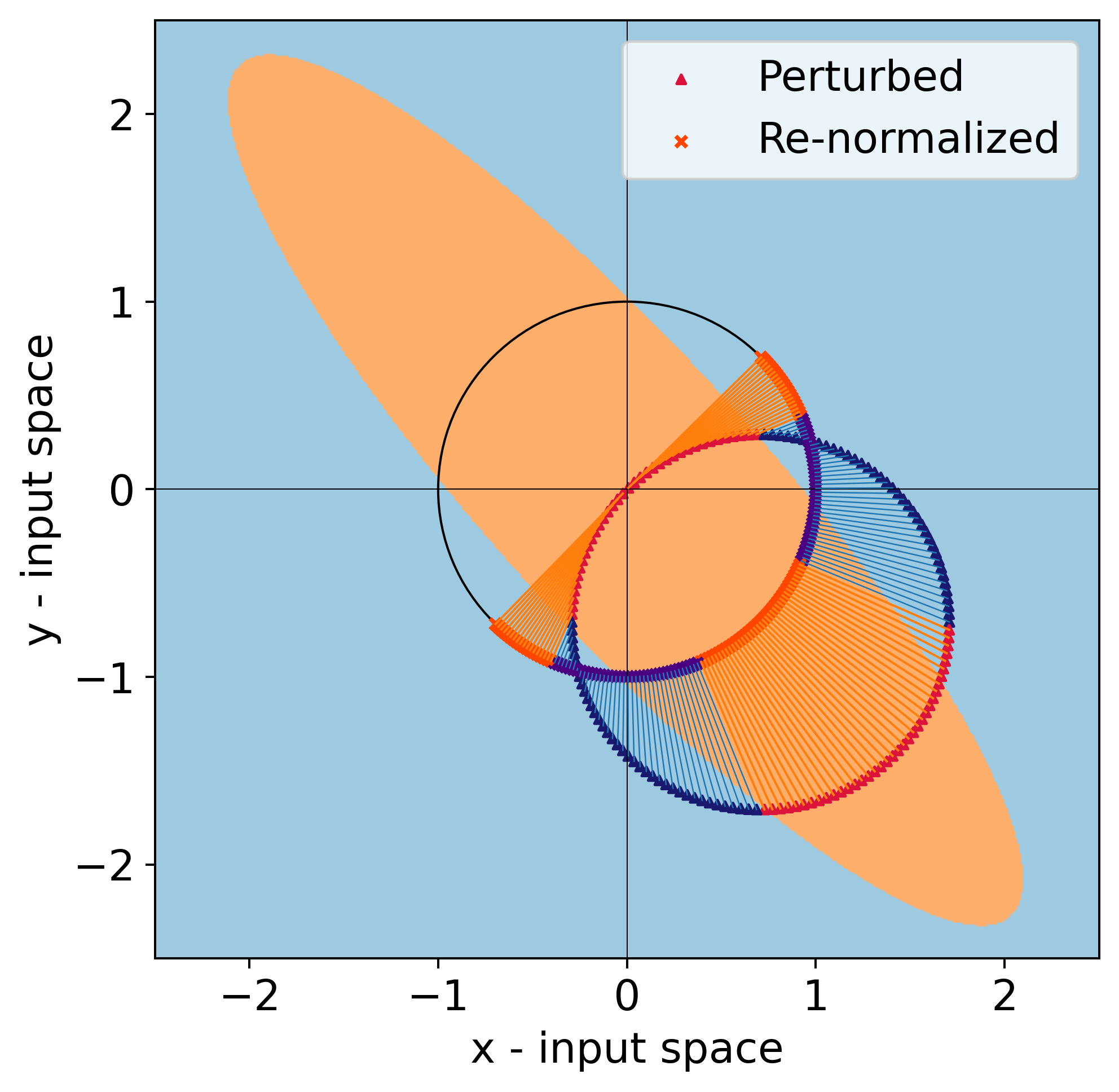}
    \caption{$\delta=1$}
    \end{subfigure}
    \begin{subfigure}{0.23\textwidth}
    \centering
    \includegraphics[width=0.8\textwidth]{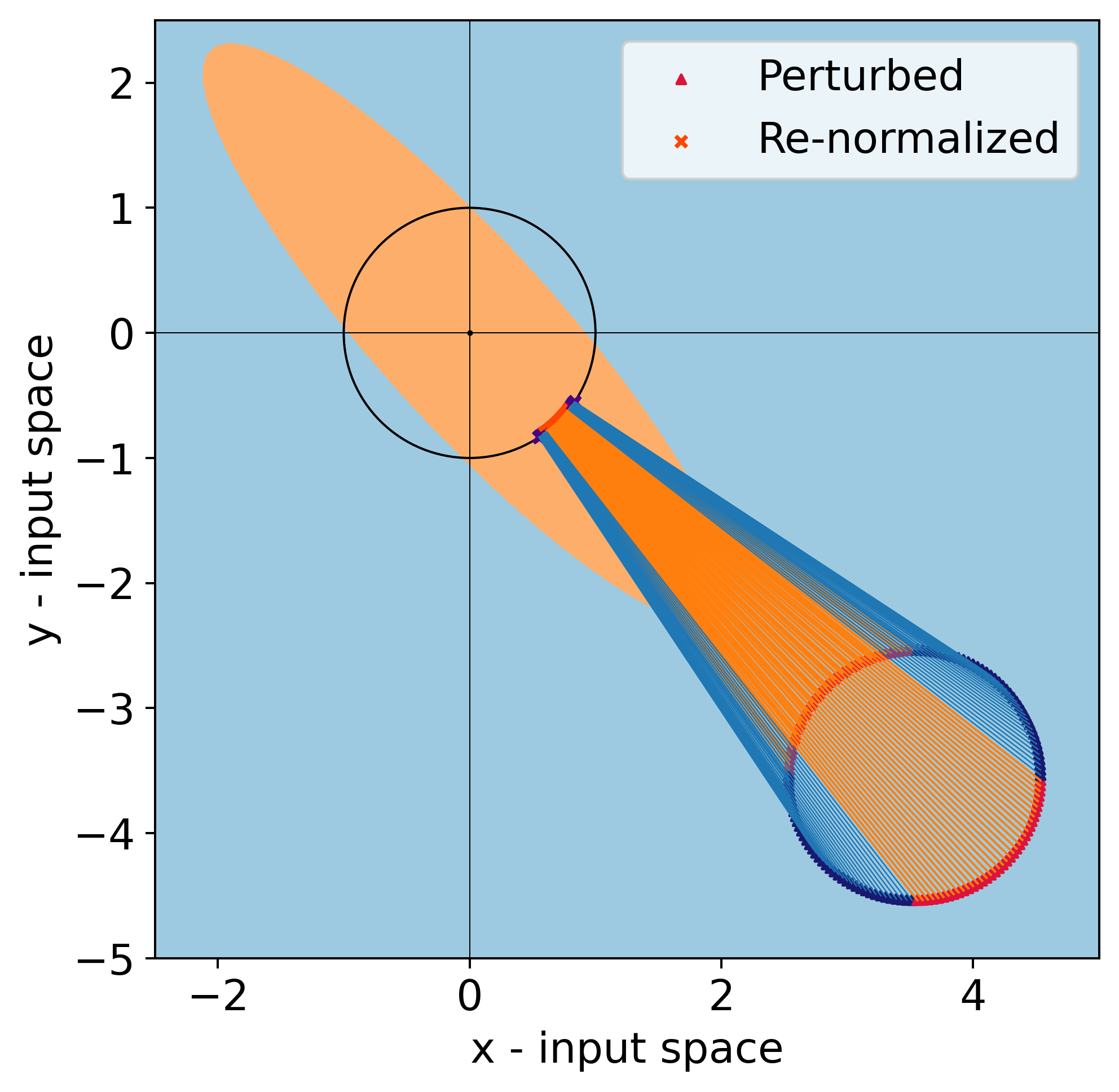}
    \caption{$\delta=5$}
    \end{subfigure}
    \begin{subfigure}{0.23\textwidth}
    \centering
    \includegraphics[width=0.8\textwidth]{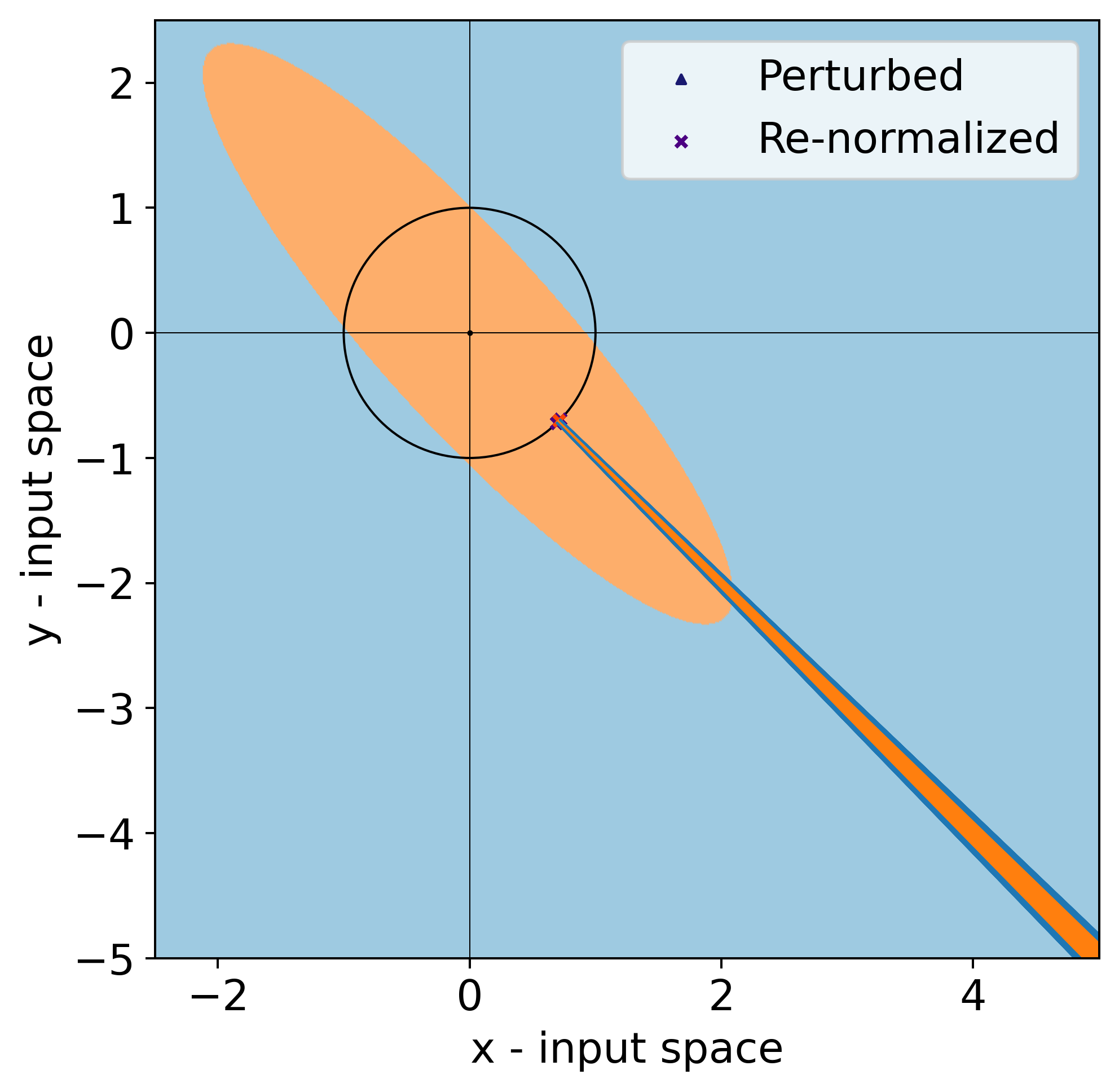}
    \caption{$\delta=50$}
    \end{subfigure}

    \caption{Effect of additive perturbation on the XOR dataset. The figures in the upper row represent the displacement of samples on the addition of perturbation. The bottom row represents that perturbed samples after re-normalization. As discussed, the ellipse defines the decision boundary of the classifier. The unit circle is marked as shown.}
    \label{fig:decisionboundary}
\end{figure*}

\subsection{Decision Boundary}

Let us denote $P(\hat{c}_x = 0)$ as p. For the particular case under consideration, invoking Lemma 1 yields:
\begin{equation}
 p = \begin{bmatrix}x_i\ & y_i\end{bmatrix} M  \begin{bmatrix}x_i\ \\ y_i\end{bmatrix}   \label{eq:debp}
\end{equation}
where
\begin{equation*}
    M = \begin{bmatrix} \abs{U_{0, 0}}^2 + \abs{U_{2, 0}}^2 & U_{0, 0}^{*}U_{0, 2}+U_{2, 0}^{*}U_{2, 2} \\ U_{0, 2}^{*}U_{0, 0}+U_{2, 2}^{*}U_{2, 0} & \abs{U_{0, 2}}^2 + \abs{U_{2, 2}}^2\end{bmatrix}
\end{equation*}
Rewriting \eqref{eq:debp}, we have:
\begin{equation*}
    \begin{bmatrix}x_i\ & y_i\end{bmatrix} M  \begin{bmatrix}x_i\ \\ y_i\end{bmatrix}  - p = 0
\end{equation*}
 Expanding out, we have:
\begin{equation}
    \alpha x_i^2 + \beta x_iy_i + \gamma y_i^2 - p = 0  \label{eq:genconic}
\end{equation}
where 
\begin{align*}
    \alpha &= \abs{U_{0, 0}}^2 + \abs{U_{2, 0}}^2 \\
    \beta &=U_{0, 0}^{*}U_{0, 2} + U_{0, 2}^{*}U_{0, 0} + U_{2, 0}^{*}U_{2, 2} + U_{2, 2}^{*}U_{2, 0}  \\
    \gamma &= \abs{U_{0, 2}}^2 + \abs{U_{2, 2}}^2
\end{align*}
Note that $\beta$ is real, since $\beta^* = \beta$. Equation \eqref{eq:genconic} represents a general conic. To identify the conic, we compute the discriminant:
\begin{equation*}
    D = \beta^2 - 4\alpha\gamma
\end{equation*}
Now using triangle inequality, we can write:
\begin{align*}
    \beta^2 \leq &\Big(\abs{U_{0, 0}^{*}U_{0, 2}} + \abs{U_{0, 2}^{*}U_{0, 0}} \\ &+ \abs{U_{2, 0}^{*}U_{2, 2}} + \abs{U_{2, 2}^{*}U_{2, 0}}\Big)^2 
\end{align*}
Which can be rewritten as:
\begin{equation}
    \beta^2 \leq 4(\abs{U_{0, 0}}\abs{U_{0, 2}} + \abs{U_{2, 0}}\abs{U_{2, 2}})^2 \label{eq:betasq}
\end{equation}
Now let $c = 4(\abs{U_{0, 0}}\abs{U_{0, 2}} + \abs{U_{2, 0}}\abs{U_{2, 2}})^2$. Then we have:
\begin{align*}
     4\alpha\gamma - c &= 4(\abs{U_{0, 0}}^2 + \abs{U_{2, 0}}^2 )(\abs{U_{0, 2}}^2 + \abs{U_{2, 2}}^2) - c \\
\end{align*}
Which we rewrite as:
\begin{align*}
    4\alpha\gamma - c = &4\Big(\abs{U_{0, 0}}^2\abs{U_{2, 2}}^2 +  \abs{U_{2, 0}}^2\abs{U_{0, 2}}^2 \\
     &\textrm{    } - 2\abs{U_{0, 0}}\abs{U_{0, 2}}\abs{U_{2, 0}}\abs{U_{2, 2}} \Big)
\end{align*}
Therefore,
\begin{equation*}
    4\alpha\gamma - c = 4\Big( \abs{U_{0, 0}}\abs{U_{2, 2}} - \abs{U_{2, 0}}\abs{U_{0, 2}} \Big)^2
\end{equation*}
Since the square of a real number is always non-negative, $4\alpha\gamma - c \geq 0$. Using \eqref{eq:betasq}, we then have:
\begin{equation*}
    D = \beta - 4\alpha\gamma \leq 0
\end{equation*}
Using the properties of conic sections, it can be seen that this case corresponds to the conic either being an ellipse or a pair of straight lines (degenerate case). Note that $\begin{bmatrix}x_i\ & y_i\end{bmatrix}^T$ must also lie on the unit circle. Therefore, samples are classified by $\mathcal{Q}$ as belonging to a particular class depending on whether the sample, which lies on the unit circle, lies inside the ellipse (or the pair of straight lines) described by equation \eqref{eq:genconic} or not.

\subsection{Effect of additive perturbation}

After training the classifier $\mathcal{Q}$, we study the effect of adding additive perturbations to the legitimate data samples. For illustration, we consider adding perturbations of the form $\delta \hat{p}$, where $\delta \in \mathbb{R}$ denotes the norm of the perturbation and $\hat{p} = \begin{bmatrix}\frac{1}{\sqrt{2}}\ & -\frac{1}{\sqrt{2}}\end{bmatrix}^T$. We add the perturbation to test samples and re-normalize the resulting vectors. The results are shown in Figure \ref{fig:decisionboundary}.

As has been discussed in the previous section, the prediction class depends on whether the sample lies inside an ellipse or not. The ellipse is plotted for reference in all cases. We observe that as the perturbation strength $\delta$ is increased, the re-normalized datapoints localize over smaller regions of the unit circle. Therefore, it seems like if you have a sufficiently large perturbation, all the re-normalized points can be localized to a required arc on the unit circle. By choosing a perturbation appropriately, this strategy may be used to project all samples into a region inside/outside the ellipse, thereby causing $\mathcal{Q}$ to predict all samples as belonging to a particular class. This intuition inspired the formalization of Theorems 1 and 2.

\section{Experimental Setup}
\subsection{Quantum Variational Models}
For all our experiments, we use parametrized quantum circuit (PQC) based quantum classifiers and generators. We assume all quantum circuits to be noiseless. As discussed in the main paper, the number of qubits of the classifier $\mathcal{Q}$ is given by $D+K$, where $D=\lceil\log_2 d\rceil$ data qubits are used to encode $d$-dimensional classical data and $K=\lceil\log_2 k\rceil$ ancillary qubits are used for a $k$-class classification task. Each ancillary qubit is set to $\ket{0}$ initially. The general structure of the models consists of $L$ layers, with each layer consisting of a rotation unit and an entanglement unit. In the rotation unit, we use an arbitrary single qubit rotation gate, provided in the Pennylane library, to carry out an arbitrary rotation over each qubit. The rotation gate used has $3$ component gates: $$\text{Rot}(\omega, \theta, \phi) = \text{RZ}(\phi)\cdot \text{RY}(\theta)\cdot \text{RZ}(\omega)$$ where $\omega$, $\theta$ and $\phi$ are trainable parameters. The rotation unit carries out an arbitrary rotation over every qubit. In order to entangle all the qubits, we use a number of CNOT gates such that each qubit is entangled with every other qubit. The overall unit is similar to the strongly entangling layers available in the Pennylane library. The general structure of a quantum classifier as described above is given in Figure \ref{fig:qclass}:
\begin{figure}[htbp]
    \centering
    \includegraphics[width=0.9\columnwidth]{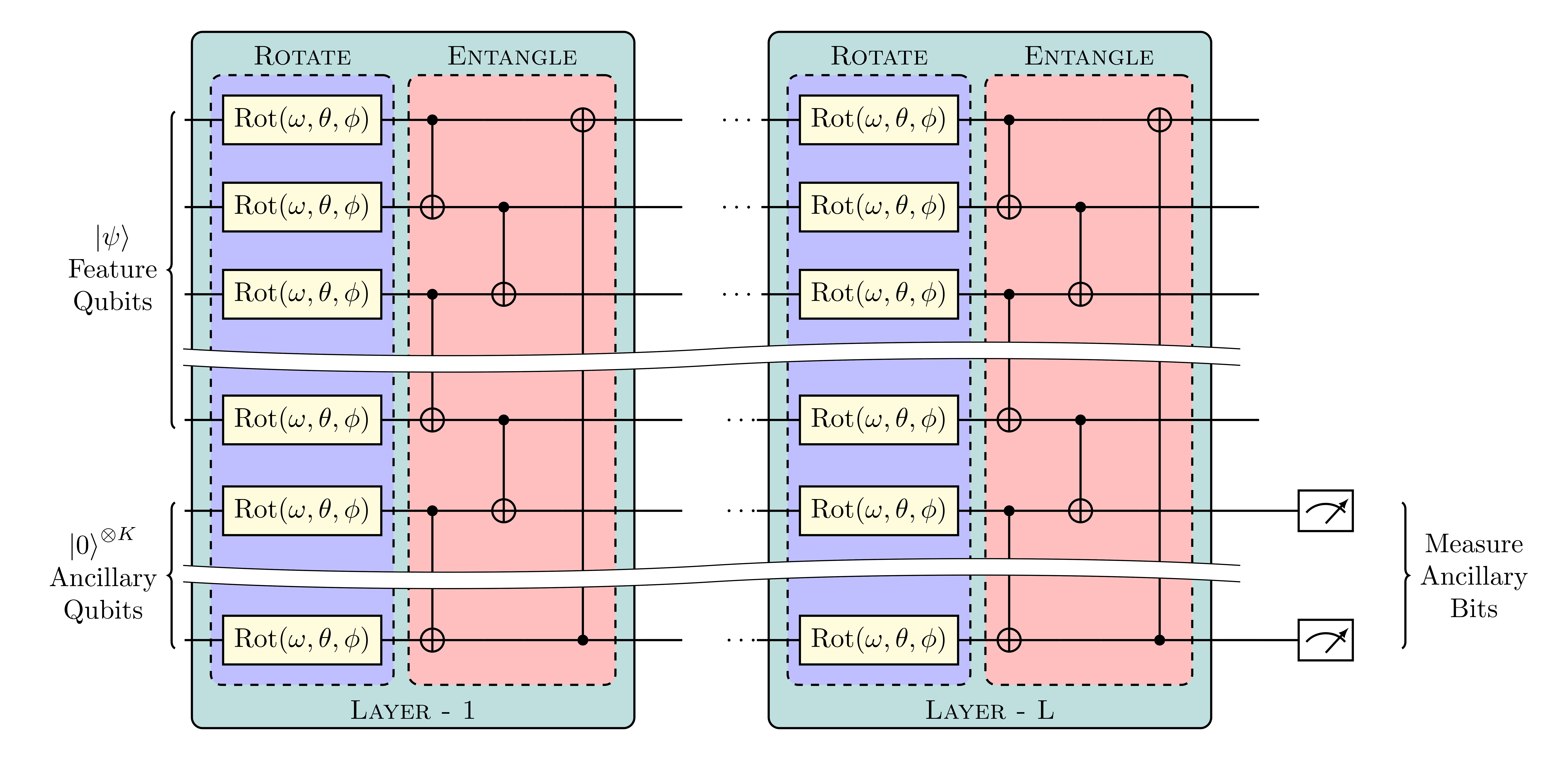}
    \caption{A quantum classifier with $L$ layers and alternating rotation and entanglement sub-units. Quantum data or encoded classical data is passed into the classifier and measurements of ancillary qubits give us the prediction probabilities.
    }
    \label{fig:qclass}
\end{figure}

After a series of rotations and entanglements are applied across the inputs, we carry out a number of circuit evaluations (shots) of the ancillary qubits to approximate the prediction probabilities of each of the $2^K$ possible classes. There are a total of $3L(D+K)$ trainable parameters in the model. The loss calculated across a batch of inputs is used to calculate gradients for all trainable parameters. It may be noted that, in our experiments, we simulate the quantum circuit and use backpropagation to calculate gradients. This is done so assuming end to end differentiability and the absence of quantum noise. In order to calculate gradients in a real quantum circuit, the analytical parameter-shift rule given below is used:
\begin{equation}
    \frac{\partial \langle\mathcal{L}(\Theta)\rangle}{\partial p} = \frac{1}{2}\Big( \langle\mathcal{L}(\Theta)\rangle_{p + \frac{\pi}{2}} - \langle\mathcal{L}(\Theta)\rangle_{p - \frac{\pi}{2}}\Big)
\end{equation}
where, $\langle\mathcal{L}(\Theta)\rangle_p$ denotes the expectation value of the cost function that depends on parameters $\Theta$, when a parameter $p$ is set to be a specific value. Simulating this, however takes significantly more time, since calculating gradients for each parameter using parameter shift requires atleast $2$ forward passes. 

Note that choosing an ansatz is a crucial task while setting up any quantum machine learning framework. In order to motivate our choice of ansatz, we tested different circuits for their expressibility similar to \cite{Azad_2023}. In order to do so, we generate a distribution of states by randomly sampling over the variational parameter space. Then we calculate the divergence in the distribution of states obtained so with the states obtained from the maximally expressive Haar distribution. The lesser the divergence, the more expressive power a circuit has. We tested a number of ansatzes before selecting the ansatz used in our experiments. 

\begin{figure}[htbp]
    \centering
    \includegraphics[width=0.9\columnwidth]{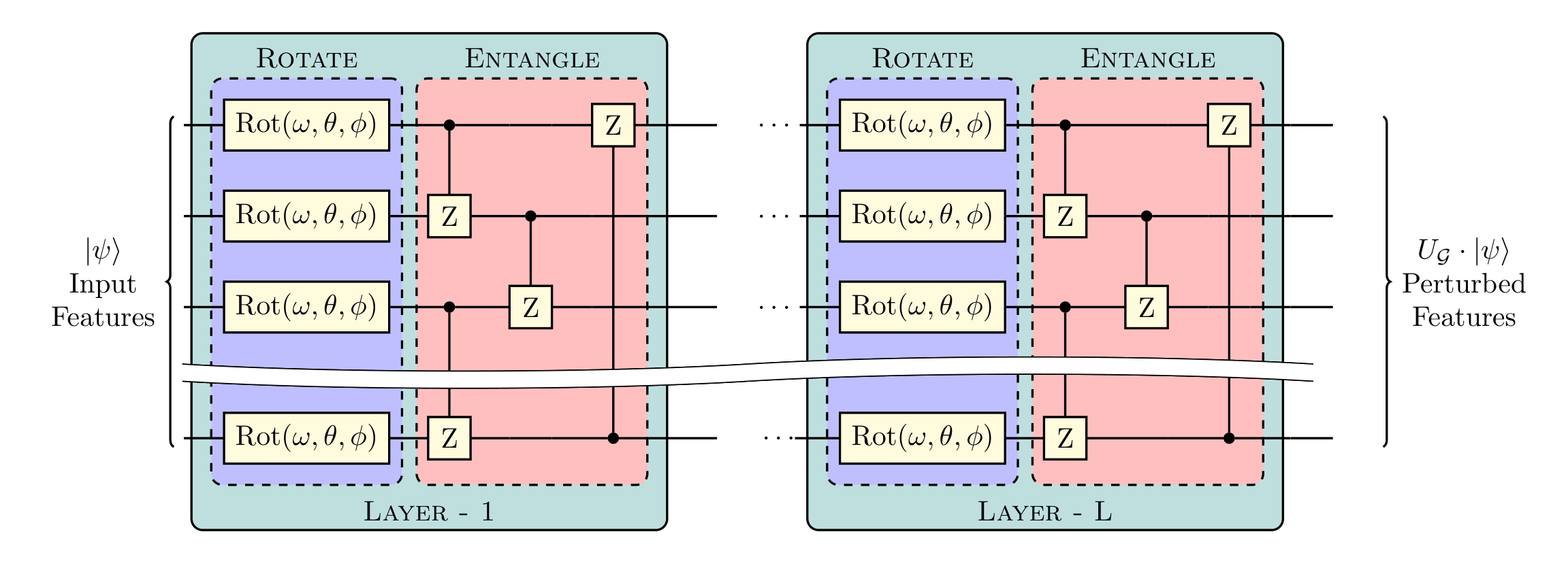}
    \caption{A quantum generator with $L$ layers acting on an initial quantum state $\ket{\psi}$. A unitary transformation $U_\mathcal{G}$ acts on it to give a perturbed quantum state $\ket{\phi}=U_\mathcal{G}\ket{\psi}$}
    \label{fig:qgen}
\end{figure}

The structure of a quantum generative model is similar to the structure discussed above with a few notable differences. The quantum generator has only $D=\lceil\log_2 d\rceil$ qubits. No ancillary bits are used in the generator. Further, we replace the CNOT gates in the classifier with CZ gates. The motivation for this is to ensure that setting all parameters to $0$ allows the global unitary transformation applied by the generator to be the identity transformation. The full circuit is shown in Figure \ref{fig:qgen}.

\subsection{QuGAP-A}

\subsubsection{Experimental setup}

All experiments are carried out in Python using Pytorch interfaced with Pennylane. We use $16 \times 16$ downsampled versions of MNIST and FMNIST; binary and four class classification is done. For MNIST binary classification, all images belonging to classes 0 and 1 of the original MNIST dataset are considered; for four class, images belonging to classes 0, 1, 2 and 3 are considered. Similarly for FMNIST, binary classification task involves images from classes T-shirt/Top and Trouser; for four class classification, images of classes T-shirt/Top, Trousers, Pullover and Dress are used. The choice of classical datasets was motivated as follows: MNIST was chosen due to its ubiquity in existing machine learning literature, both in the context of adversarial attacks as well as quantum machine learning; FMNIST was chosen as it offers a significantly more difficult classification task as compared to MNIST and can better benchmark machine learning methods \cite{xiao2017fashion}

We train PQCs of depth 10 and depth 20 for binary classification tasks and train PQCs of depth 20, 40 and 60 for the four class classification tasks. Apart from the PQCs, we also train a convolutional neural network (CNN) for comparison of attack performance as well as to study the transferability of attacks from classical to quantum classifiers, discussed in a later section. The CNN used has $3$ convolutional blocks with convolutional, max pooling and activation layers. This is followed by a series of dense layers before the output layer. It may be noted that all models perform poorly on FMNIST by a margin of around $5\%$ as compared to MNIST. Furthermore, quantum classifiers achieve competitive results in binary classification tasks, but are outperformed by the CNN in 4-class classification. 

\begin{figure}[htbp]
\centering
\begin{subfigure}{0.45\columnwidth}
\centering
\includegraphics[width=0.95\textwidth]{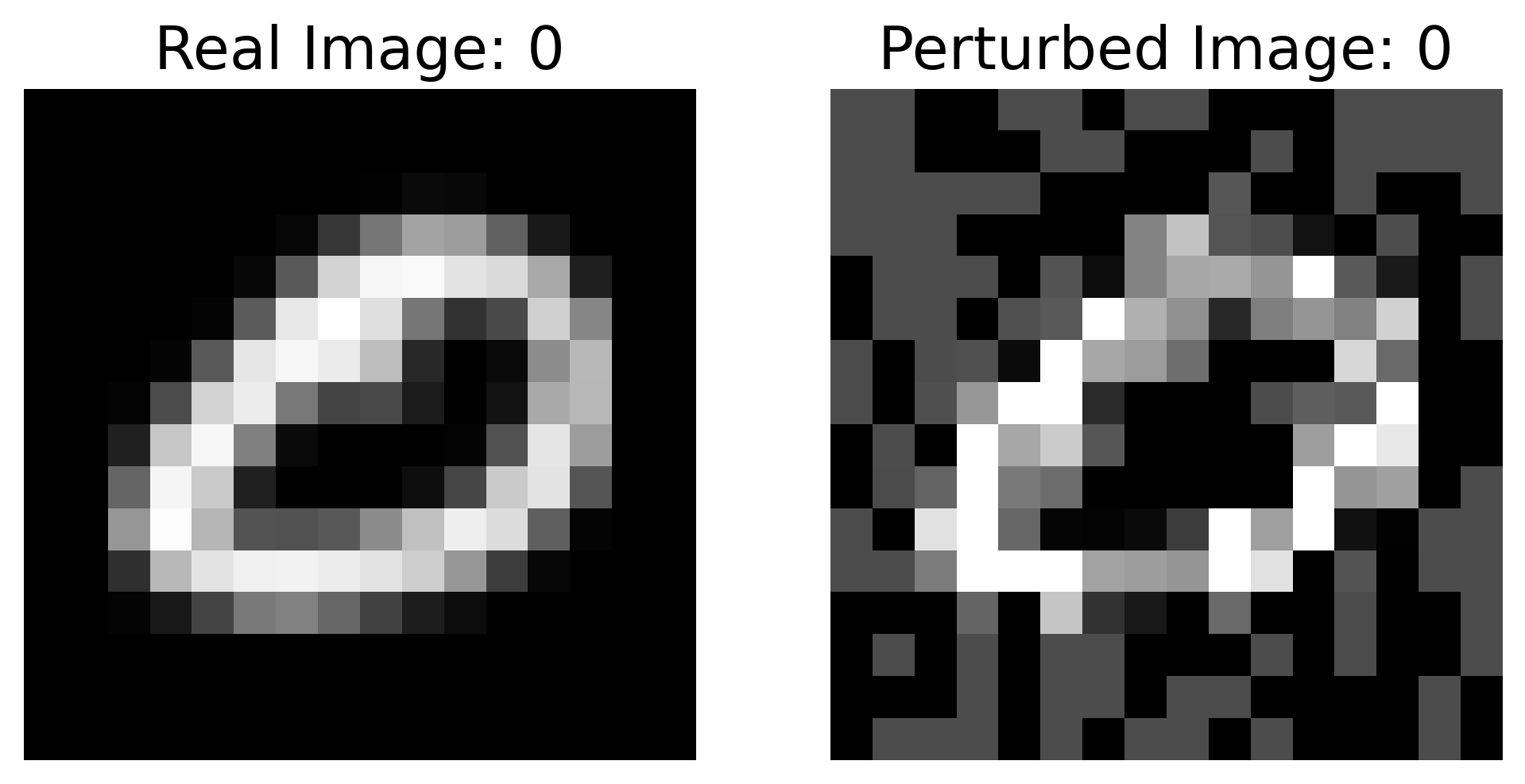}
\end{subfigure}
\begin{subfigure}{0.45\columnwidth}
\centering
\includegraphics[width=0.95\textwidth]{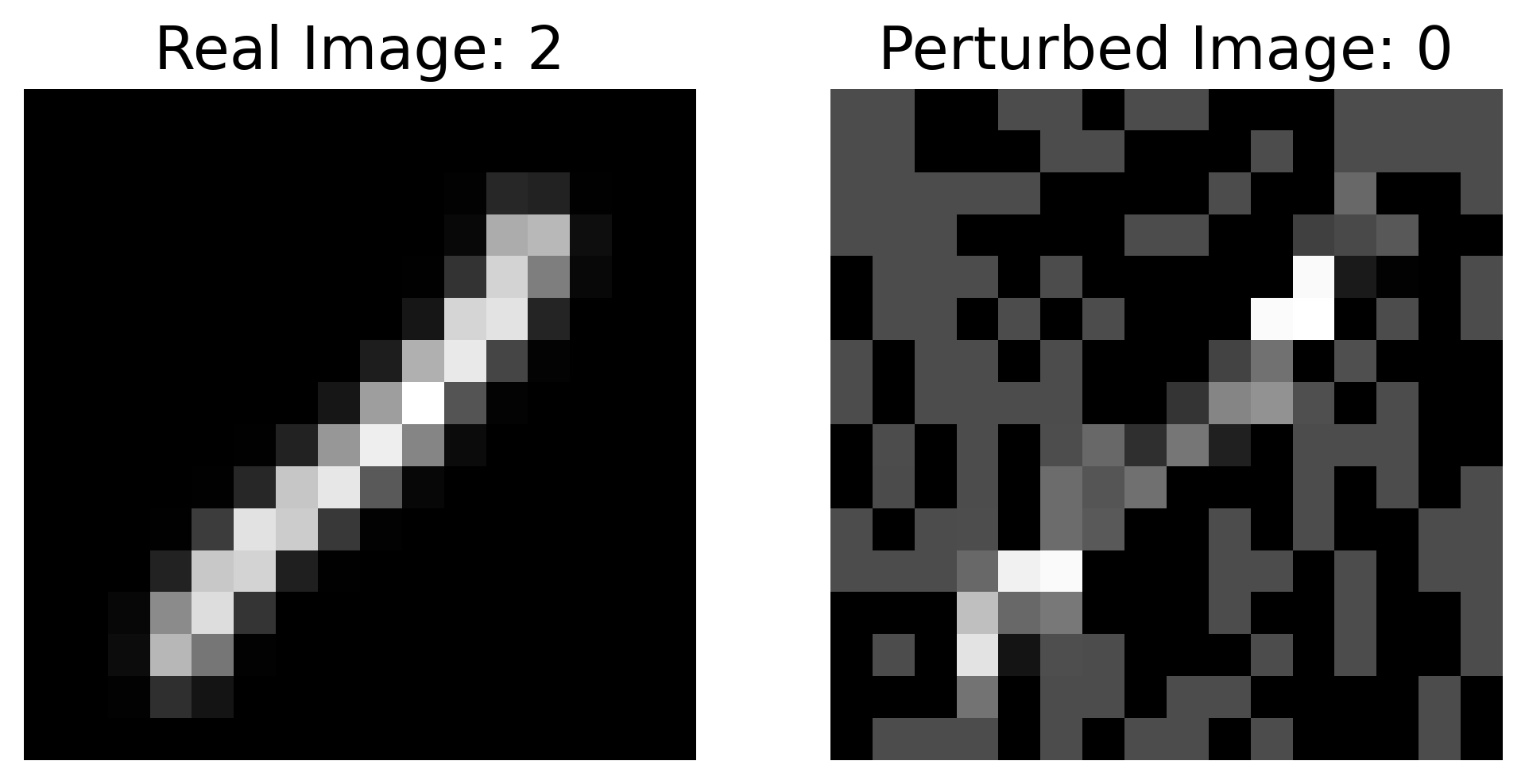}
\end{subfigure}
\begin{subfigure}{0.45\columnwidth}
\centering
\includegraphics[width=0.95\textwidth]{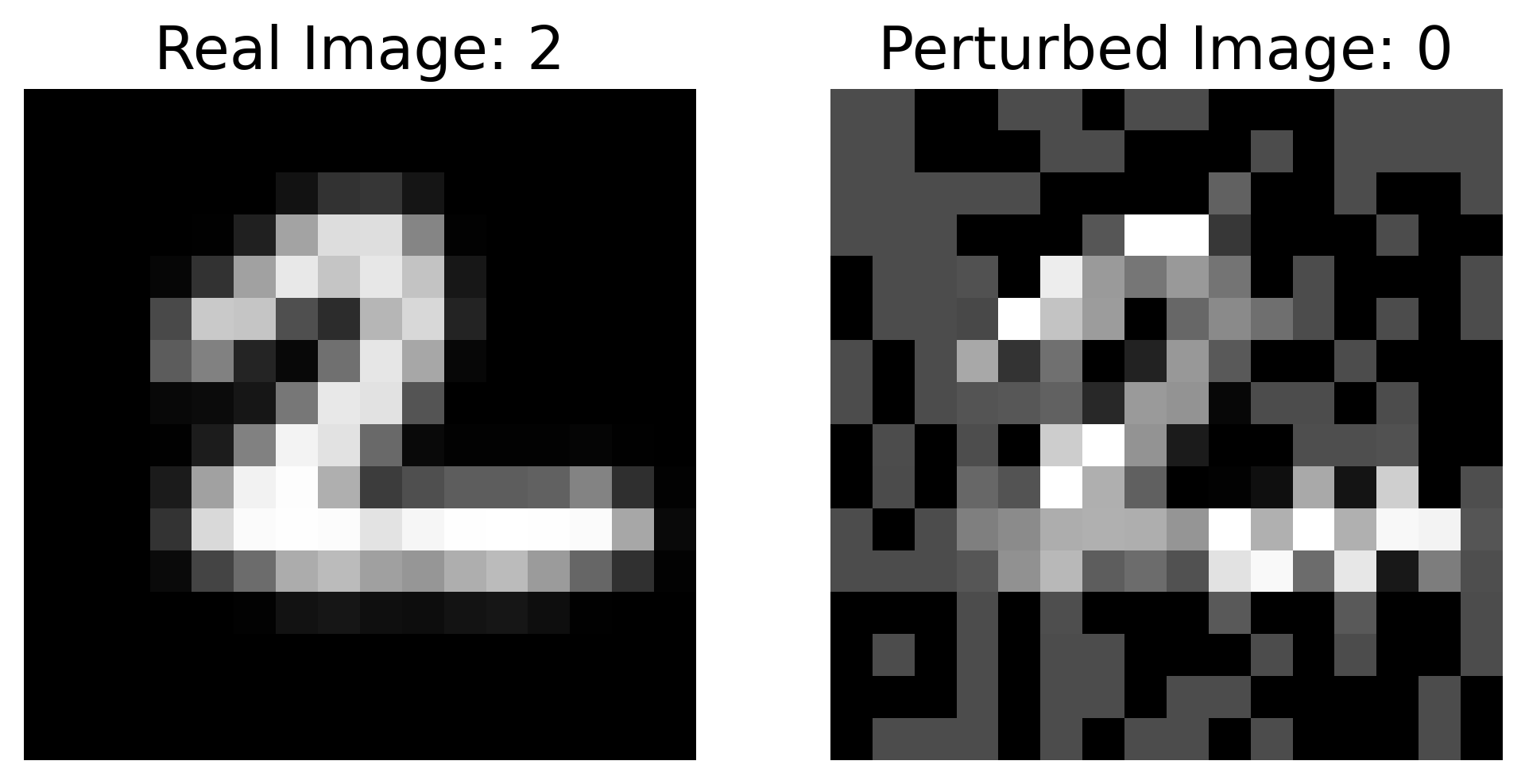}
\end{subfigure}
\begin{subfigure}{0.45\columnwidth}
\centering
\includegraphics[width=0.95\textwidth]{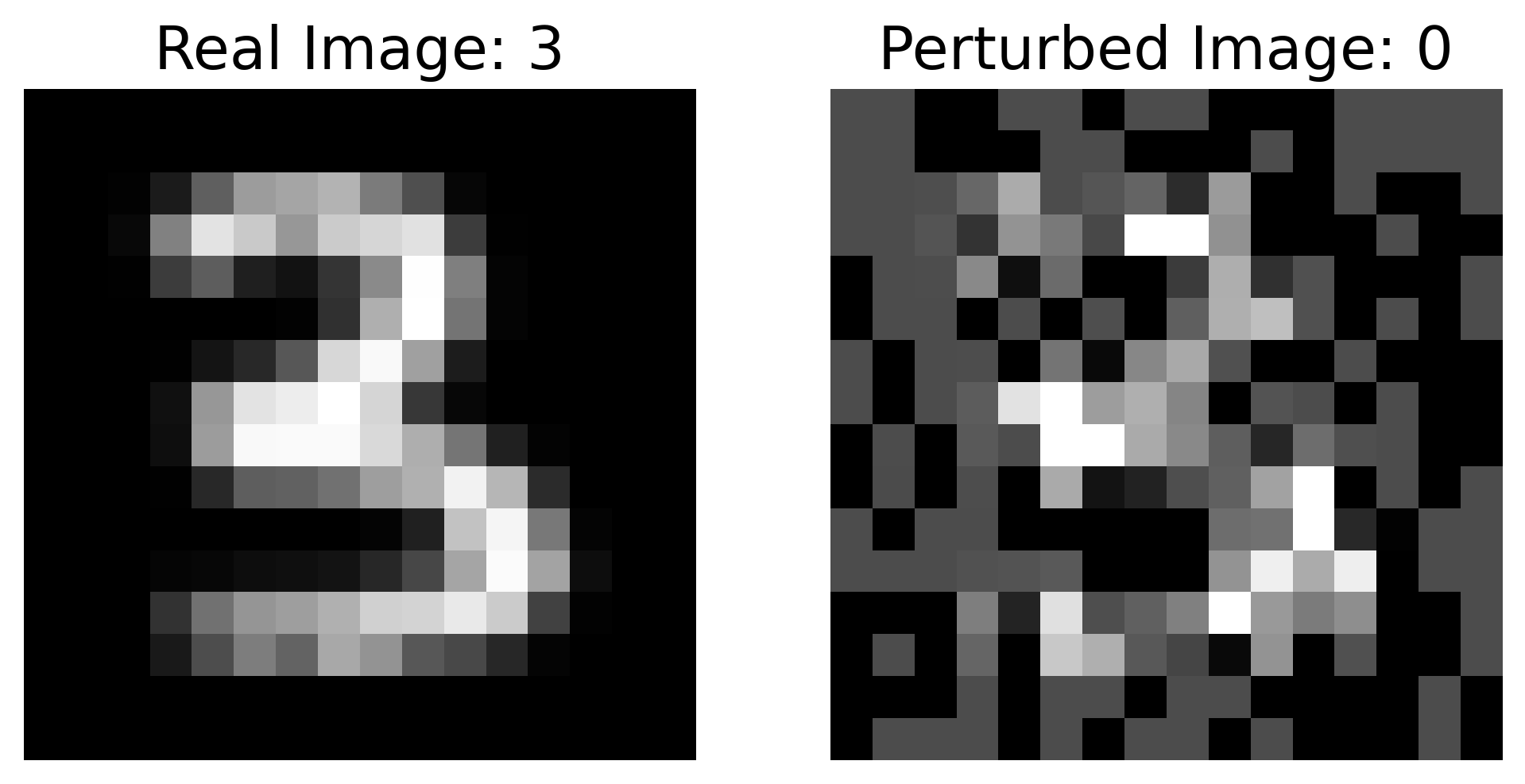}
\end{subfigure}
\caption{We show here an illustration of the generated untargeted UAP for PQC20 along with predictions for real and perturbed images. The strength of perturbation is $\epsilon=0.30$}
\label{fig:example}
\end{figure}

The quantum target models are trained for binary classification with a learning rate of $0.001$ for $5$ epochs and $0.0001$ for $5$ more epochs. They achieve approximately $100\%$ and $94\%$ accuracy respectively for MNIST and FMNIST. For 4 class classification, the quantum classifiers are trained for $30$ epochs with a learning rate of $0.001$ that decays by $0.1$ for $10$ epochs each and achieve approximately $94\%$ and $88\%$ accuracy for MNIST and FMNIST respectively. We use the Adam optimizer \cite{kingma2017adam} with a batch size of $64$ for training. 

The random vector $z$ is sampled from $\mathbb{R}^{256}$. For the classical generator, we use a fully-connected neural network with 3 hidden layers containing $512$, $1024$ and $512$ nodes each. Each layer uses a Leaky ReLU activation function except the output layer which uses the tanh activation function. It may be noted that we also experimented with deep convolutional generators, however results were unsatisfactory. The trained quantum classifiers are classically simulated by computing their equivalent unitary transformation $U$. Note that once you have $U$, the output probabilities can be computed using the result presented in Lemma 1 in the main paper. Backpropogation is carried out using autograd framework of Pytorch so as to speed up computation. The generator training is carried out using Adam optimizer for $10$ epochs, with a decaying learning initialized at $0.001$, decaying by a factor of $0.3$ every 4 epochs. Note that we set the momentum term of the Adam optimizer, $\beta_1=0.5$ for generator training.

\subsubsection{Clipping of images}

To ensure that the generated images after applying additive attacks are still valid, we clip each pixel of the adversarial image generated to the range $[0, 1]$ after normalizing. While we have not accounted for this step in the developed theory, the effect of this additional step is rather straightforward. Note that we first normalize the adversarial image so that it lies on a $(d-1)-$sphere of unit norm. The clipping step can then be interpreted as a projection into the closed orthant of space $\mathbb{R}^d$ defined by:
\begin{equation}
    \{x_i \geq 0\}_{i = 0}^{d-1} \label{eq:orthant}
\end{equation}
where $x_i$ denotes the $i^{\textrm{th}}$ element of vector $x$. The projected points are then re-normalized to lie on the $(d-1)-$sphere of unit norm. Therefore, this additional step essentially constraints the generated adversarial image to lie on the part of the $(d-1)-$sphere in the closed orthant defined by \eqref{eq:orthant}. While this introduces an additional constraint on the generated images, we still expect the generator to be able to compute a perturbation which can fool classifiers while adhering to this constraint. Our experimental results validate this claim.

\subsection{QuGAP-U: Classical Simulation}

As mentioned in the main paper, we simulate QuGAP-U framework classically to validate its viability. To perform the simulation, we represent the quantum data classically using complex vectors $x \in \mathbb{C}^d$. We also represent the trained quantum classifier $\mathcal{Q}$ using its equivalent unitary $U_{\mathcal{Q}}$. Instead of using a quantum generator, an arbitrary matrix $T \in \mathbb{C}^{d\times d}$ is used. However, note that $T$ has to be constrained to be a unitary to ensure that it is equivalent to a quantum generator. After transforming the inputs using $T$, the probabilities are computed. Loss is then computed and $T$ is optimized using gradient descent. The pseudocode is given in Algorithm \ref{alg:algorithm}. It may be noted that, the purpose of this simulation is to use gradient descent to classically identify unitary UAPs. This is done so because PQCs, by construction, cannot simulate any arbitrary transformation. In order to do so, we would need a PQC with infinite depth \cite{perez2021one, perez2020data}.
\begin{algorithm}[tb]
\caption{Classical simulation of QuGAP-U}
\label{alg:algorithm}
\textbf{Input}: Training set $\{ (x^{i}, c_{x^{i}}) \}_{i = 1}^{N}$, $U_{\mathcal{Q}}$, $T$\\
\textbf{Parameter}: Fidelity control $\alpha$, Learning rate $\eta$, Batches $B$, Total epochs $E$\\
\textbf{Output}: Trained unitary $T$
\begin{algorithmic}[1] 
\STATE Initialize $T$ randomly
\FOR{epoch in E}

 \FOR{batch in B}
  \STATE Compute SVD of $T$: $T = W\Sigma V^{\dagger}$
  \STATE $X \longleftarrow WV^{\dagger}$
  \STATE  $t^{i} \longleftarrow Tx^{i}$ for every $x^{i}$ in batch. 
  \STATE  $y^{i} \longleftarrow U_{\mathcal{Q}}t^{i}$ for every $t^{i}$ in batch.
  \STATE Compute prediction probabilities $p^{i}$ using $y^{i}$
  \STATE Calculate $\mathcal{L}_{fool}$ and $\mathcal{L}_{fid}$
  \STATE $L_U \longleftarrow L_{fool} + \alpha L_{fid}$
  \STATE Update $T$ using Adam; learning rate = $\eta$
 \ENDFOR
\ENDFOR
\end{algorithmic}
\end{algorithm}
A non-trivial part of the algorithm is handling the constraint that $T$ should be unitary. To achieve this, we utilize a projected gradient descent approach. The approach is based on the fact that the closest unitary matrix, $U_T$, to $T$ in terms of Frobenius norm ($\norm{T-U_T}_F$) is $U$ if $T = UP$ is the polar decomposition of $T$. We refer the readers to \cite{keller1975closest} for a proof. Furthermore, we note that $U = WV^{\dagger}$ if $T = W\Sigma V^{\dagger}$ is the singular value decomposition (SVD) of $T$ as shown in \cite{higham1986computing}. Based on these two facts, after each step of gradient descent on $T$, we project $T$ onto the space of unitaries by setting $T = WV^{\dagger}$.

Training is carried out for $15$ epochs with a batch size of $64$ using Adam optimizer with an initial learning rate of $0.001$ which decays by $0.3$ after every $5$ epochs. It may be noted that other optimizers like SGD were experimented with but discarded due to unsatisfactory results. Further, the value of the momentum parameter $\beta_1$, of the Adam optimizer, is set to an experimentally determined value of $0.5$.

\subsection{QuGAP-U}
All experiments are carried out in Python using Pytorch interfaced with Pennylane. We attack quantum classifiers of depth 10 and 20 (of the structure given in Figure \ref{fig:targ}) trained for binary classification on the TIM Dataset (described in the next section) as well as $8 \times 8$ downsampled MNIST. Furthermore, due to computational limitations, MNIST dataset was subsampled by a factor of 10. The classifiers achieve accuracies of approximately $92\%$ on the test set of TIM and $99\%$ on the test set of MNIST.  Quantum generators are of the structure described in Figure \ref{fig:qgen}. For the results reported in the main paper, we use quantum generators of depth 30 for TIM and depth 200 for MNIST. The choice of depth was based on the dimension $d$ of the Hilbert space of the dataset as described in the main paper. We choose a generator depth such that the total number of parameters is at least $d^2$. More precisely, we used the following relation to compute the depth: 
\begin{equation}
    \text{Depth} \geq \frac{d \times d}{3 \dot \lceil\log_2 d\rceil}
    \label{eq: depth}
\end{equation} We can see from this relation that the required depth increases very quickly as the dimensions of the input data increases. A motivation for choosing the depth so, can be found in section D4. Training is done by using `backprop' method in Pennylane library. Gradients computed are used to update the generator parameters using a learning rate of $0.001$. Training is done in batches of size $64$ for MNIST/FMNIST and $50$ for TIM dataset, over $10$ epochs. We use Adam optimizer with the momentum term set as $\beta_1=0.5$. 

\subsection{TIM Dataset}
The transverse-field Ising model (TIM) is a quantum version of the classical Ising model \cite{ising1967}, a mathematical model of ferromagnetism in statistical mechanics. It consists of a lattice, with nearest neighbour interactions determined by the alignment of spins along the $z$-axis as well as a transverse magnetic field applied along the $x$-axis. We consider a chain lattice. The hamiltonian for a system with spin $\pm \frac{1}{2}$ can be described as:
\begin{equation}
    H = -J\Big( \sum_i \sigma_{i}^z \sigma_{i+1}^z + g\;\sum_i \sigma_i^x\Big) \label{eq:hamilt}
\end{equation}
where $\sigma_i^z$ and $\sigma_i^x$ are the Pauli matrices for the $i^{th}$ spin, $J$ is an energy prefactor and $g$ is a coefficient that denotes the relative strength of the applied transverse magnetic field compared to the nearest neighbour interactions. It is known that the model is said to be in ordered state when $\abs{g}\leq 1$; for $J>0$ this is a ferromagnetic ordering and for $J<0$ it corresponds to an anti-ferromagnetic ordering of the lattice. Moreover, for $\abs{g}<0$, the system is said to be in disordered state, i.e. paramagnetic ordering.

In order to model a binary classification task based on the Ising model, we consider the problem of classifying the phase of the model from its quantum state. In order to set this up, we sample values of $g$ uniformly from the range $[0,2]$. The Hamiltonian of the system is then constructed using \eqref{eq:hamilt} and the ground state of the of system is calculated. We then construct data-label pairs, with the data being the quantum state, and the label descriptor, assigned to the system based on its phase. Class $0$ corresponds to paramagnetic (disordered) phase and class $1$ corresponds to ferromagnetic (ordered) phase. It may be noted that such a dataset can be constructed based on the system only because it has a distinct phase transition at $\abs{g}=1$. The synthesized dataset has $5000$ training images and $1000$ testing images, evenly split across both classes. This dataset is used in all our experiments. Note that the quantum state, can be represented using complex tensors for the purpose of simulating QuGAP-U classically as discussed in previous sections.

\section{Additional Experiments}

\subsection{Targeted Attacks}
\begin{figure}[h!tbp]
\centering
\begin{subfigure}{0.45\columnwidth}
\centering
\includegraphics[width=0.95\textwidth]{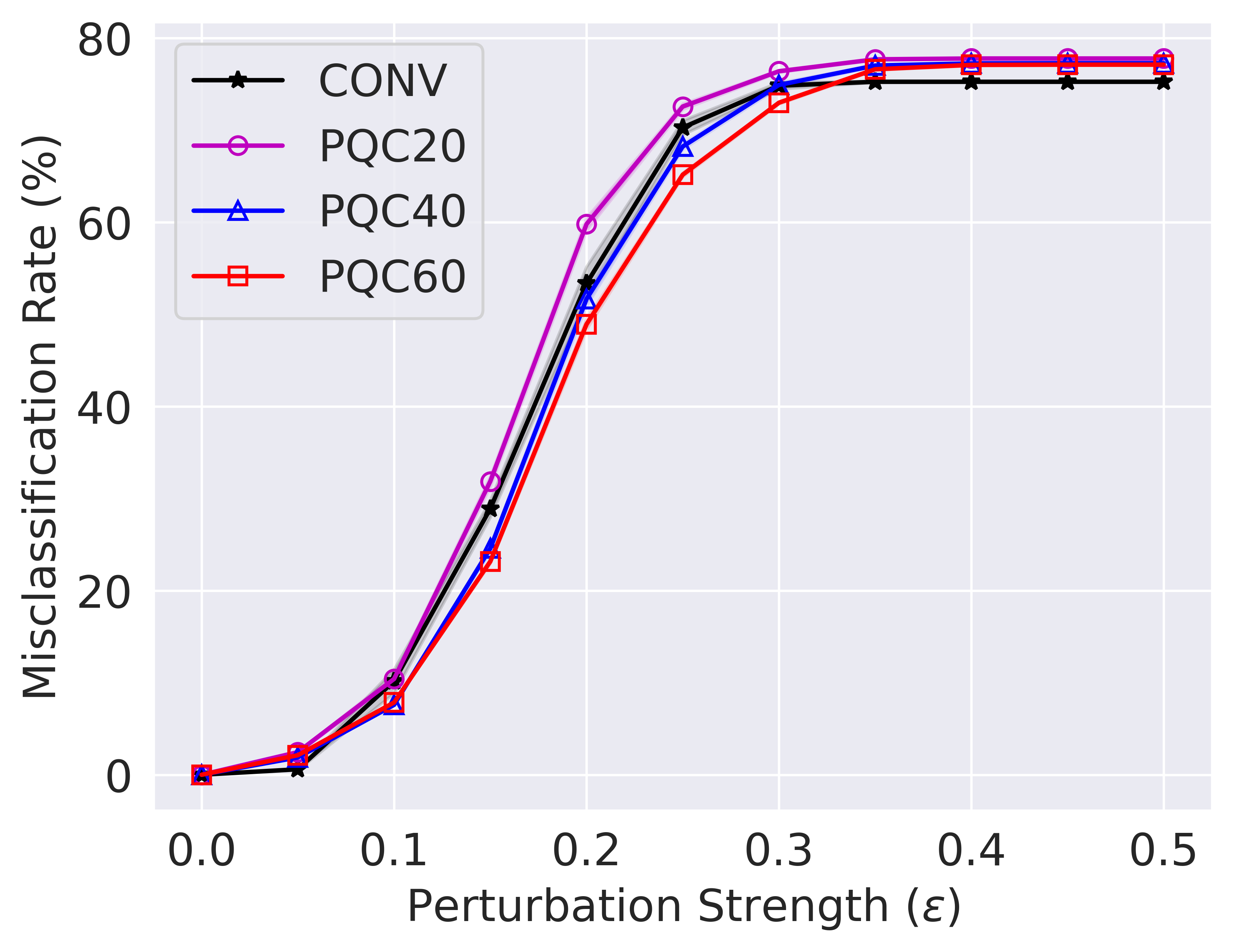}
\caption{MNIST: 4 Class}
\end{subfigure}
\begin{subfigure}{0.45\columnwidth}
\centering
\includegraphics[width=0.95\textwidth]{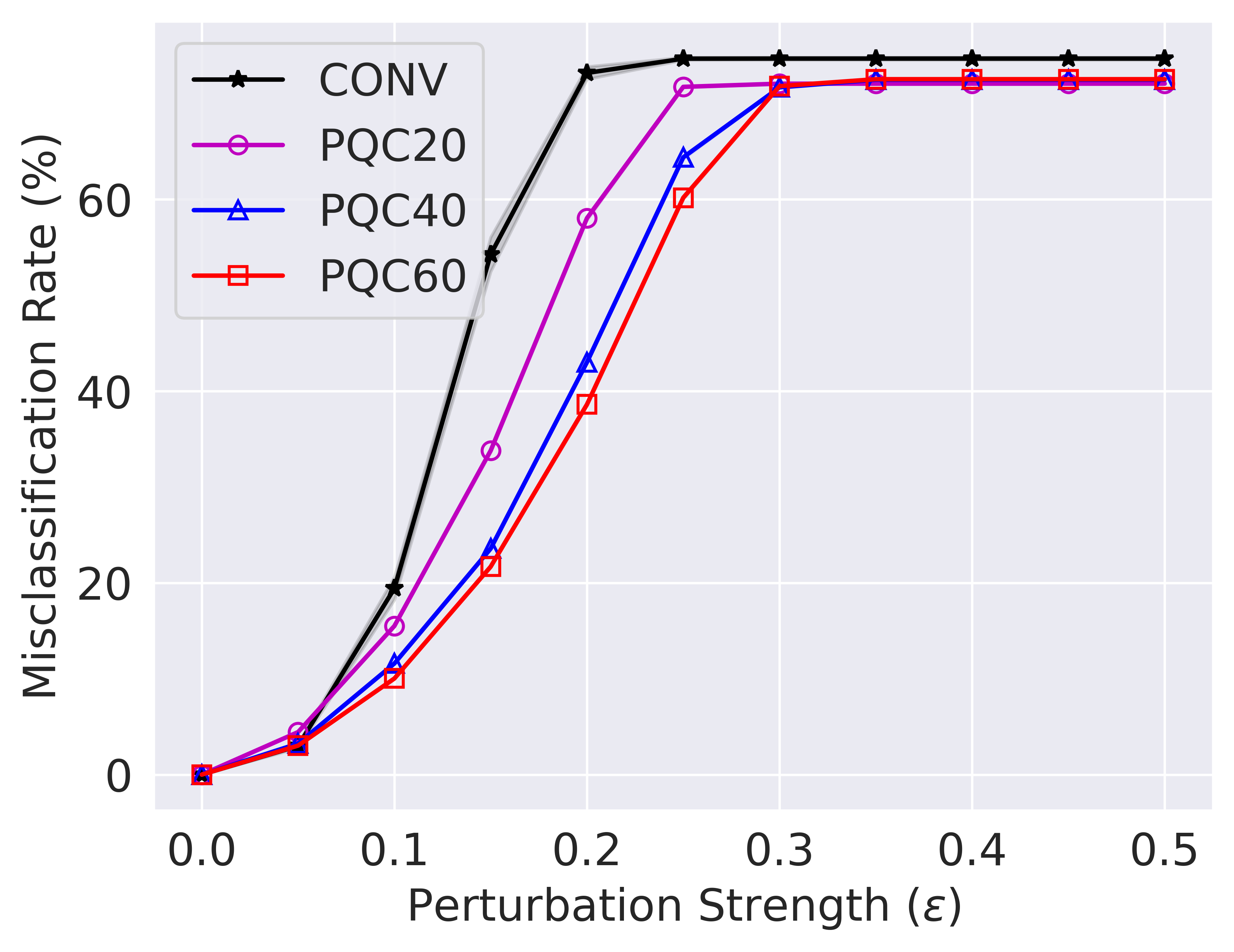}
\caption{FMNIST: 4 Class}
\end{subfigure}
\caption{Misclassification rates for targeted additive UAPs}
\label{fig:targ}
\end{figure}

In this section, we report results for targeted attacks on $16 \times 16$ MNIST and FMNIST four class classification. It may be noted that all training settings were identical to the untargeted case. As with the untargeted attacks, we generate ten different attacks and report mean and standard deviation of misclassification rates for these attacks on the test set. As the target class, we choose class 2 for MNIST and class Trouser for FMNIST. The results are given in Figure \ref{fig:targ}. We observe high misclassification rates competitive to untargeted attacks, thereby validating the efficacy of our proposed approach.

\subsection{Transferability}

As discussed in the main paper, a naive approach for generating additive UAPs for quantum classifiers would be to generate additive UAPs for classical classifiers using existing methods and transferring them to quantum classifiers. However, here we experimentally show that attacks on classical classifiers do not transfer to quantum classifiers.
\begin{figure}[htbp]
\centering
\begin{subfigure}{0.45\columnwidth}
\centering
\includegraphics[width=0.95\textwidth]{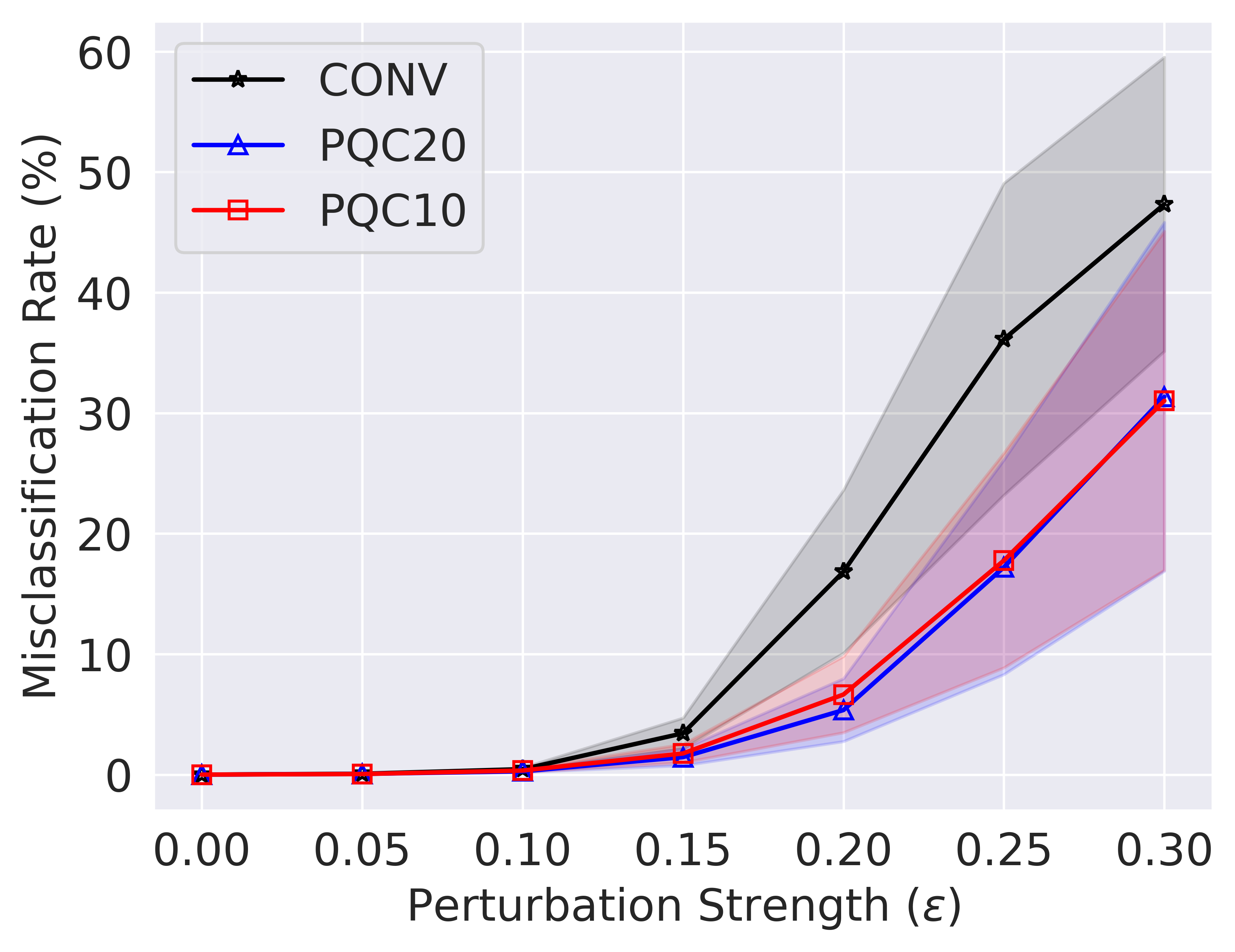}
\caption{MNIST: 2 Class}
\end{subfigure}
\begin{subfigure}{0.45\columnwidth}
\centering
\includegraphics[width=0.95\textwidth]{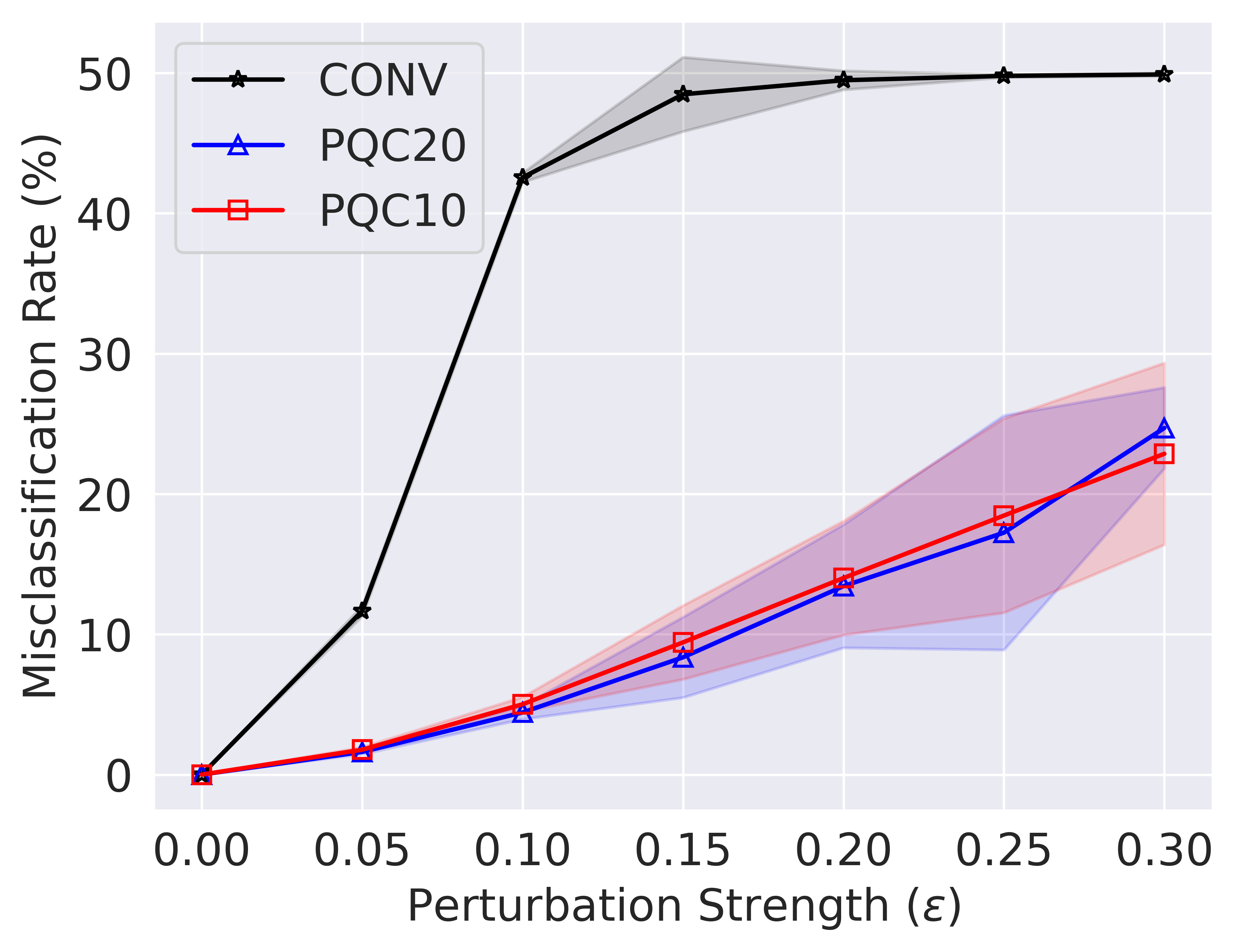}
\caption{FMNIST: 2 Class}
\end{subfigure}
\begin{subfigure}{0.45\columnwidth}
\centering
\includegraphics[width=0.95\textwidth]{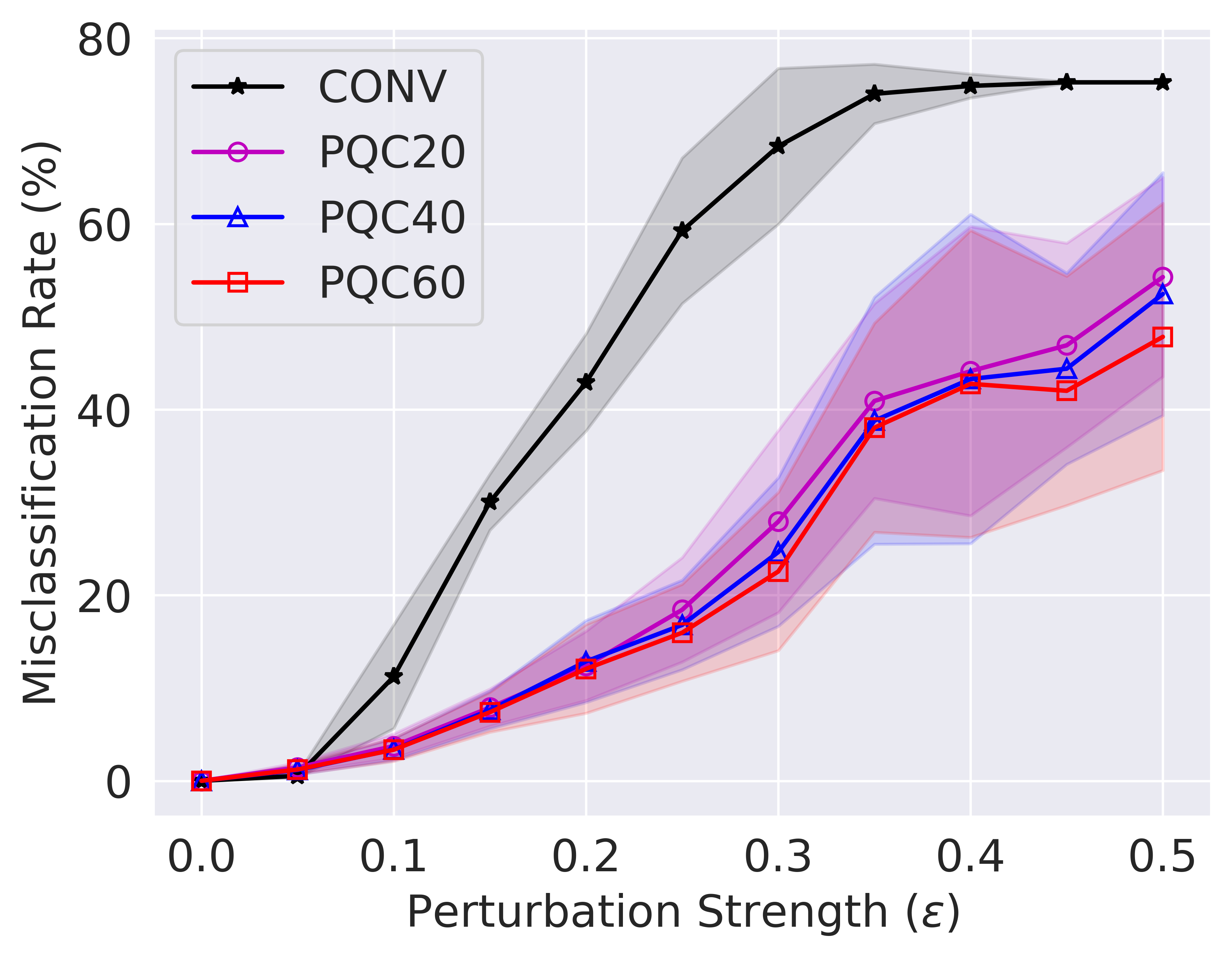}
\caption{MNIST: 4 Class}
\end{subfigure}
\begin{subfigure}{0.45\columnwidth}
\centering
\includegraphics[width=0.95\textwidth]{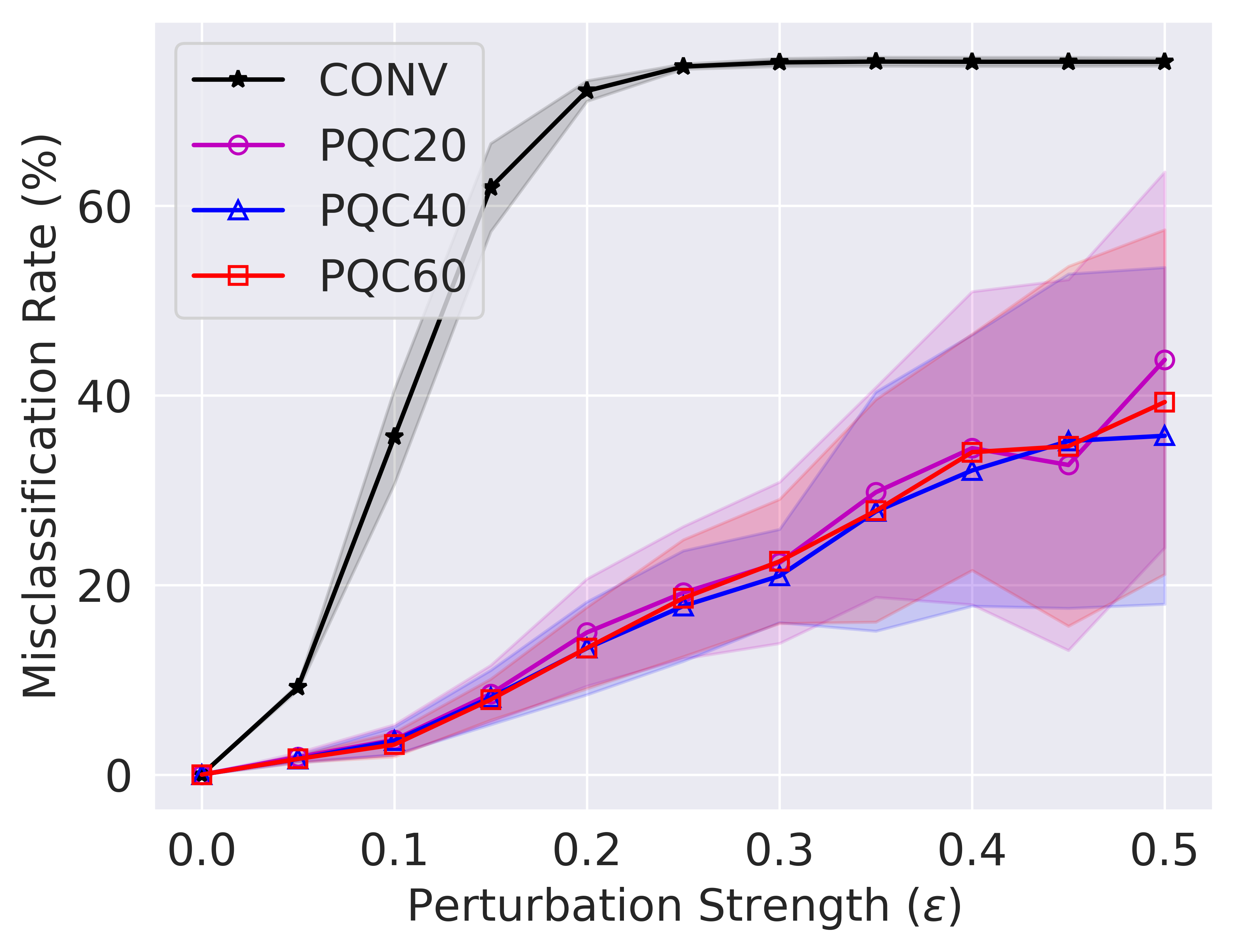}
\caption{FMNIST: 4 Class}
\end{subfigure}
\caption{Transferability of additive UAPs. We report transferability for attacks trained on CONV for binary and four class classification for $16\times16$ MNIST and FMNIST. Attacks show very limited transferability, if at all, from classical to quantum classifiers.}
\label{fig:transf}
\end{figure}
To show this, we first generate attacks on the CNN classifier trained on MNIST and FMNIST for binary and four-class classification tasks. After generating 10 such attacks, we test the effectiveness if these attacks on the quantum classifiers. More specifically, we compute the mean and standard deviation of the misclassification rates produced by these attacks on the quantum classifiers. The results are reported in Figure \ref{fig:transf}. We observe attacks generated by training on the CONV network do not transfer to the PQCs. This is especially pronounced in the more complex FMNIST dataset. Therefore, we require a dedicated framework for generating additive UAPs for quantum classifiers. This provides an additional motivation for using the proposed QuGAP-A framework.

\subsection{qBIM: Variation of fidelity with depth}
We utilize the adaptation of qBIM to generate unitary UAPs proposed in \cite{gong2022universal}. A single variational layer is used for generating the attack. The strength of the perturbation is controlled by constraining the variational layer to lie near the identity operator. In practice, this is ensured by clamping the parameters of the rotation gates in $[-\epsilon, \epsilon]$. By increasing $\epsilon$, we trade-off fidelity for higher misclassification. 

However, having only a single layer severely limits the search space for the perturbation. A naive approach to circumvent this problem would be to add multiple variational layers, but still constrain the parameters of each layer to be near zero, so that the overall perturbation remains close to identity. We experimentally evaluate this approach by varying the number of variational layers used in qBIM. 

\begin{figure}[htbp]
    \centering
    \includegraphics[width=0.8\columnwidth]{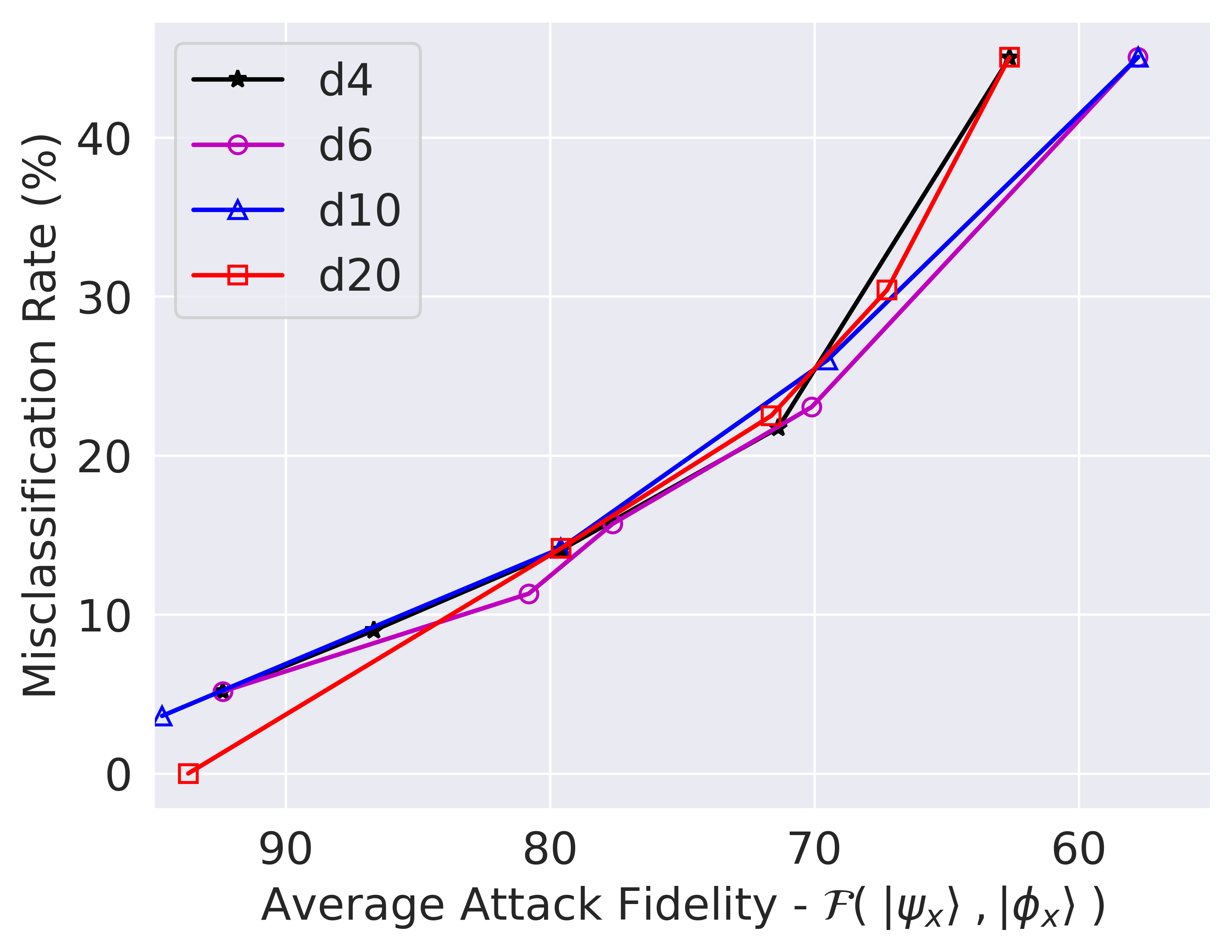}
    \caption{Misclassification rates for qBIM on TIM binary classification with various depths of variational layers: 4, 6 10 and 20 layers.}
    \label{fig:qbim}
\end{figure}

The results are presented in Figure \ref{fig:qbim}. We observe that despite increasing depth, the misclassification rate does not improve at the same fidelity. This is because though the depth is being increased, the constraint on each individual parameter has to be decreased to maintain the same fidelity. This makes the approach ineffective with respect to scaling and provides motivation for the proposed QuGAP-U framework. 

\subsection{Effect of generator depth on QuGAP-U}
As discussed in the main paper, a quantum generator of insufficient depth may be unable to generate effect unitary UAPs. We experimentally vary the depth of the generator and observe the impact on the performance of generated UAPs. Towards this, we use generators of varying depth to attack the PQC10 classifier trained on the $8\times8$ MNIST dataset (chosen because TIM dataset shows performance saturation at very low generator depths).
\begin{figure}[htbp]
    \centering
    \includegraphics[width=0.8\columnwidth]{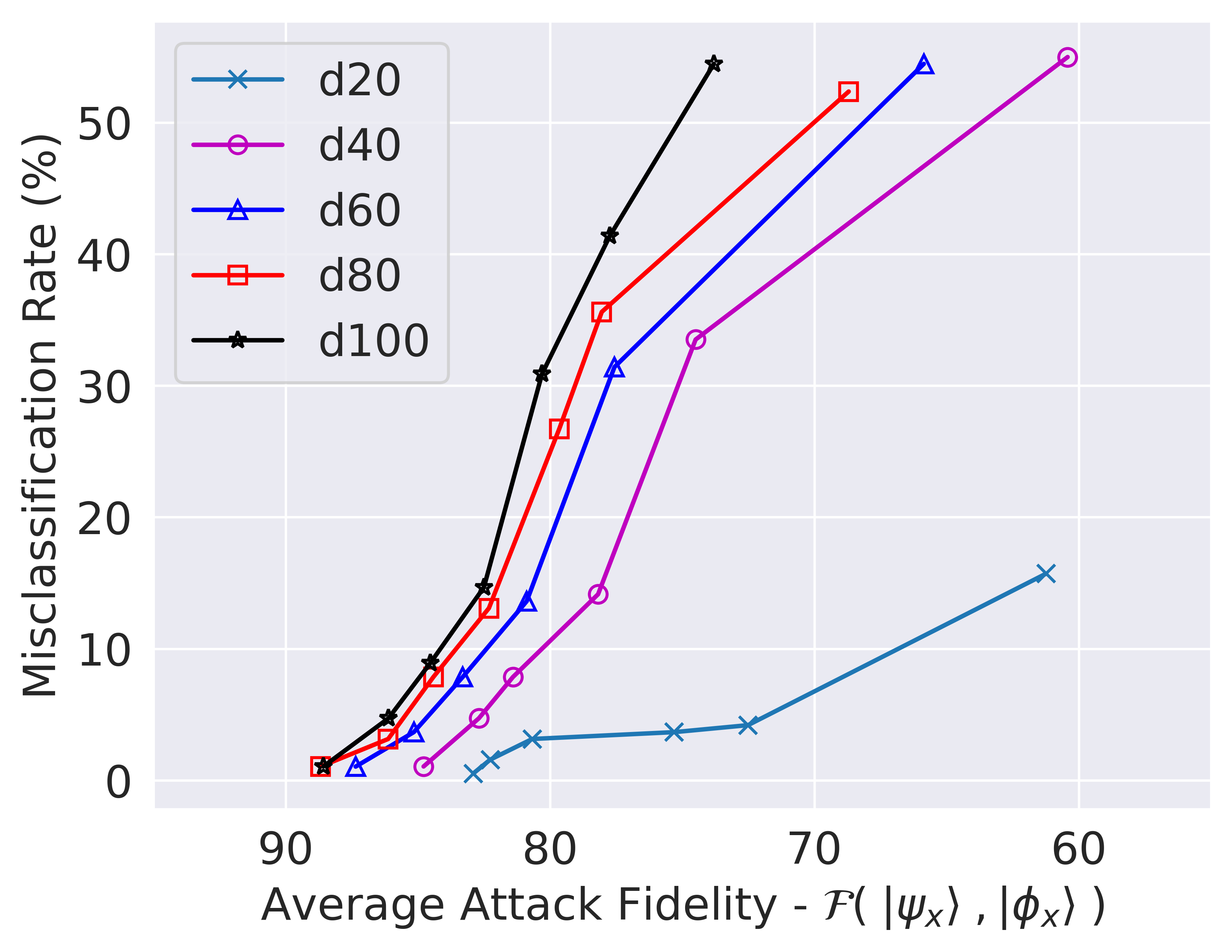}
    \caption{Misclassification rates for QuGAP-U trained using generators of various depths. Generative models of depths 20, 40, 60, 80 and 100 have been studied.}
    \label{fig:qugapu}
\end{figure}

The results are presented in Figure \ref{fig:qugapu}. We observe that the performance of the generator increases with increase in depth as we expect. Furthermore, the performance improvement is especially significant over lower depths. This further motivates our choice of depth to have $\mathcal{O}(d^2)$ parameters in the PQC. 

\subsection{Computational Resources}
The training of all classical models and simulations of quantum models was carried out on two NVIDIA RTX A$6000$ GPUs with VRAM of $48$GiB each. The training of all quantum classifiers and generative models was carried out on a cluster of $32$ Intel Xeon Gold $6226$R CPU cores. Due to the high memory requirements of the `backprop' gradient computation method in Pennylane, it is recommended to conduct all experiments described in this paper on machines with a RAM of at least $20$GB.

\bibliography{references}

\begin{thebibliography}{57}
\providecommand{\natexlab}[1]{#1}

\bibitem[{Abbas et~al.(2021)Abbas, Sutter, Zoufal, Lucchi, Figalli, and Woerner}]{abbas2021power}
Abbas, A.; Sutter, D.; Zoufal, C.; Lucchi, A.; Figalli, A.; and Woerner, S. 2021.
\newblock The power of quantum neural networks.
\newblock \emph{Nature Computational Science}, 1(6): 403–409.

\bibitem[{Akhtar and Mian(2018)}]{akhtar2018threat}
Akhtar, N.; and Mian, A. 2018.
\newblock Threat of Adversarial Attacks on Deep Learning in Computer Vision: A Survey.
\newblock arXiv:1801.00553.

\bibitem[{Azad and Sinha(2023)}]{Azad_2023}
Azad, U.; and Sinha, A. 2023.
\newblock {qLEET}: visualizing loss landscapes, expressibility, entangling power and training trajectories for parameterized quantum circuits.
\newblock \emph{Quantum Information Processing}, 22(6).

\bibitem[{Bai et~al.(2021)Bai, Zhao, Zhu, Han, Chen, Li, and Kot}]{bai2021ai}
Bai, T.; Zhao, J.; Zhu, J.; Han, S.; Chen, J.; Li, B.; and Kot, A. 2021.
\newblock Ai-gan: Attack-inspired generation of adversarial examples.
\newblock In \emph{2021 IEEE International Conference on Image Processing (ICIP)}, 2543--2547. IEEE.

\bibitem[{Benedetti et~al.(2019)Benedetti, Lloyd, Sack, and Fiorentini}]{benedetti2019parameterized}
Benedetti, M.; Lloyd, E.; Sack, S.; and Fiorentini, M. 2019.
\newblock Parameterized quantum circuits as machine learning models.
\newblock \emph{Quantum Science and Technology}, 4(4): 043001.

\bibitem[{Bergholm et~al.(2022)Bergholm, Izaac, Schuld, Gogolin, Ahmed, Ajith, Alam, Alonso-Linaje, AkashNarayanan, Asadi, Arrazola, Azad, Banning, Blank, Bromley, Cordier, Ceroni, Delgado, Matteo, Dusko, Garg, Guala, Hayes, Hill, Ijaz, Isacsson, Ittah, Jahangiri, Jain, Jiang, Khandelwal, Kottmann, Lang, Lee, Loke, Lowe, McKiernan, Meyer, Montañez-Barrera, Moyard, Niu, O'Riordan, Oud, Panigrahi, Park, Polatajko, Quesada, Roberts, Sá, Schoch, Shi, Shu, Sim, Singh, Strandberg, Soni, Száva, Thabet, Vargas-Hernández, Vincent, Vitucci, Weber, Wierichs, Wiersema, Willmann, Wong, Zhang, and Killoran}]{bergholm2022pennylane}
Bergholm, V.; Izaac, J.; Schuld, M.; Gogolin, C.; Ahmed, S.; Ajith, V.; Alam, M.~S.; Alonso-Linaje, G.; AkashNarayanan, B.; Asadi, A.; Arrazola, J.~M.; Azad, U.; Banning, S.; Blank, C.; Bromley, T.~R.; Cordier, B.~A.; Ceroni, J.; Delgado, A.; Matteo, O.~D.; Dusko, A.; Garg, T.; Guala, D.; Hayes, A.; Hill, R.; Ijaz, A.; Isacsson, T.; Ittah, D.; Jahangiri, S.; Jain, P.; Jiang, E.; Khandelwal, A.; Kottmann, K.; Lang, R.~A.; Lee, C.; Loke, T.; Lowe, A.; McKiernan, K.; Meyer, J.~J.; Montañez-Barrera, J.~A.; Moyard, R.; Niu, Z.; O'Riordan, L.~J.; Oud, S.; Panigrahi, A.; Park, C.-Y.; Polatajko, D.; Quesada, N.; Roberts, C.; Sá, N.; Schoch, I.; Shi, B.; Shu, S.; Sim, S.; Singh, A.; Strandberg, I.; Soni, J.; Száva, A.; Thabet, S.; Vargas-Hernández, R.~A.; Vincent, T.; Vitucci, N.; Weber, M.; Wierichs, D.; Wiersema, R.; Willmann, M.; Wong, V.; Zhang, S.; and Killoran, N. 2022.
\newblock PennyLane: Automatic differentiation of hybrid quantum-classical computations.
\newblock arXiv:1811.04968.

\bibitem[{Bharti et~al.(2022)Bharti, Cervera-Lierta, Kyaw, Haug, Alperin-Lea, Anand, Degroote, Heimonen, Kottmann, Menke et~al.}]{bharti2022noisy}
Bharti, K.; Cervera-Lierta, A.; Kyaw, T.~H.; Haug, T.; Alperin-Lea, S.; Anand, A.; Degroote, M.; Heimonen, H.; Kottmann, J.~S.; Menke, T.; et~al. 2022.
\newblock Noisy intermediate-scale quantum algorithms.
\newblock \emph{Reviews of Modern Physics}, 94(1): 015004.

\bibitem[{Biamonte et~al.(2017)Biamonte, Wittek, Pancotti, Rebentrost, Wiebe, and Lloyd}]{biamonte2017quantum}
Biamonte, J.; Wittek, P.; Pancotti, N.; Rebentrost, P.; Wiebe, N.; and Lloyd, S. 2017.
\newblock Quantum machine learning.
\newblock \emph{Nature}, 549(7671): 195–202.

\bibitem[{BRUSH(1967)}]{ising1967}
BRUSH, S.~G. 1967.
\newblock History of the Lenz-Ising Model.
\newblock \emph{Rev. Mod. Phys.}, 39: 883--893.

\bibitem[{Cao et~al.(2019)Cao, Romero, Olson, Degroote, Johnson, Kieferov{\'a}, Kivlichan, Menke, Peropadre, Sawaya et~al.}]{cao2019quantum}
Cao, Y.; Romero, J.; Olson, J.~P.; Degroote, M.; Johnson, P.~D.; Kieferov{\'a}, M.; Kivlichan, I.~D.; Menke, T.; Peropadre, B.; Sawaya, N.~P.; et~al. 2019.
\newblock Quantum chemistry in the age of quantum computing.
\newblock \emph{Chemical reviews}, 119(19): 10856--10915.

\bibitem[{Carlini and Wagner(2017)}]{carlini2017towards}
Carlini, N.; and Wagner, D. 2017.
\newblock Towards evaluating the robustness of neural networks.
\newblock In \emph{2017 IEEE Symposium on Security and Privacy (sp)}, 39--57. IEEE.

\bibitem[{Cerezo et~al.(2021)Cerezo, Arrasmith, Babbush, Benjamin, Endo, Fujii, McClean, Mitarai, Yuan, Cincio, and Coles}]{cerezo2021variational}
Cerezo, M.; Arrasmith, A.; Babbush, R.; Benjamin, S.~C.; Endo, S.; Fujii, K.; McClean, J.~R.; Mitarai, K.; Yuan, X.; Cincio, L.; and Coles, P.~J. 2021.
\newblock Variational quantum algorithms.
\newblock \emph{Nature Reviews Physics}, 3(9): 625–644.

\bibitem[{Dallaire-Demers and Killoran(2018)}]{dallaire2018quantum}
Dallaire-Demers, P.-L.; and Killoran, N. 2018.
\newblock Quantum generative adversarial networks.
\newblock \emph{Physical Review A}, 98(1).

\bibitem[{Du et~al.(2021)Du, Hsieh, Liu, Tao, and Liu}]{du2021quantum}
Du, Y.; Hsieh, M.-H.; Liu, T.; Tao, D.; and Liu, N. 2021.
\newblock Quantum noise protects quantum classifiers against adversaries.
\newblock \emph{Physical Review Research}, 3(2): 023153.

\bibitem[{Gong and Deng(2022)}]{gong2022universal}
Gong, W.; and Deng, D.-L. 2022.
\newblock Universal adversarial examples and perturbations for quantum classifiers.
\newblock \emph{National Science Review}, 9(6): nwab130.

\bibitem[{Gong et~al.(2022)Gong, Yuan, Li, and Deng}]{gong2022enhancing}
Gong, W.; Yuan, D.; Li, W.; and Deng, D.-L. 2022.
\newblock Enhancing Quantum Adversarial Robustness by Randomized Encodings.
\newblock arXiv:2212.02531.

\bibitem[{Goodfellow, Shlens, and Szegedy(2015)}]{goodfellow2015explaining}
Goodfellow, I.~J.; Shlens, J.; and Szegedy, C. 2015.
\newblock Explaining and Harnessing Adversarial Examples.
\newblock arXiv:1412.6572.

\bibitem[{He et~al.(2015)He, Zhang, Ren, and Sun}]{he2015delving}
He, K.; Zhang, X.; Ren, S.; and Sun, J. 2015.
\newblock Delving deep into rectifiers: Surpassing human-level performance on imagenet classification.
\newblock In \emph{Proceedings of the IEEE international conference on computer vision}, 1026--1034.

\bibitem[{Higham(1986)}]{higham1986computing}
Higham, N.~J. 1986.
\newblock Computing the polar decomposition—with applications.
\newblock \emph{SIAM Journal on Scientific and Statistical Computing}, 7(4): 1160--1174.

\bibitem[{Huang et~al.(2021)Huang, Broughton, Mohseni, Babbush, Boixo, Neven, and McClean}]{Huang2021}
Huang, H.-Y.; Broughton, M.; Mohseni, M.; Babbush, R.; Boixo, S.; Neven, H.; and McClean, J.~R. 2021.
\newblock Power of data in quantum machine learning.
\newblock \emph{Nature Communications}, 12(1): 2631.

\bibitem[{Huang et~al.(2023)Huang, Tsai, Yang, Su, Yu, Chen, and Kuo}]{huang2023certified}
Huang, J.-C.; Tsai, Y.-L.; Yang, C.-H.~H.; Su, C.-F.; Yu, C.-M.; Chen, P.-Y.; and Kuo, S.-Y. 2023.
\newblock Certified Robustness of Quantum Classifiers against Adversarial Examples through Quantum Noise.
\newblock In \emph{ICASSP 2023-2023 IEEE International Conference on Acoustics, Speech and Signal Processing (ICASSP)}, 1--5. IEEE.

\bibitem[{Jordan and Mitchell(2015)}]{jordan2015machine}
Jordan, M.~I.; and Mitchell, T.~M. 2015.
\newblock Machine learning: Trends, perspectives, and prospects.
\newblock \emph{Science}, 349(6245): 255--260.

\bibitem[{Keller(1975)}]{keller1975closest}
Keller, J.~B. 1975.
\newblock Closest unitary, orthogonal and hermitian operators to a given operator.
\newblock \emph{Mathematics Magazine}, 48(4): 192--197.

\bibitem[{Kiani et~al.(2022)Kiani, Balestriero, LeCun, and Lloyd}]{kiani2022projunn}
Kiani, B.; Balestriero, R.; LeCun, Y.; and Lloyd, S. 2022.
\newblock projUNN: Efficient method for training deep networks with unitary matrices.
\newblock \emph{Advances in Neural Information Processing Systems}, 35: 14448--14463.

\bibitem[{Kingma and Ba(2017)}]{kingma2017adam}
Kingma, D.~P.; and Ba, J. 2017.
\newblock Adam: A Method for Stochastic Optimization.
\newblock arXiv:1412.6980.

\bibitem[{LaRose and Coyle(2020)}]{larose2020robust}
LaRose, R.; and Coyle, B. 2020.
\newblock Robust data encodings for quantum classifiers.
\newblock \emph{Physical Review A}, 102(3): 032420.

\bibitem[{Lau et~al.(2022)Lau, Lim, Shrotriya, and Kwek}]{lau2022nisq}
Lau, J. W.~Z.; Lim, K.~H.; Shrotriya, H.; and Kwek, L.~C. 2022.
\newblock NISQ computing: where are we and where do we go?
\newblock \emph{AAPPS Bulletin}, 32(1): 27.

\bibitem[{LeCun, Bengio, and Hinton(2015)}]{lecun2015deep}
LeCun, Y.; Bengio, Y.; and Hinton, G. 2015.
\newblock Deep learning.
\newblock \emph{nature}, 521(7553): 436--444.

\bibitem[{LeCun, Cortes, and Burges(2010)}]{lecun2010mnist}
LeCun, Y.; Cortes, C.; and Burges, C. 2010.
\newblock MNIST handwritten digit database.
\newblock \emph{ATT Labs [Online]. Available: http://yann.lecun.com/exdb/mnist}, 2.

\bibitem[{Liu and Wittek(2020)}]{liu2020vulnerability}
Liu, N.; and Wittek, P. 2020.
\newblock Vulnerability of quantum classification to adversarial perturbations.
\newblock \emph{Physical Review A}, 101(6).

\bibitem[{Liu et~al.(2017)Liu, Chen, Liu, and Song}]{liu2017delving}
Liu, Y.; Chen, X.; Liu, C.; and Song, D. 2017.
\newblock Delving into Transferable Adversarial Examples and Black-box Attacks.
\newblock arXiv:1611.02770.

\bibitem[{Lloyd, Mohseni, and Rebentrost(2014)}]{lloyd2014quantum}
Lloyd, S.; Mohseni, M.; and Rebentrost, P. 2014.
\newblock Quantum principal component analysis.
\newblock \emph{Nature Physics}, 10(9): 631–633.

\bibitem[{Lu, Duan, and Deng(2020)}]{lu2020quantum}
Lu, S.; Duan, L.-M.; and Deng, D.-L. 2020.
\newblock Quantum adversarial machine learning.
\newblock \emph{Physical Review Research}, 2(3).

\bibitem[{Madry et~al.(2019)Madry, Makelov, Schmidt, Tsipras, and Vladu}]{madry2019towards}
Madry, A.; Makelov, A.; Schmidt, L.; Tsipras, D.; and Vladu, A. 2019.
\newblock Towards Deep Learning Models Resistant to Adversarial Attacks.
\newblock arXiv:1706.06083.

\bibitem[{Mitarai et~al.(2018)Mitarai, Negoro, Kitagawa, and Fujii}]{mitarai2018quantum}
Mitarai, K.; Negoro, M.; Kitagawa, M.; and Fujii, K. 2018.
\newblock Quantum circuit learning.
\newblock \emph{Physical Review A}, 98(3): 032309.

\bibitem[{Moosavi-Dezfooli et~al.(2017)Moosavi-Dezfooli, Fawzi, Fawzi, and Frossard}]{dezfooli2017}
Moosavi-Dezfooli, S.-M.; Fawzi, A.; Fawzi, O.; and Frossard, P. 2017.
\newblock Universal Adversarial Perturbations.
\newblock In \emph{2017 IEEE Conference on Computer Vision and Pattern Recognition (CVPR)}, 86--94.

\bibitem[{Nielsen and Chuang(2010)}]{nielsen2010quantum}
Nielsen, M.~A.; and Chuang, I.~L. 2010.
\newblock \emph{Quantum computation and quantum information}.
\newblock Cambridge university press.

\bibitem[{P{\'e}rez-Salinas et~al.(2020)P{\'e}rez-Salinas, Cervera-Lierta, Gil-Fuster, and Latorre}]{perez2020data}
P{\'e}rez-Salinas, A.; Cervera-Lierta, A.; Gil-Fuster, E.; and Latorre, J.~I. 2020.
\newblock Data re-uploading for a universal quantum classifier.
\newblock \emph{Quantum}, 4: 226.

\bibitem[{P{\'e}rez-Salinas et~al.(2021)P{\'e}rez-Salinas, L{\'o}pez-N{\'u}{\~n}ez, Garc{\'\i}a-S{\'a}ez, Forn-D{\'\i}az, and Latorre}]{perez2021one}
P{\'e}rez-Salinas, A.; L{\'o}pez-N{\'u}{\~n}ez, D.; Garc{\'\i}a-S{\'a}ez, A.; Forn-D{\'\i}az, P.; and Latorre, J.~I. 2021.
\newblock One qubit as a universal approximant.
\newblock \emph{Physical Review A}, 104(1): 012405.

\bibitem[{Pfeuty(1970)}]{PFEUTY197079}
Pfeuty, P. 1970.
\newblock The one-dimensional Ising model with a transverse field.
\newblock \emph{Annals of Physics}, 57(1): 79--90.

\bibitem[{Poursaeed et~al.(2018)Poursaeed, Katsman, Gao, and Belongie}]{poursaeed2018}
Poursaeed, O.; Katsman, I.; Gao, B.; and Belongie, S. 2018.
\newblock Generative Adversarial Perturbations.
\newblock In \emph{2018 IEEE/CVF Conference on Computer Vision and Pattern Recognition}, 4422--4431.

\bibitem[{Preskill(2018)}]{preskill2018quantum}
Preskill, J. 2018.
\newblock Quantum Computing in the NISQ era and beyond.
\newblock \emph{Quantum}, 2: 79.

\bibitem[{Rebentrost, Mohseni, and Lloyd(2014)}]{rebentrost2014quantum}
Rebentrost, P.; Mohseni, M.; and Lloyd, S. 2014.
\newblock Quantum Support Vector Machine for Big Data Classification.
\newblock \emph{Physical Review Letters}, 113(13).

\bibitem[{Ren et~al.(2022)Ren, Li, Xu, Wang, Jiang, Jin, Zhu, Chen, Song, Zhang et~al.}]{ren2022experimental}
Ren, W.; Li, W.; Xu, S.; Wang, K.; Jiang, W.; Jin, F.; Zhu, X.; Chen, J.; Song, Z.; Zhang, P.; et~al. 2022.
\newblock Experimental quantum adversarial learning with programmable superconducting qubits.
\newblock \emph{Nature Computational Science}, 2(11): 711--717.

\bibitem[{Schuld et~al.(2019)Schuld, Bergholm, Gogolin, Izaac, and Killoran}]{schuld2019evaluating}
Schuld, M.; Bergholm, V.; Gogolin, C.; Izaac, J.; and Killoran, N. 2019.
\newblock Evaluating analytic gradients on quantum hardware.
\newblock \emph{Physical Review A}, 99(3): 032331.

\bibitem[{Schuld et~al.(2020)Schuld, Bocharov, Svore, and Wiebe}]{schuld2020circuit}
Schuld, M.; Bocharov, A.; Svore, K.~M.; and Wiebe, N. 2020.
\newblock Circuit-centric quantum classifiers.
\newblock \emph{Physical Review A}, 101(3).

\bibitem[{Shende, Bullock, and Markov(2005)}]{shende2005synthesis}
Shende, V.~V.; Bullock, S.~S.; and Markov, I.~L. 2005.
\newblock Synthesis of quantum logic circuits.
\newblock In \emph{Proceedings of the 2005 Asia and South Pacific Design Automation Conference}, 272--275.

\bibitem[{Shor(1994)}]{shor1994algorithms}
Shor, P.~W. 1994.
\newblock Algorithms for Quantum Computation: Discrete Logarithms and Factoring.
\newblock In \emph{Proceedings 35th annual Symposium on Foundations of Computer Science}, 124--134. IEEE.

\bibitem[{Silver et~al.(2016)Silver, Huang, Maddison, Guez, Sifre, Van Den~Driessche, Schrittwieser, Antonoglou, Panneershelvam, Lanctot et~al.}]{silver2016mastering}
Silver, D.; Huang, A.; Maddison, C.~J.; Guez, A.; Sifre, L.; Van Den~Driessche, G.; Schrittwieser, J.; Antonoglou, I.; Panneershelvam, V.; Lanctot, M.; et~al. 2016.
\newblock Mastering the game of Go with deep neural networks and tree search.
\newblock \emph{nature}, 529(7587): 484--489.

\bibitem[{Stein et~al.(2021)Stein, Baheri, Chen, Mao, Guan, Li, Fang, and Xu}]{stein2021qugan}
Stein, S.~A.; Baheri, B.; Chen, D.; Mao, Y.; Guan, Q.; Li, A.; Fang, B.; and Xu, S. 2021.
\newblock Qugan: A quantum state fidelity based generative adversarial network.
\newblock In \emph{2021 IEEE International Conference on Quantum Computing and Engineering (QCE)}, 71--81. IEEE.

\bibitem[{Szegedy et~al.(2014)Szegedy, Zaremba, Sutskever, Bruna, Erhan, Goodfellow, and Fergus}]{szegedy2014intriguing}
Szegedy, C.; Zaremba, W.; Sutskever, I.; Bruna, J.; Erhan, D.; Goodfellow, I.; and Fergus, R. 2014.
\newblock Intriguing properties of neural networks.
\newblock arXiv:1312.6199.

\bibitem[{West et~al.(2023)West, Erfani, Leckie, Sevior, Hollenberg, and Usman}]{west2023benchmarking}
West, M.~T.; Erfani, S.~M.; Leckie, C.; Sevior, M.; Hollenberg, L. C.~L.; and Usman, M. 2023.
\newblock Benchmarking adversarially robust quantum machine learning at scale.
\newblock \emph{Physical Review Research}, 5(2).

\bibitem[{Xiao et~al.(2019)Xiao, Li, Zhu, He, Liu, and Song}]{xiao2019generating}
Xiao, C.; Li, B.; Zhu, J.-Y.; He, W.; Liu, M.; and Song, D. 2019.
\newblock Generating Adversarial Examples with Adversarial Networks.
\newblock arXiv:1801.02610.

\bibitem[{Xiao, Rasul, and Vollgraf(2017)}]{xiao2017fashion}
Xiao, H.; Rasul, K.; and Vollgraf, R. 2017.
\newblock Fashion-MNIST: a Novel Image Dataset for Benchmarking Machine Learning Algorithms.
\newblock arXiv:1708.07747.

\bibitem[{Xu et~al.(2020)Xu, Ma, Liu, Deb, Liu, Tang, and Jain}]{xu2020adversarial}
Xu, H.; Ma, Y.; Liu, H.-C.; Deb, D.; Liu, H.; Tang, J.-L.; and Jain, A.~K. 2020.
\newblock Adversarial attacks and defenses in images, graphs and text: A review.
\newblock \emph{International Journal of Automation and Computing}, 17: 151--178.

\bibitem[{Zhang et~al.(2020)Zhang, Sheng, Alhazmi, and Li}]{zhang2020adversarial}
Zhang, W.~E.; Sheng, Q.~Z.; Alhazmi, A.; and Li, C. 2020.
\newblock Adversarial Attacks on Deep-Learning Models in Natural Language Processing: A Survey.
\newblock \emph{ACM Transactions on Intelligent Systems and Technology}, 11(3): 1--41.

\bibitem[{Zhao et~al.(2023)Zhao, Zhou, Li, Tang, Wang, Hou, Min, Zhang, Zhang, Dong, Du, Yang, Chen, Chen, Jiang, Ren, Li, Tang, Liu, Liu, Nie, and Wen}]{zhao2023survey}
Zhao, W.~X.; Zhou, K.; Li, J.; Tang, T.; Wang, X.; Hou, Y.; Min, Y.; Zhang, B.; Zhang, J.; Dong, Z.; Du, Y.; Yang, C.; Chen, Y.; Chen, Z.; Jiang, J.; Ren, R.; Li, Y.; Tang, X.; Liu, Z.; Liu, P.; Nie, J.-Y.; and Wen, J.-R. 2023.
\newblock A Survey of Large Language Models.
\newblock arXiv:2303.18223.

\end{thebibliography}
\end{document}